\documentclass[lettersize,journal]{IEEEtran}
\usepackage{amsmath,amsfonts}
\usepackage{algorithmic}
\usepackage{algorithm}
\usepackage{array}
\usepackage[caption=false,font=normalsize,labelfont=sf,textfont=sf]{subfig}
\usepackage{textcomp}
\usepackage{stfloats}
\usepackage{url}
\usepackage{verbatim}
\usepackage{graphicx}
\usepackage{cite}
\usepackage{booktabs}
\usepackage[normalem]{ulem}
\usepackage{amsmath,amsfonts,bm}

\def\eqref#1{equation~\ref{#1}}
\def\1{\bm{1}}

\def\vx{{\bm{x}}}

\def\mM{{\bm{M}}}

\def\mX{{\bm{X}}}

\DeclareMathAlphabet{\mathsfit}{\encodingdefault}{\sfdefault}{m}{sl}
\SetMathAlphabet{\mathsfit}{bold}{\encodingdefault}{\sfdefault}{bx}{n}

\def\gG{{\mathcal{G}}}

\def\sV{{\mathbb{V}}}

\newcommand{\E}{\mathbb{E}}

\newcommand{\R}{\mathbb{R}}

\usepackage{xcolor}
\usepackage{subfig}
\usepackage{placeins}

\begin{document}

\title{Online Continual Graph Learning}

\author{Giovanni Donghi, Luca Pasa, Daniele Zambon \IEEEmembership{Member, IEEE},\\ Cesare Alippi \IEEEmembership{Fellow, IEEE}, Nicolò Navarin \IEEEmembership{Member, IEEE}
\thanks{This research was supported by the European Union - NextGenerationEU as part of the Italian National Recovery and Resilience Plan (PNRR), and the project ``Lifelong Learning on large-scale and structured data'', funded by the EC-funded OCRE project.} 
\thanks{Giovanni Donghi, Luca Pasa and Nicolò Navarin are with the Department of Mathematics, University of Padua, Padua, Italy. Daniele Zambon is with the Faculty of Informatics, Università della Svizzera Italiana, Switzerland. Cesare Alippi is with the Faculty of Informatics, Università della Svizzera Italiana, Switzerland, and also with the Department of Electronics, Information and Bioengineering, Politecnico di Milano, Milan, Italy.}
\thanks{\textit{Corresponding author: Giovanni Donghi, giovanni.donghi@phd.unipd.it}}
}

\markboth{Preprint submitted to IEEE TRANSACTIONS ON NEURAL NETWORKS AND LEARNING SYSTEMS}{}

\maketitle

\begin{abstract}
Continual Learning (CL) aims to incrementally acquire new knowledge while mitigating catastrophic forgetting. Within this setting, Online Continual Learning (OCL) focuses on updating models promptly and incrementally from single or small batches of observations from a data stream. 
Extending OCL to graph-structured data is crucial, as many real-world networks evolve over time and require timely, online predictions. 
However, existing continual or streaming graph learning methods typically assume access to entire graph snapshots or multiple passes over tasks, violating the efficiency constraints of the online setting. 
To address this gap, we introduce the Online Continual Graph Learning (OCGL) setting, which formalizes node-level continual learning on evolving graphs under strict memory and computational budgets. 
OCGL defines how a model incrementally processes a stream of node-level information while maintaining anytime inference and respecting resource constraints. 
We further establish a comprehensive benchmark comprising seven datasets and nine CL strategies, suitably adapted to the OCGL setting, enabling a standardized evaluation setup. 
Finally, we present a minimalistic yet competitive baseline for OCGL, inspired by our benchmarking results, 
that achieves strong empirical performance with high efficiency.
\end{abstract}


\section{Introduction}

In standard machine learning, models are trained once on a fixed dataset, assuming independent and identically distributed samples. 
Real-world environments, however, often generate data in chunks or streams, undergoing shifts in the data distribution or even variations in tasks to be solved, which requests often impractical and expensive periodic retraining. Continual Learning (CL) \cite{parisi_continual_2019,de_lange_continual_2022} addresses this challenge by enabling models to learn incrementally while retaining past knowledge and without requiring all data to remain available.

In the more restrictive \emph{online} learning setting, training data points are collected sequentially and must be processed by the learning method as soon as they appear and  in real-time. Once processed, each sample is typically discarded leaving the learner with no access to past data \cite{chaudhry_efficient_2018,mai_online_2022}. Such strict environments are found in monitoring and control problems \cite{zliobaite_overview_2016,gunasekara_survey_2023} where decisions must be made continuously and under limited resources.
Building on this, Online Continual Learning (OCL) represents a particularly challenging setting in which models must rapidly adapt to evolving data streams using small batches that are observed only once. 
Additionally, they are expected to support anytime inference, meaning they are expected to generate accurate predictions at arbitrary points in time, even before training can be considered completed.
At the same time, OCL systems must operate with minimal computational and memory cost, and mitigate forgetting of previously acquired knowledge.

Recently, CL has been extended to graph-structured data, giving rise to the field of Continual Graph Learning (CGL) \cite{yuan_continual_2023}.  Indeed, many machine learning tasks involve graph representations of data, such as social networks, citation networks, biological systems, and transaction networks.
Additionally, most graphs in the real world are not static: they continuously evolve, experiencing the addition/removal of nodes and changes to their topology. Examples include the growth of social networks, the appearance of new publications in citation networks, and changes in road conditions \cite{liu_overcoming_2021,zhou_overcoming_2021}. However, most existing CGL methods operate in \emph{offline} and task-wise fashions, training on subgraph snapshots with multiple passes until convergence. 
Such settings fail to meet the core requirements of online learning -- single-pass updates, limited budget, and anytime predictions -- and overlook specific challenges of dynamic graphs. 
For instance, multi-hop message passing in Graph Neural Networks (GNNs) leads to unbounded computational growth as graphs densify due to the inclusion of new nodes and edges. We refer to this issue as the neighborhood expansion problem, which we analyze in this work.
These additional constraints of the OCGL setting render most existing methods unsuitable for realistic, inherently evolving environments.

To address these shortcomings, we introduce the Online Continual Graph Learning (OCGL) setting, which unifies continual and online learning principles for node-level graph streams. 
We formalize the incremental acquisition of graph information under these constraints and analyze the performance of existing methods when adapted to this setting, along with their practical implications.
Our main contributions are summarized as follows.
\begin{enumerate}
\item We formalize OCGL as a principled setting bridging OCL and CGL, establishing a foundation for CL in environments characterized by data streams of node-level information.
\item We highlight and discuss the neighborhood expansion problem, which OCGL introduces, and that can break the computation and memory requirements if not properly handled.  
We present a simple viable solution to address this problem.
\item We design a benchmarking environment for OCGL, encompassing seven datasets and several existing methods from the literature, suitably adapted to operate within the OCGL setting. Our findings reveal higher performance of replay-based methods that are tailored to preserve topological information.
\item We introduce LINEAR, a simple and lightweight method inspired by the observations drawn from the benchmark results.
Despite its simplicity, LINEAR achieves competitive performance, and it constitutes a strong baseline to assess the quality of new OCGL methods.
\end{enumerate}
We believe that this work lays a strong foundation for systematic progress in CL, fostering the development of sound and more effective approaches for OCGL, while LINEAR offers a strong and reliable reference for future research, setting a clear benchmark that new methods should meet or exceed to demonstrate meaningful progress.

\section{Background and related works}\label{sec:related_works}

\subsection{Continual Learning} 

Depending on the type of shift in the data distribution, CL has been categorized into three main scenarios \cite{van_de_ven_three_2022}: in \textit{task-incremental} learning, the model sequentially learns distinct tasks, which requires availability of task identifiers to make predictions; 
\textit{class-incremental} learning consists in classifying instances with an increasing number of classes, without task identifiers; finally, \textit{domain-incremental} learning requires solving the same problem in different contexts. 
In the past, CL was mainly applied to reinforcement learning \cite{kirkpatrick_overcoming_2017,rolnick_experience_2019} and computer vision \cite{rebuffi_icarl_2017,lopez-paz_gradient_2017,aljundi_memory_2018,li_learning_2018,masana_class-incremental_2022,mai_online_2022}, but most of the methods that have been developed to address these problem domains can be used for a wide range of other machine learning tasks. CL approaches to mitigate forgetting fall into three general categories \cite{de_lange_continual_2022}: \textit{regularization}, \textit{replay} and \textit{architectural} methods. Regularization methods \cite{kirkpatrick_overcoming_2017,zenke_continual_2017,aljundi_memory_2018,li_learning_2018,chaudhry_riemannian_2018} introduce additional loss terms to preserve important parameters to retain previously acquired knowledge. Replay methods \cite{rebuffi_icarl_2017,lopez-paz_gradient_2017,chaudhry_tiny_2019,chaudhry_efficient_2018} use a memory buffer to store some representative samples from old tasks, to use them jointly with new samples to update model parameters. Architectural methods \cite{fernando_pathnet_2017,masse_alleviating_2018,rusu_progressive_2022} avoid changes to model weights either by gating mechanisms or by introducing new parameters, allowing the model to grow.

\subsection{Online Continual Learning} 

In the usual CL scenarios described above, data arrive one task at a time, allowing \textit{offline} training with multiple passes and shuffles over the data for the current task \cite{de_lange_continual_2022}. \textit{Online Continual Learning (OCL)} \cite{chaudhry_efficient_2018,mai_online_2022,soutifcormerais_comprehensive_2023} addresses the more realistic case where data arrive in small batches of only few samples, without the possibility for the model to store all the data for the current task, either for privacy reasons or memory limitations. 
In this setting, the algorithm must efficiently learn from each mini-batch in a nonstationary stream. Additionally, whereas for CL we assume to know the task boundaries, OCL can be performed in a boundary-agnostic setting, or task-free, allowing for diverse distribution shifts \cite{koh_online_2021}. However, many CL methods are not suited to this setting and require modifications. An additional characteristic of OCL is anytime inference: the model should always be up-to-date and ready to make predictions online after each training batch, reacting quickly to distribution shifts \cite{koh_online_2021}.

\subsection{Learning on graphs} 

\textit{Graph Neural Networks (GNN)} \cite{sperduti_supervised_1997,scarselli_graph_2009,micheli_neural_2009,kipf2017semisupervised} have emerged as the state-of-the-art approach for dealing with network data, generalizing convolution to graph structures. The core mechanism of most GNNs is message passing \cite{gilmer_neural_2017}: at each layer, the hidden embedding $h_v^{(k)}$ of each node $v$ is updated using information from its neighborhood $\mathcal{N}(v)$ as $\smash{h_v^{(l+1)} = \text{UPDATE}(h_v^{(l)}, \text{AGGREGATE}(\{h_u^{(l)}: u \in \mathcal{N}(v)\}))}$. Here AGGREGATE and UPDATE are differentiable functions specified by the particular model. Specifically, as at each step each node updates its embedding using the information (message) coming from its neighbors, after $l$ layers it will depend on its $l$-hop neighborhood. 
Graph-based processing of temporal data has a relatively short history, primarily encompassing the study of temporal graphs \cite{kazemi2020representation, gravina2024deep, Liu_2025_Multiview, Liu_2025_Rethinking} and time series data \cite{cini2023graph, jin2024survey} with dedicated adaptation strategies to deal with evolving graphs \cite{cini2023taming} and benchmarks \cite{huang2023temporal}.

\subsection{Continual Graph Learning} 

In the last few years, researchers have started to develop CL strategies tailored to graph data \cite{wang_streaming_2020}, with applications such as recommender systems \cite{xu_graphsail_2020} and traffic prediction \cite{ijcai2021p498}. Most \textit{Continual Graph Learning (CGL)} methods adapt general CL strategies, focusing on preserving topological information with a loss term on neighborhood aggregation parameters \cite{liu_overcoming_2021}, or specific node selection policies to replay informative nodes \cite{zhou_overcoming_2021}. Recently, a number of surveys have been published on the topic \cite{febrinanto_graph_2023,yuan_continual_2023,zhang_continual_2024,tian_continual_2024}, and a benchmark has been proposed \cite{zhang_cglb_2022}.
Importantly, CGL differs from other problem domains because of dependencies introduced by graph structure, requiring careful consideration. Specifically, we can distinguish between \textit{graph-level} CGL and \textit{node-level} CGL \cite{zhang_cglb_2022}. In graph-level CGL, each sample is an independent graph and standard CL methods apply directly \cite{carta_catastrophic_2022}, while node-level CGL performs predictions within a single evolving graph.
In node-level CGL, each task consists of a new subgraph, for example with new classes of nodes. Specifically, the task subgraph arrives all at once, and offline training is performed on it. 
A key issue is the treatment of inter-task edges \cite{tian_continual_2024}: since GNNs aggregate neighbor information, edges to nodes from previous tasks may implicitly expose past data.
In practice, inter-task edges are often kept, but without access to the labels from past tasks \cite{zhang_continual_2024}. Finally, adding new nodes changes previous neighborhoods, causing structural shift \cite{Su_2023_RobustGraphIncremental}, a source of backward interference.

\section{Online Continual Graph Learning}\label{sec:ocgl}

The OCL setting has been explored in domains such as computer vision \cite{mai_online_2022,soutifcormerais_comprehensive_2023} and sequences \cite{parisi_online_2020}, but it has not been thoroughly investigated for graph-structured data. Some papers on CGL consider a setting referred to as \textit{streaming} \cite{wang_streaming_2020,perini_learning_2022}, yet the approaches can be categorized as offline CL as the streams consist of graph snapshots, on which models are trained with multiple passes. While some CGL methods could be used in an online fashion \cite{febrinanto_graph_2023}, the practical implications and constraints of the online setting for graph data have not been investigated yet.
Table \ref{tab:properties} summarizes these differences and motivates the new setting introduced here.

We introduce \textit{Online Continual Graph Learning (OCGL)}, a new setting that ports CGL to the online problem setting. Specifically, OCGL is applicable to dynamic real-world scenarios such as social networks or recommender systems, where sudden distribution changes occur, and quick model adjustments are essential for anytime predictions. We focus on the study of node classification, although the setting can be adapted to regression problems or edge-level tasks. In this section we describe the general setting and its associated principles, requirements and challenges, while in Section \ref{sec:setup} we will specify the instantiation of OCGL, such as the particular stream construction, used in our experiments.

\begin{figure*}[ht!]
    \centering
    \includegraphics[width=1\textwidth]{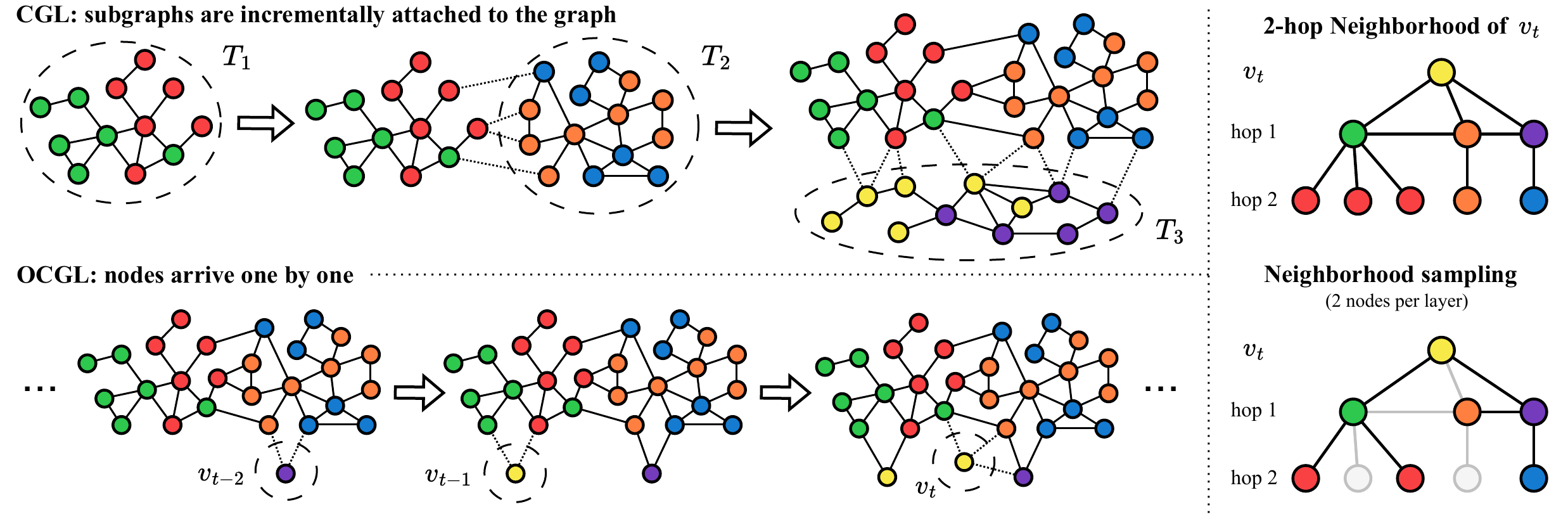}%
    \caption{Illustration of the different graph evolution under CGL and OCGL. \textbf{Top left}: in CGL task subgraphs are incrementally attached to the existing graph (and training is performed \textit{offline} until convergence on the subgraphs). \textbf{Bottom left}: in OCGL individual nodes are attached to the graph in order of their arrival (and training is performed \textit{online} in one pass on individual or mini-batches of nodes). \textbf{Right}: the size of the observed 2-hop neighborhood of a node is kept bounded by using neighborhood sampling. In the example, 2 neighbors of $v_t$ are sampled, and then recursively 2 neighbors for each of them.}
    \label{fig:diagram}
\end{figure*}

\subsection{A growing network} 

We model the data associated with an OCGL problem as an evolving graph $\gG$ induced by a stream of nodes $v_1, v_2, \ldots, v_t, \ldots$ added in succession. At each time step $t$, the graph snapshot $\gG^t = (\sV^t, \E^t, \mX^t)$ is defined by node set $\sV^t=\{v_i\}_{i\le t}=\{v_1,\dots,v_t\}$, edge set $\E^t\subseteq  \sV^t \times\sV^t$, and a set of node attributes $\mX^t=\{\vx^i\}_{i\le t}\subset \mathbb R^{F}$. Edge attributes can be considered as well, however they are not discussed here to ease the presentation. 
The graph nodes $v_i$ can be associated with class labels $y_i\in \{1,\dots, C\}$ to be predicted and/or used as training samples to learn the model.
At each time step $t$ a new node is added and connected to graph $\gG^{t-1}$. Specifically, a tuple $(v_t, \mathcal{N}(v_t), \vx^t)$ containing a new node index $v_t\not\in\sV^{t-1}$, associated node features $\vx_t$, and a set of neighbors $\mathcal{N}(v_t) \subseteq \sV^{t-1}$ is presented and used to connect $v_t$ to graph $\gG^{t-1}$ according to the relations contained in $\mathcal{N}(v_t)$. Finally, the target class label $y_t$ of node $v_t$ may or may not be acquired contextually to $(v_t, \mathcal{N}(v_t), \vx^t)$: for instance, a prediction for node $v_t$ might be requested at time $t$ while the true class label $y_t$ is observed only at a later time.

The formulation of the evolving graph is general, as it does not make assumptions on the distribution shifts happening in the node stream. It can be easily adapted or made more specific: while a real-world stream could be induced by a time-stamp on the nodes, this setting can be derived from any static graph by establishing an ordering on the nodes. The three CL scenarios of task-, class- and domain-incremental can thus be easily adapted to this online setting by ordering nodes by task, similarly to what is done in other domains \cite{mai_online_2022,soutifcormerais_comprehensive_2023}.

\newcommand{\present}{{\color{green!60!black}$\boldsymbol\surd$}}
\newcommand{\absent}{{\color{red!60!black}$\boldsymbol\times$}}
\newcommand{\kinda}{{\color{blue!60!black}$\boldsymbol\sim$}}
\begin{table}[]
\caption{Properties of the different continual learning settings.}
\label{tab:properties}
\begin{center}
\renewcommand{\arraystretch}{1.25}
\begin{tabular}{lcccc}
\toprule
\textbf{Properties}                  & \textbf{CL} & \textbf{OCL} & \textbf{CGL} & \textbf{OCGL} \\ \midrule
Incremental learning                 & \present     & \present      & \present      & \present       \\
Graph representation learning        & \absent    & \absent     & \present      & \present       \\
Single pass over stream              & \absent    & \present      & \absent     & \present       \\
Anytime predictions                  & \absent    & \present      & \absent     & \present       \\
Efficiency constraints (memory)      & \present     & \present      & \present      & \present       \\
Efficiency constraints (computation) & \absent    & \present      & \absent     & \present       \\
Neighborhood expansion problem       & \absent    & \absent     & \kinda     & \present       \\ \bottomrule
\end{tabular}
\vspace{-1em}
\end{center}
\end{table}

\subsection{Problem statement}

Given a model $f_\theta$, the objective of OCGL is to incrementally update $\theta$ using only the information from the current node $v_t$ and its $L$-hop neighborhood in $\gG^t$, or using small mini-batches, slightly weakening the online setting as commonly done in the literature \cite{chaudhry_efficient_2018}.
In line with CL principles, while adapting to the evolving stream the model must also retain previously acquired knowledge.
The key difference with standard CGL is that in OCGL, as in OCL, the model needs to adapt quickly in order to perform anytime predictions on the node stream \cite{koh_online_2021}. This objective introduces specific constraints (see also Table \ref{tab:properties}):
\begin{itemize}
  \item New nodes arrive individually or in small mini-batches that are processed once; after training, each batch is discarded, possibly except for a limited replay buffer.
  \item Differently from CGL, small mini-batches may not form meaningful self-contained subgraphs, hence inter-batch edges are naturally present and must be handled consistently.
  \item Each update must have bounded computational and memory cost, regardless of the total size of $\gG^t$.
  \item The model must provide anytime inference, quickly adapting to new knowledge while retaining previously acquired one: after each update it can be queried on past nodes $v_{t-k}$ using the current graph state $\gG^t$.
\end{itemize}

Satisfying these constraints guarantees efficiency and scalability as the graph grows to arbitrary size, but it poses non-negligible issues associated with reiterated message passing within multi-layer GNNs. We delve deeper into these complexities below, and we illustrate the difference of graph evolution compared to GCL in Figure \ref{fig:diagram}.

\begin{figure}[b!]
\centering
\subfloat[CoraFull, $|\sV|=19{,}793$]{%
  \includegraphics[width=0.41\textwidth]{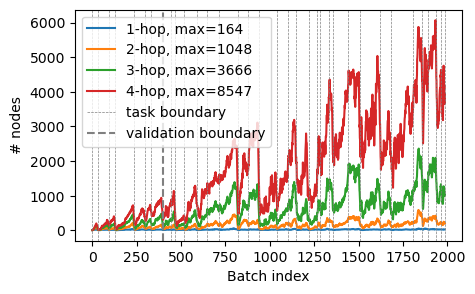}%
  \label{fig:nodes_cora}}
\\
\subfloat[Arxiv, $|\sV|=169,343$]{%
  \includegraphics[width=0.41\textwidth]{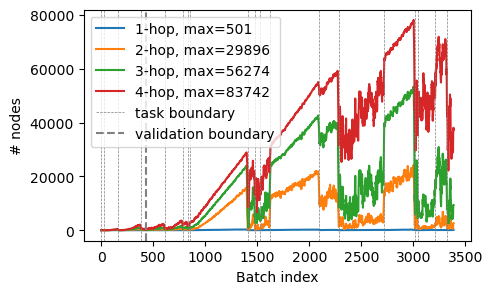}%
  \label{fig:nodes_arxiv}}
\\
\subfloat[Reddit, $|\sV|=227,853$]{%
  \includegraphics[width=0.41\textwidth]{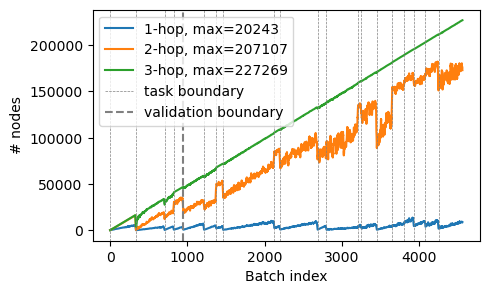}%
  \label{fig:nodes_reddit}}
\\
\subfloat[Amazon Computer, $|\sV|=13,752$]{%
  \includegraphics[width=0.41\textwidth]{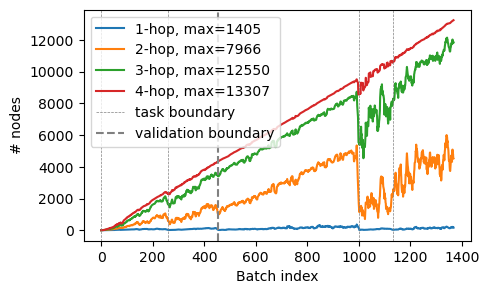}%
  \label{fig:nodes_computer}}
\caption{Number of nodes in the union of $l$-hop neighborhoods of each training batch. Smoothed with rolling average over windows of 10 batches for readability, maximum is reported in the legend. 
}
\label{fig:neighborhood_expansion}
\end{figure}

\subsection{Neighborhood expansion problem} 

At each of the $L$ layers of the GNN, the network aggregates the embeddings of neighboring nodes, thus requiring access to their $L$-hop neighborhoods. The neighborhood size however scales as $O(d^L)$ where $d$ is the average degree, which tends to increase as the graph grows: real-world graphs exhibit densification over time, with the number of edges following a power law of the number of nodes, and a shrinking of graph diameter over time \cite{leskovec_graphs_2005,shah_growing_2019}.
In many real graphs, only a few hops suffice to connect most nodes, so large $d$ or $L$ implies processing almost the entire graph per mini-batch, violating OCGL’s efficiency constraints.
It is therefore of utmost importance to limit $L$ or to introduce a strategy to deal with  $d$. We note that neighborhood expansion is also present in CGL, but it is not in contrast with the requirements of that setting, and is thus not discussed in the literature.

To illustrate this phenomenon, Figure \ref{fig:neighborhood_expansion} shows the size of the $l$-hop neighborhood across mini-batches in the node stream for four datasets considered in our experiments (see Section \ref{sec:setup} for further details). The Reddit graph in particular is very well connected, with two hops containing the majority of the graph. CoraFull and Arxiv show a more moderate neighborhood expansion, while two hops on the Amazon Computer graph cover about half of the nodes.

\subsection{Mini-batch acquisition}\label{sec:minibatch_acquisition}

As discussed, the growth of node neighborhoods makes it infeasible to aggregate information from all $L$-hop neighbors while maintaining a bounded computation. To satisfy the efficiency requirements, the model must therefore operate on a limited subset of nodes.
This issue is similar to the problem of scaling static GNNs for large graphs, where mini-batch training is required both for memory and efficiency reasons. Numerous approaches have been developed, such as fixing a number of neighbors sampled for aggregation \cite{hamilton_inductive_2017,chen_fastgcn_2018} or training on partitions of the graph \cite{chiang_cluster-gcn_2019}.
In our context, the simplest solution to guarantee compliance with OCGL requirements is to cap the number of neighboring nodes through any form of sub-sampling, thus guaranteeing an upper bound on the size of the computational graph for each batch.
Thus, for any node $v$, an OCGL model cannot leverage the full $L$-layer ego-graph $\gG^t_{v,L}$ within $\gG^t$. Instead, it will only receive a sub-sampled $\widetilde\gG^t_{v,L}$.

The way in which the sub-sampled $L$-hop neighborhoods are obtained starting from the data generating process of the growing graph may depend on the specific application, but we can generally conceptualize two distinct systems, which are nevertheless equivalent regarding the learning process of the model. In one case, an up-to-date snapshot $\gG^t$ is assumed to be stored in a {Past Information Store (PIS)} system -- distinct from an eventual memory buffer associated with predictive model $f_\theta$ -- as in the more general lifelong-learning system setup \cite{chen_lifelong_2018}. 
We do not impose memory limitations on the PIS to allow graph growth, but we still require the training on each mini-batch to have bounded computational time and memory cost, and assume to only have access to the labels of nodes in the current batch. 
In the other case, mini-batches are directly received in the form of sub-sampled neighborhoods from an external data source. This could be the case when privacy concerns are involved, or in case of huge graphs where a local exploration is performed. 

\section{Strategies for OCGL}

\subsection{Adaptation of existing methods}

Having defined the Online Continual Graph Learning setting, we consider and evaluate some popular CL techniques, most of which are agnostic with respect to the type of the input data. 
Some CGL learning strategies are easily adapted to the online setting, such as PDGNN \cite{Zhang_2024} and SSM \cite{zhang_sparsified_2022}, while others are not directly applicable, such as ER-GNN \cite{zhou_overcoming_2021}, which stores representative nodes according to metrics computed offline on an entire graph snapshot, and thus we simplify it as described below.
Similar efficiency considerations exclude many more recent CGL methods, which may require expensive steps and/or access to most of the graph: to name a few, CaT \cite{liu_cat_2023} requires access to the task subgraph for condensation, UGCL \cite{hoang_universal_2023} uses a global structure distillation which requires to compute node embeddings for the entire graph, and a local structure distillation which still requires to compute the full embedding of buffer nodes and all their neighbors, and MSCGL \cite{Cai_2022_MultimodalContinualGraph} performs a costly neural architecture search for each task. Additionally, baselines that require expensive fine-tuning steps such as GDumb \cite{prabhu_gdumb_2020} are excluded, as they would violate the online setting. Several strategies natively require task boundaries, and have been modified for the task-free setting as described below.
\begin{itemize}
    \item \textbf{A-GEM.} Averaged GEM \cite{chaudhry_efficient_2018} is a more efficient version of GEM \cite{lopez-paz_gradient_2017}, which ensures that the average loss for past tasks does not increase. It achieves this by projecting the gradient of the incoming batch in the orthogonal space of the gradient computed on samples from a memory buffer, if their scalar product is negative. We select buffer nodes with reservoir sampling \cite{vitter_random_1985}
    \item \textbf{ER.} Experience replay \cite{chaudhry_tiny_2019} is a simple yet powerful replay-based method, which selects samples to be stored in a memory buffer by reservoir sampling \cite{vitter_random_1985}. New incoming batches for training are then augmented with nodes sampled uniformly from the buffer. 
    \item \textbf{EWC.} Elastic Weight Consolidation \cite{kirkpatrick_overcoming_2017} adds a quadratic term to the loss to penalize the modification of important parameters. Parameter importance is approximated by the diagonal of the Fisher information matrix, which needs to be computed offline for each task. We therefore modify the algorithm to keep one single Fisher information matrix updated with a running average over the batches, similarly to the MAS approach detailed later. Another approach would be to keep a moving average, as done in EWC++ \cite{chaudhry_riemannian_2018}.
    \item \textbf{LwF.} Learning without Forgetting \cite{li_learning_2018} uses distillation \cite{hinton_distilling_2015} to regularize the loss with logits from a previous version of the model (teacher) on the current batch. To use it in a task-free setting, we introduce an additional hyperparameter: the number of batches after which the teacher is updated.
    \item \textbf{MAS.} Memory Aware Synapses \cite{aljundi_memory_2018} is a quadratic regularization similar to EWC, but it calculates importance as the sensitivity of the output on parameters. MAS is natively an online method, as the importance scores are accumulated with each new data point.
    \item \textbf{PDGNN.} Parameter Decoupled GNNs \cite{Zhang_2024} use a Topology-aware Embedding Memory, storing embedding vectors obtained with SGC \cite{Wu_2019_SimplifyingGraphConvolutional}, thus preserving neighborhood information. These vectors are used for experience replay with an MLP model.
    \item \textbf{SSM.} Sparsified Subgraph Memory \cite{zhang_sparsified_2022} consists in storing sparsified computational graphs of nodes in the memory buffer, allowing to leverage topological information in replay methods. We use this memory buffer by selecting nodes via reservoir sampling \cite{vitter_random_1985}, as support for both ER and A-GEM strategies.
    \item \textbf{TWP.} Topology-aware Weight Preserving \cite{liu_overcoming_2021} is another regularization method, which preserves important weights for topological aggregation in GAT \cite{velickovic_graph_2018}, generalized also to other GNNs. We modify it for the online setting as EWC.
\end{itemize}

\subsection{A new baseline method}

The main purpose of this section is introducing a minimal, efficient and competitive method that can serves as reference for future OCGL research. 
Following the insights obtained from our results in Section~\ref{sec:neighborhood}, we introduce \textbf{LINEAR} (Lightweight Incremental NEighborhood Aggregation with Replay), a new baseline inspired by PDGNN and pushing its design philosophy to the limit with regard to simplicity.
For each node, LINEAR first computes a feature vector by averaging features of its neighbors. A lightweight linear classifier is then applied to these aggregated features.

More specifically, LINEAR consists of a single SGC layer \cite{Wu_2019_SimplifyingGraphConvolutional} that averages up to $r$ feature vectors sampled from the 1-hop neighborhoods, followed by a multinomial logistic regression classifier -- no hidden layers, no nonlinearities, and no iterative message passing are employed; $r$ is a model hyperparameter to meet the requested computational budget as described in Section \ref{sec:minibatch_acquisition}. We couple this with an experience replay mechanism, storing directly the neighborhood-averaged features via reservoir sampling. 

The computational complexity of LINEAR is significantly lower than standard GNN-based models. Assume for convenience fixed feature dimension $F$ and number of sampled neighbors $r$ throughout all layers, and a $C$-class linear classifier on top. Due to message passing, an $L$-layer GNN requires processing $O(r^{L})$ nodes to compute the embedding of a single target node. Each node transformation applies an $F \times F$ linear map, and when $F > r + C$ this cost dominates both aggregation and classification. The overall complexity of predicting the class of a single node is therefore $O(r^{L} F^{2})$, reflecting the neighborhood expansion problem.
In contrast, LINEAR requires only $O(rF + FC)$ cost, coming from the neighbor aggregation and the final linear classifier, respectively.

Despite its simplicity, we show in Section \ref{sec:neighborhood} that LINEAR exhibits strong empirical performance in the OCGL benchmarks, often surpassing GNN-based methods.

\section{Experimental Setup}\label{sec:setup}

In this section, we introduce the specific experimental setup used, describing the construction of the node streams from  benchmark datasets, and the details of model training and evaluation.

\subsection{Benchmarks}

Seven node classification graph datasets are used in our experiments: five homophilous multiclass datasets CoraFull \cite{bojchevski_deep_2018}, Arxiv \cite{hu_open_2021}, Reddit \cite{hamilton_inductive_2017}, Amazon Computer \cite{shchur_pitfalls_2019} and Products \cite{hu_open_2021}, the heterophilous Roman Empire \cite{Platonov_2022_CriticalLookEvaluation} and the binary classification dataset Elliptic \cite{Weber_2019_AntiMoneyLaunderingBitcoin}. The datasets are described in Appendix \ref{app:dataset}. As is common in the OCL literature \cite{mai_online_2022,soutifcormerais_comprehensive_2023}, in order to position our experiments close to the rest of the CL literature, for all datasets except for Elliptic we devise a node stream derived from the class-incremental CL setting, which is considered the most challenging one for catastrophic forgetting \cite{masana_class-incremental_2022}. We divide the nodes in the graph into groups with fixed order consisting of 2 classes: this would be the sequence of two-class tasks in class-incremental learning (resulting in 35 tasks for CoraFull, 20 tasks for Arxiv and Reddit, 5 for Amazon Computer, 23 for Products and 9 for Roman Empire).
Then, we fix an ordering on the nodes of each task, and we stream the nodes accordingly. Therefore, the graph will gradually grow with mini-batches of nodes from two new classes at a time, which are processed in an online fashion. This allows us to consider metrics from the CL literature which require task boundaries, even though in our experiments the learning algorithm itself is task agnostic and simply adds a new output neuron when an instance of a new class is observed. 
For Elliptic and Arxiv, that have time information available, we construct a time-incremental stream, that is according to the node real timestamps, and we divide the nodes into 10 tasks solely for the purposes of evaluation.
For each dataset, we split the graph into 60\% for training, 20\% for validation and 20\% for testing. A transductive setting is used: validation and test nodes are not used for loss computation, but they are still used for message passing. 

\subsection{Performance assessment} 

We consider three widely adopted metrics in the literature: \textit{Average Performance (AP)}, \textit{Average Forgetting (AF)} \cite{lopez-paz_gradient_2017}, and \textit{Average Anytime Performance (AAP)} \cite{caccia_new_2021}.
AAP is obtained by evaluating the model after each training batch, which we refer to as anytime evaluation, and thus allows us to understand model performance over time. For the highly unbalanced Elliptic dataset, we consider the F1 score as performance metric, while for all other datasets accuracy is used.
More details are reported in Appendix~\ref{app:Metrics}.

\subsection{Training details}

In our experiments the backbone for all CL strategies is the \textit{Graph Convolutional Network (GCN)} \cite{kipf2017semisupervised}, specifically a 2-layer GCN with a fixed hidden dimension of 256 units as done by Zhang et~al. \cite{zhang_cglb_2022}, with the exception of PDGNN which uses 2 layers of SGC \cite{Wu_2019_SimplifyingGraphConvolutional} followed by a 2 layer MLP with 256 units, and LINEAR which simply averages features in the immediate neighborhood and uses a linear classifier on top.
We use Adam optimizer \cite{kingma_adam_2017} without weight decay, tuning the learning rate as a hyperparameter with the protocol defined below. 
We consider the batch size to be fixed, as it could depend on the real-world problem. For the smaller datasets CoraFull, Amazon Computer, and Roman Empire we consider batches of 10 nodes, while for the larger Arxiv, Reddit, and Elliptic we use 50, and for the much larger Products we use 250 nodes.
To address the neighborhood expansion problem raised in Section \ref{sec:ocgl}, we perform neighborhood sampling, fixing the number of neighbors to 10.
As suggested by Aljundi et~al. \cite{aljundi_online_2019}, multiple passes on the same mini-batch before passing to the next can be beneficial. We therefore considered as an additional tuned hyperparameter whether to perform multiple passes (5) on each mini-batch. We stress how multiple passes on an individual mini-batch are compliant with the constraints of the online setting, as they do not require storing past mini-batches beyond the current one and processing time is multiplied by a constant factor, the number of passes.
As a baseline, we use a \textit{bare} model that is simply fine-tuned on the incoming stream without applying any CL strategy. Additionally, we provide an upper bound in the form of a model that is jointly trained offline on the entire graph.
For replay based methods, we consider different sizes of the memory buffer: 1\%, 2\% and 4\% of the nodes, except for the much larger Products where we use 0.1\%, 0.2\% and 0.4\%, to avoid having an unrealistically large buffer of several tens of thousands of nodes.

\subsection{Hyperparameter selection}

Many works in the CL literature use a learning protocol that is akin to the classic machine learning setting, selecting hyperparameters by performing as many full passes over the task sequence as required by a grid search. This protocol violates stricter definitions of Lifelong Learning, where the stream is observed only once, and is indeed unrealistic for real applications where a model needs to quickly adapt to changes in data distribution. Chaudhry et~al. \cite{chaudhry_efficient_2018} therefore proposed a more sensible hyperparameter selection protocol, which has now been used in several works \cite{xu_graphsail_2020,mai_online_2022,soutifcormerais_comprehensive_2023} and that we use for our experiments. With this protocol, only the first few tasks are used for hyperparameter selection, allowing the model to perform multiple passes, with the same online setting, over them to select the hyperparameters that lead to the best performance on validation nodes. In our case, we considered approximately 20\% of the tasks for this validation, with the exception of Amazon Computer where it was set to 2 as there are only 5 total tasks. Hyperparameters are then selected based on AP on the validation set at this validation boundary (details on the hyperparameters are reported in Appendix \ref{app:hyperparams}).

\section{Results}\label{sec:neighborhood}

\begin{table}[b]
\caption{Results on CoraFull.}
\label{tab:cora}
\begin{center}
\vspace{-1em}
\begin{tabular}{@{}lccc@{}}
\toprule
Method    & AAP\% $\uparrow$                    & AP\% $\uparrow$                             & AF\% $\uparrow$                  \\ \midrule
A-GEM     & $22.43 {\scriptstyle \pm 1.44}$             & $10.65 {\scriptstyle \pm 1.55}$             & $-31.77 {\scriptstyle \pm 4.19}$ \\
ER        & $37.42 {\scriptstyle \pm 0.43}$             & $\underline{29.73 {\scriptstyle \pm 1.08}}$ & $-63.66 {\scriptstyle \pm 1.62}$ \\
EWC       & $30.14 {\scriptstyle \pm 2.57}$             & $8.50 {\scriptstyle \pm 1.45}$              & $-19.64 {\scriptstyle \pm 2.98}$ \\
LwF       & $33.24 {\scriptstyle \pm 0.57}$             & $12.19 {\scriptstyle \pm 1.24}$             & $-39.41 {\scriptstyle \pm 1.24}$ \\
MAS       & $\smash{\dashuline{39.56 {\scriptstyle \pm 2.60}}}$             & $18.04 {\scriptstyle \pm 3.11}$             & $-19.21 {\scriptstyle \pm 2.74}$ \\
PDGNN     & $\underline{49.95 {\scriptstyle \pm 0.29}}$ & $\mathbf{30.78 {\scriptstyle \pm 1.99}}$    & $-61.17 {\scriptstyle \pm 2.15}$ \\
SSM-A-GEM & $22.92 {\scriptstyle \pm 0.49}$             & $11.33 {\scriptstyle \pm 1.84}$             & $-32.29 {\scriptstyle \pm 2.60}$ \\
SSM-ER    & $33.07 {\scriptstyle \pm 1.61}$             & $19.43 {\scriptstyle \pm 1.83}$             & $-22.74 {\scriptstyle \pm 2.80}$ \\
TWP       & $23.28 {\scriptstyle \pm 0.95}$             & $9.97 {\scriptstyle \pm 0.81}$              & $-33.55 {\scriptstyle \pm 2.09}$ \\ \midrule
LINEAR      & $\mathbf{55.70 {\scriptstyle \pm 0.33}}$    & $\smash{\dashuline{27.13 {\scriptstyle \pm 1.29}}}$             & $-67.02 {\scriptstyle \pm 1.30}$ \\ \midrule
bare      & $23.66 {\scriptstyle \pm 0.20}$             & $15.19 {\scriptstyle \pm 2.69}$             & $-67.19 {\scriptstyle \pm 3.85}$ \\
Joint     & -                                           & $67.55 {\scriptstyle \pm 0.05}$             & -                                \\ \bottomrule
\end{tabular}
\vspace{-1em}
\end{center}
\end{table}

\begin{table}[]
\caption{Results on Arxiv.}
\label{tab:arxiv}
\begin{center}
\vspace{-1em}
\begin{tabular}{@{}lccc@{}}
\toprule
Method    & AAP\% $\uparrow$                    & AP\% $\uparrow$                             & AF\% $\uparrow$                  \\ \midrule
A-GEM     & $16.97 {\scriptstyle \pm 0.21}$             & $9.24 {\scriptstyle \pm 0.67}$              & $-80.89 {\scriptstyle \pm 0.57}$ \\
ER        & $36.09 {\scriptstyle \pm 0.19}$             & $20.77 {\scriptstyle \pm 1.38}$             & $-72.98 {\scriptstyle \pm 1.22}$ \\
EWC       & $12.98 {\scriptstyle \pm 0.33}$             & $4.79 {\scriptstyle \pm 0.55}$              & $-56.96 {\scriptstyle \pm 7.95}$ \\
LwF       & $12.96 {\scriptstyle \pm 0.02}$             & $4.61 {\scriptstyle \pm 0.47}$              & $-70.73 {\scriptstyle \pm 1.37}$ \\
MAS       & $13.50 {\scriptstyle \pm 0.43}$             & $6.66 {\scriptstyle \pm 0.68}$              & $-72.52 {\scriptstyle \pm 2.23}$ \\
PDGNN     & $\underline{52.45 {\scriptstyle \pm 0.42}}$ & $\underline{37.83 {\scriptstyle \pm 1.47}}$ & $-50.99 {\scriptstyle \pm 1.69}$ \\
SSM-A-GEM & $23.07 {\scriptstyle \pm 0.47}$             & $15.01 {\scriptstyle \pm 2.04}$             & $-77.04 {\scriptstyle \pm 2.24}$ \\
SSM-ER    & $\smash{\dashuline{44.19 {\scriptstyle \pm 0.58}}}$             & $\smash{\dashuline{24.76 {\scriptstyle \pm 1.14}}}$             & $-66.99 {\scriptstyle \pm 1.16}$ \\
TWP       & $13.92 {\scriptstyle \pm 0.22}$             & $5.20 {\scriptstyle \pm 0.57}$              & $-77.50 {\scriptstyle \pm 1.48}$ \\ \midrule
LINEAR      & $\mathbf{53.15 {\scriptstyle \pm 0.28}}$    & $\mathbf{41.16 {\scriptstyle \pm 1.46}}$    & $-49.78 {\scriptstyle \pm 1.74}$ \\ \midrule
bare      & $12.33 {\scriptstyle \pm 0.02}$             & $4.82 {\scriptstyle \pm 0.13}$              & $-90.07 {\scriptstyle \pm 0.49}$ \\
Joint     & -                                           & $58.58 {\scriptstyle \pm 0.28}$             & -                                \\ \bottomrule
\end{tabular}
\vspace{-1em}
\end{center}
\end{table}

\begin{table}[]
\caption{Results on Reddit.}
\label{tab:reddit}
\begin{center}
\vspace{-1em}
\begin{tabular}{@{}lccc@{}}
\toprule
Method    & AAP\% $\uparrow$                    & AP\% $\uparrow$                             & AF\% $\uparrow$                  \\ \midrule
A-GEM     & $46.02 {\scriptstyle \pm 1.50}$             & $16.76 {\scriptstyle \pm 1.60}$             & $-81.63 {\scriptstyle \pm 1.62}$ \\
ER        & $60.40 {\scriptstyle \pm 0.82}$             & $36.59 {\scriptstyle \pm 3.75}$             & $-61.08 {\scriptstyle \pm 3.67}$ \\
EWC       & $38.86 {\scriptstyle \pm 1.33}$             & $16.05 {\scriptstyle \pm 1.19}$             & $-81.38 {\scriptstyle \pm 1.28}$ \\
LwF       & $33.93 {\scriptstyle \pm 0.04}$             & $14.26 {\scriptstyle \pm 0.21}$             & $-81.18 {\scriptstyle \pm 0.24}$ \\
MAS       & $34.35 {\scriptstyle \pm 1.57}$             & $13.42 {\scriptstyle \pm 1.27}$             & $-80.69 {\scriptstyle \pm 1.63}$ \\
PDGNN     & $\underline{86.66 {\scriptstyle \pm 0.32}}$ & $\mathbf{78.92 {\scriptstyle \pm 0.53}}$    & $-18.64 {\scriptstyle \pm 0.56}$ \\
SSM-A-GEM & $43.61 {\scriptstyle \pm 1.49}$             & $20.25 {\scriptstyle \pm 0.75}$             & $-77.02 {\scriptstyle \pm 0.82}$ \\
SSM-ER    & $\smash{\dashuline{83.85 {\scriptstyle \pm 0.46}}}$             & $\smash{\dashuline{67.95 {\scriptstyle \pm 1.99}}}$             & $-29.22 {\scriptstyle \pm 1.99}$ \\
TWP       & $35.96 {\scriptstyle \pm 1.72}$             & $13.88 {\scriptstyle \pm 1.91}$             & $-83.35 {\scriptstyle \pm 2.21}$ \\ \midrule
LINEAR      & $\mathbf{87.75 {\scriptstyle \pm 0.06}}$    & $\underline{78.38 {\scriptstyle \pm 0.28}}$ & $-17.93 {\scriptstyle \pm 0.36}$ \\ \midrule
bare      & $37.73 {\scriptstyle \pm 1.22}$             & $15.81 {\scriptstyle \pm 3.80}$             & $-81.43 {\scriptstyle \pm 3.85}$ \\
Joint     & -                                           & $90.02 {\scriptstyle \pm 0.12}$             & -                                \\ \bottomrule
\end{tabular}
\vspace{-1em}
\end{center}
\end{table}

\begin{table}[]
\caption{Results on Amazon Computer.}
\label{tab:computer}
\begin{center}
\vspace{-1em}
\begin{tabular}{@{}lccc@{}}
\toprule
Method    & AAP\% $\uparrow$                    & AP\% $\uparrow$                             & AF\% $\uparrow$                  \\ \midrule
A-GEM     & $47.02 {\scriptstyle \pm 0.74}$             & $20.05 {\scriptstyle \pm 0.45}$             & $-77.84 {\scriptstyle \pm 0.68}$ \\
ER        & $56.94 {\scriptstyle \pm 1.00}$             & $45.45 {\scriptstyle \pm 6.07}$             & $-50.68 {\scriptstyle \pm 4.99}$ \\
EWC       & $41.41 {\scriptstyle \pm 0.24}$             & $19.00 {\scriptstyle \pm 0.73}$             & $-77.68 {\scriptstyle \pm 0.35}$ \\
LwF       & $44.49 {\scriptstyle \pm 0.28}$             & $24.63 {\scriptstyle \pm 2.10}$             & $-63.29 {\scriptstyle \pm 3.54}$ \\
MAS       & $45.32 {\scriptstyle \pm 2.23}$             & $21.86 {\scriptstyle \pm 4.19}$             & $-63.52 {\scriptstyle \pm 8.57}$ \\
PDGNN     & $\underline{82.78 {\scriptstyle \pm 0.31}}$ & $\underline{75.17 {\scriptstyle \pm 2.08}}$ & $-19.47 {\scriptstyle \pm 3.42}$ \\
SSM-A-GEM & $50.87 {\scriptstyle \pm 1.28}$             & $32.82 {\scriptstyle \pm 8.61}$             & $-65.03 {\scriptstyle \pm 8.85}$ \\
SSM-ER    & $\smash{\dashuline{70.16 {\scriptstyle \pm 0.99}}}$             & $\smash{\dashuline{56.20 {\scriptstyle \pm 8.32}}}$             & $-40.88 {\scriptstyle \pm 8.44}$ \\
TWP       & $42.51 {\scriptstyle \pm 0.49}$             & $19.34 {\scriptstyle \pm 1.51}$             & $-68.10 {\scriptstyle \pm 6.64}$ \\ \midrule
LINEAR      & $\mathbf{86.53 {\scriptstyle \pm 0.20}}$    & $\mathbf{81.55 {\scriptstyle \pm 0.64}}$    & $-15.18 {\scriptstyle \pm 0.61}$ \\ \midrule
bare      & $42.37 {\scriptstyle \pm 0.47}$             & $18.99 {\scriptstyle \pm 0.69}$             & $-77.65 {\scriptstyle \pm 1.30}$ \\
Joint     & -                                           & $83.07 {\scriptstyle \pm 1.30}$             & -                                \\ \bottomrule
\end{tabular}
\vspace{-1em}
\end{center}
\end{table}

\begin{table}[]
\caption{Results on Products.}
\label{tab:products}
\begin{center}
\vspace{-1em}
\begin{tabular}{@{}lccc@{}}
\toprule
Method    & AAP\% $\uparrow$                            & AP\% $\uparrow$                             & AF\% $\uparrow$                  \\ \midrule
A-GEM     & $49.12 {\scriptstyle \pm 0.34}$             & $32.50 {\scriptstyle \pm 0.60}$             & $-56.44 {\scriptstyle \pm 0.63}$ \\
ER        & $\underline{54.13 {\scriptstyle \pm 0.26}}$ & $\underline{45.37 {\scriptstyle \pm 1.30}}$ & $-50.83 {\scriptstyle \pm 0.91}$ \\
EWC       & $26.85 {\scriptstyle \pm 1.45}$             & $5.13 {\scriptstyle \pm 1.06}$              & $-78.13 {\scriptstyle \pm 8.01}$ \\
LwF       & $23.59 {\scriptstyle \pm 0.05}$             & $4.44 {\scriptstyle \pm 0.05}$              & $-85.24 {\scriptstyle \pm 0.08}$ \\
MAS       & $28.63 {\scriptstyle \pm 1.39}$             & $9.85 {\scriptstyle \pm 2.34}$              & $-65.89 {\scriptstyle \pm 1.94}$ \\
PDGNN     & $\mathbf{62.24 {\scriptstyle \pm 0.70}}$    & $\mathbf{52.41 {\scriptstyle \pm 0.52}}$    & $-42.15 {\scriptstyle \pm 0.66}$ \\
SSM-A-GEM & $48.01 {\scriptstyle \pm 0.19}$             & $31.44 {\scriptstyle \pm 1.81}$             & $-59.72 {\scriptstyle \pm 2.42}$ \\
SSM-ER    & $45.70 {\scriptstyle \pm 0.50}$             & $20.71 {\scriptstyle \pm 1.13}$             & $-75.32 {\scriptstyle \pm 1.18}$ \\
TWP       & $27.22 {\scriptstyle \pm 0.22}$             & $4.32 {\scriptstyle \pm 0.01}$              & $-83.11 {\scriptstyle \pm 0.39}$ \\ \midrule
LINEAR      & $\smash{\dashuline{51.91 {\scriptstyle \pm 0.23}}}$             & $\smash{\dashuline{40.66 {\scriptstyle \pm 0.50}}}$             & $-48.81 {\scriptstyle \pm 0.54}$ \\ \midrule
bare      & $28.15 {\scriptstyle \pm 0.41}$             & $8.56 {\scriptstyle \pm 1.33}$              & $-68.66 {\scriptstyle \pm 1.20}$ \\
Joint     & -                                           & $62.92 {\scriptstyle \pm 0.44}$             & -                                \\ \bottomrule
\end{tabular}
\vspace{-1em}
\end{center}
\end{table}

In this section, we discuss empirical results of the experiments described in Section \ref{sec:setup}. A comparison of AAP, AP and AF for the considered CL methods on the various benchmarks is reported in Tables \ref{tab:cora}-\ref{tab:arxivTI}.
The results for replay-based methods in the tables are those obtained with the larger memory buffer sizes (4\% of nodes, or 0.4\% on Products), while a comparison between different buffer sizes is shown in Figure \ref{fig:buffer_results} and in Appendix \ref{app:plots}. All experiments were repeated five times with different initializations, and the metrics are reported as mean and standard deviation across runs. In the tables, the best results are highlighted in bold, the second-best underlined, and the third with dashed underline.

\subsection{Overall results}

Across benchmarks, the results confirm the difficulty of the OCGL setting, especially for a class-incremental stream: none of the considered strategies approaches the upper bound of AP provided by joint offline training, except for the proposed LINEAR on Amazon Computer. In general, the considered replay methods, specifically SSM-ER, PDGNN and LINEAR achieve higher performance compared to the baseline and regularization methods, and often with a large margin. This can be expected, as rehearsal methods in CL are generally known to achieve most of the state-of-the-art results \cite{van_de_ven_three_2022}; this holds true also for CGL~\cite{zhang_cglb_2022}.
Regularization methods struggle more than the replay strategies, as only in few cases, such as with MAS on CoraFull, they score significantly better than the bare baseline, while their performances are often closer to the lower bound of simple fine-tuning.

\subsection{Anytime evaluation}

\begin{figure*}[htbp]
\centering
\subfloat[CoraFull]{%
  \includegraphics[width=0.49\textwidth]{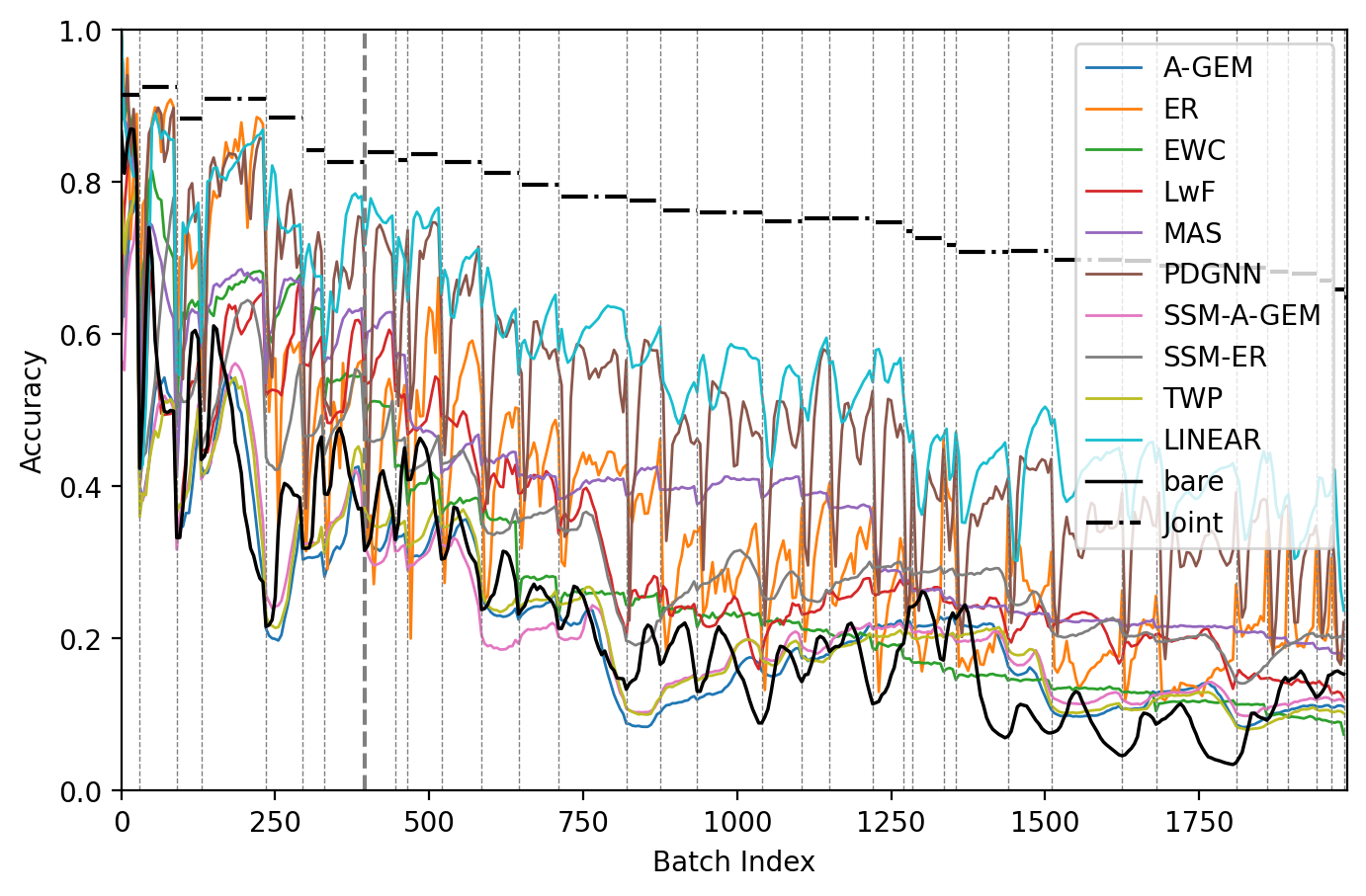}%
  \label{fig:cora_main}}
\hfill
\subfloat[Reddit]{%
  \includegraphics[width=0.49\textwidth]{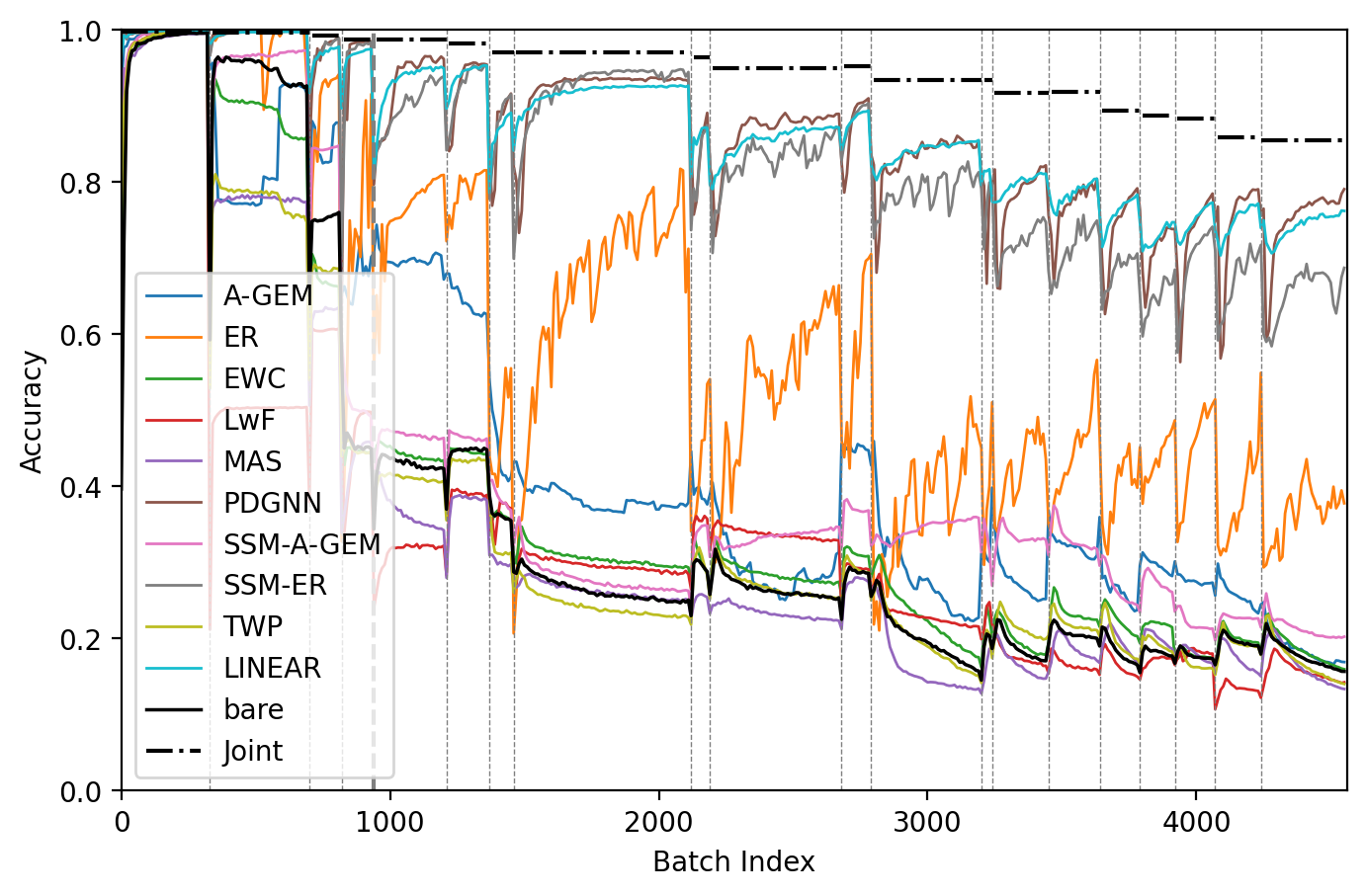}%
  \label{fig:reddit_main}}
\caption{Anytime evaluation on the CoraFull and Reddit datasets: for each method, the line shows average accuracy measured after each training batch. We also indicate with vertical dotted lines the task boundaries, highlighting the validation boundary, and report the upper bound of jointly training up to the current task. We remark that it is natural and expected that accuracy tends to decrease with the batch index, as new classes are introduced and the classification task gets increasingly complex. Similar plots for all datasets are provided in Appendix~\ref{app:plots}.}
\label{fig:anytime_main}
\end{figure*}

Looking at Average Performance gives us an easy way to compare the performance after the entire learning process. However, since in the online setting we expect the model to be ready to make predictions at any time, Average Anytime Performance is a more useful metric of performance over time. Although overall AP and AAP rankings agree, there are cases in which they differ.
Figure \ref{fig:anytime_main} can give us some additional insights into the performance trends through the node stream. For example, on CoraFull we have that LINEAR scores a lower final AP compared to PDGNN, despite having a higher AAP. Yet, looking at Figure \ref{fig:cora_main} we see that this is due to a drop in performance just at the end of the stream, while for the rest of the time LINEAR stays consistently above PDGNN. Or in Figure \ref{fig:reddit_main}, for the Reddit dataset, we see how while LINEAR overall closely follows PDGNN, it suffers from significantly more contained drops at task boundaries. Thus, these cases confirm AAP as more suitable metric than AP, as the behavior of LINEAR is preferable for anytime evaluation.

\subsection{Strategy comparison}

\begin{figure*}[htbp]
\centering
\subfloat[MAS]{%
  \includegraphics[width=0.33\textwidth]{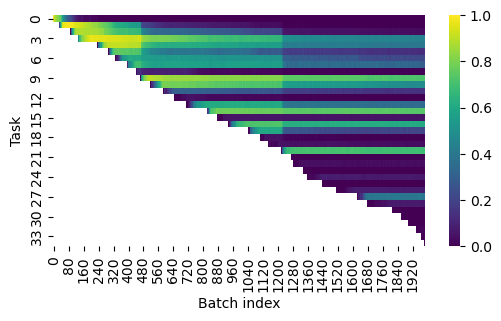}%
  \label{fig:heatmap_cora_mas}}
\hfill
\subfloat[PDGNN]{%
  \includegraphics[width=0.33\textwidth]{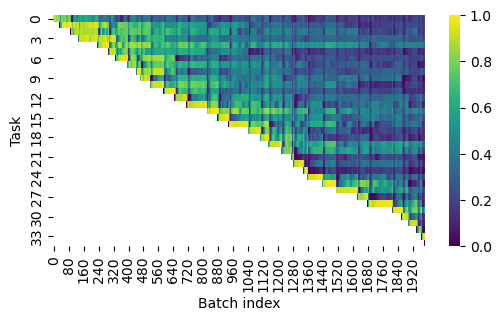}%
  \label{fig:heatmap_cora_pdgnn}}
\hfill
\subfloat[LINEAR]{%
  \includegraphics[width=0.33\textwidth]{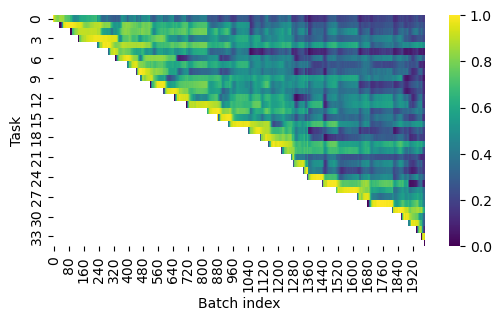}%
  \label{fig:heatmap_cora_malc}}
\caption{Anytime evaluation by task: a breakdown of model performance at the end of each training batch for three selected methods on the CoraFull dataset. Similar plots for all datasets and CL methods are provided in Appendix~\ref{app:plots}.}
\label{fig:heatmaps}
\end{figure*}

To better understand the different results in light of the choice of strategy, we report in Figure \ref{fig:heatmaps} a more detailed breakdown of accuracy by task for three representative methods on the CoraFull dataset. With this comparison we clearly see how MAS (\ref{fig:heatmap_cora_mas}) retains past knowledge for some tasks thanks to its regularization, but struggles to learn new information. Instead, PDGNN (\ref{fig:heatmap_cora_pdgnn}) and LINEAR (\ref{fig:heatmap_cora_malc}) strike a better balance between stability and plasticity. They obtain similar results, but as observed above LINEAR appears more robust to sudden performance drops compared to PDGNN.

\subsection{Time-incremental stream}

Compared to the abrupt and challenging drifts of the class-incremental stream, with a time-incremental stream, determined by the node timestamps, we have a tamer distribution shift which naturally reduces the usefulness of CL. This is reflected in the results of Tables \ref{tab:elliptic}-\ref{tab:arxivTI}, where many methods exhibit positive forgetting, indicating some backward transfer. Although CL offers limited improvement in these conditions, considering AAP and looking at appendix Figure \ref{fig:elliptic_anytime} (Appendix \ref{app:plots}), we see how PDGNN maintains an overall better and more stable performance, only with a slight drop at the end of the stream. Thus, while on these two dataset using CL methods may not be crucial as in cases with more pronounced drifts, using the right strategy can still provide benefits.

\begin{table}[]
\caption{Results on Elliptic.}
\label{tab:elliptic}
\begin{center}
\vspace{-1em}
\begin{tabular}{@{}lccc@{}}
\toprule
Method    & AAP\% $\uparrow$                            & AP\% $\uparrow$                             & AF\% $\uparrow$                  \\ \midrule
A-GEM     & $43.99 {\scriptstyle \pm 0.43}$             & $47.74 {\scriptstyle \pm 1.37}$             & $-3.11 {\scriptstyle \pm 0.79}$  \\
ER        & $\smash{\dashuline{44.81 {\scriptstyle \pm 0.44}}}$             & $44.17 {\scriptstyle \pm 1.19}$             & $-8.18 {\scriptstyle \pm 1.41}$  \\
EWC       & $43.69 {\scriptstyle \pm 0.82}$             & $\underline{51.08 {\scriptstyle \pm 1.10}}$ & $2.06 {\scriptstyle \pm 2.03}$   \\
LwF       & $43.44 {\scriptstyle \pm 0.08}$             & $50.79 {\scriptstyle \pm 1.36}$             & $1.58 {\scriptstyle \pm 0.61}$   \\
MAS       & $43.69 {\scriptstyle \pm 0.82}$             & $\underline{51.08 {\scriptstyle \pm 1.10}}$             & $2.06 {\scriptstyle \pm 2.03}$   \\
PDGNN     & $\mathbf{51.85 {\scriptstyle \pm 0.53}}$    & $49.17 {\scriptstyle \pm 1.01}$             & $-14.49 {\scriptstyle \pm 0.79}$ \\
SSM-A-GEM & $37.39 {\scriptstyle \pm 1.14}$             & $38.40 {\scriptstyle \pm 2.47}$             & $-10.45 {\scriptstyle \pm 2.50}$ \\
SSM-ER    & $31.79 {\scriptstyle \pm 1.38}$             & $28.23 {\scriptstyle \pm 3.18}$             & $-14.18 {\scriptstyle \pm 2.52}$ \\
TWP       & $43.73 {\scriptstyle \pm 0.64}$             & $\mathbf{51.13 {\scriptstyle \pm 1.74}}$    & $2.30 {\scriptstyle \pm 0.99}$   \\ \midrule
LINEAR      & $\underline{46.46 {\scriptstyle \pm 0.25}}$ & $40.57 {\scriptstyle \pm 1.88}$             & $-13.50 {\scriptstyle \pm 2.07}$ \\ \midrule
bare      & $43.76 {\scriptstyle \pm 1.01}$             & $51.28 {\scriptstyle \pm 2.37}$             & $2.37 {\scriptstyle \pm 2.03}$   \\
Joint     & -                                           & $71.97 {\scriptstyle \pm 0.83}$             & -                                \\ \bottomrule
\end{tabular}
\vspace{-1em}
\end{center}
\end{table}

\begin{table}[]
\caption{Results on Arxiv time-incremental.}
\label{tab:arxivTI}
\begin{center}
\vspace{-1em}
\begin{tabular}{@{}lccc@{}}
\toprule
Method    & AAP\% $\uparrow$                            & AP\% $\uparrow$                             & AF\% $\uparrow$                 \\ \midrule
A-GEM     & $59.31 {\scriptstyle \pm 0.09}$             & $64.29 {\scriptstyle \pm 0.51}$             & $3.62 {\scriptstyle \pm 0.36}$  \\
ER        & $60.07 {\scriptstyle \pm 0.10}$             & $64.98 {\scriptstyle \pm 0.64}$             & $3.75 {\scriptstyle \pm 0.21}$  \\
EWC       & $\smash{\dashuline{60.43 {\scriptstyle \pm 0.04}}}$             & $65.72 {\scriptstyle \pm 0.21}$             & $3.76 {\scriptstyle \pm 0.34}$  \\
LwF       & $60.25 {\scriptstyle \pm 0.04}$             & $\mathbf{65.93 {\scriptstyle \pm 0.30}}$    & $3.91 {\scriptstyle \pm 0.38}$  \\
MAS       & $59.38 {\scriptstyle \pm 0.32}$             & $62.16 {\scriptstyle \pm 0.31}$             & $1.99 {\scriptstyle \pm 0.29}$  \\
PDGNN     & $\mathbf{62.92 {\scriptstyle \pm 0.06}}$    & $\smash{\dashuline{65.78 {\scriptstyle \pm 0.15}}}$             & $1.56 {\scriptstyle \pm 0.44}$  \\
SSM-A-GEM & $\underline{60.68 {\scriptstyle \pm 0.04}}$ & $\underline{65.90 {\scriptstyle \pm 0.38}}$ & $4.07 {\scriptstyle \pm 0.27}$  \\
SSM-ER    & $60.06 {\scriptstyle \pm 0.08}$             & $64.80 {\scriptstyle \pm 0.19}$             & $3.31 {\scriptstyle \pm 0.30}$  \\
TWP       & $60.41 {\scriptstyle \pm 0.09}$             & $65.59 {\scriptstyle \pm 0.46}$             & $3.83 {\scriptstyle \pm 0.49}$  \\ \midrule
LINEAR      & $58.01 {\scriptstyle \pm 0.07}$             & $58.08 {\scriptstyle \pm 0.27}$             & $-1.97 {\scriptstyle \pm 0.34}$ \\ \midrule
bare      & $60.37 {\scriptstyle \pm 0.08}$             & $65.52 {\scriptstyle \pm 0.30}$             & $3.68 {\scriptstyle \pm 0.26}$  \\
Joint     & -                                           & $69.72 {\scriptstyle \pm 0.09}$             & -                               \\ \bottomrule
\end{tabular}
\vspace{-1em}
\end{center}
\end{table}

\subsection{Heterophily}

To the best of our knowledge, the issue of heterophily in CGL has been addressed only by Zhao et al. \cite{Zhao_2024_AGALEGraphAwareContinual}. However, their setting ignores inter-task edges, effectively making the graph less heterophilous. Since that requires unrealistic supervisory information on test nodes, we consider the Roman Empire dataset in our setting, showing results in Table \ref{tab:roman}. As can be expected, the results of all considered strategies are much lower than the upper bound. This can be the effect of two factors: first, the considered graph convolution may not be the best to deal with heterophily, and second, since nodes tend to connect with nodes of different classes, many connections are inter-task, and may thus not be observed when training on a certain class. This catastrophic drift in neighborhood composition is of course due to the class-incremental setting, which itself may not be realistic for heterophilous phenomena. While Roman Empire is not a real-world dataset, since it comes from words in a Wikipedia page, our results show that training GCNs in the OCGL setting poses significant challenges due to the drift in neighborhood composition.

\begin{table}[]
\caption{Results on Roman Empire.}
\label{tab:roman}
\begin{center}
\vspace{-1em}
\begin{tabular}{@{}lccc@{}}
\toprule
Method    & AAP\% $\uparrow$                            & AP\% $\uparrow$                             & AF\% $\uparrow$                  \\ \midrule
A-GEM     & $37.85 {\scriptstyle \pm 0.15}$             & $8.97 {\scriptstyle \pm 0.17}$              & $-80.05 {\scriptstyle \pm 0.19}$ \\
ER        & $42.40 {\scriptstyle \pm 0.09}$             & $10.24 {\scriptstyle \pm 0.24}$             & $-77.55 {\scriptstyle \pm 0.39}$ \\
EWC       & $38.61 {\scriptstyle \pm 0.20}$             & $8.85 {\scriptstyle \pm 0.05}$              & $-80.14 {\scriptstyle \pm 0.15}$ \\
LwF       & $38.37 {\scriptstyle \pm 0.00}$             & $8.81 {\scriptstyle \pm 0.01}$              & $-79.98 {\scriptstyle \pm 0.01}$ \\
MAS       & $41.46 {\scriptstyle \pm 0.49}$             & $\smash{\dashuline{11.02 {\scriptstyle \pm 1.36}}}$             & $-41.79 {\scriptstyle \pm 2.72}$ \\
PDGNN     & $\mathbf{46.56 {\scriptstyle \pm 0.33}}$    & $\underline{14.41 {\scriptstyle \pm 0.65}}$ & $-69.46 {\scriptstyle \pm 0.75}$ \\
SSM-A-GEM & $37.77 {\scriptstyle \pm 0.22}$             & $9.09 {\scriptstyle \pm 0.08}$              & $-80.75 {\scriptstyle \pm 0.19}$ \\
SSM-ER    & $\smash{\dashuline{42.82 {\scriptstyle \pm 0.18}}}$             & $9.82 {\scriptstyle \pm 0.53}$              & $-72.30 {\scriptstyle \pm 1.03}$ \\
TWP       & $38.35 {\scriptstyle \pm 0.44}$             & $8.99 {\scriptstyle \pm 0.06}$              & $-79.89 {\scriptstyle \pm 0.16}$ \\ \midrule
LINEAR      & $\underline{45.89 {\scriptstyle \pm 0.19}}$ & $\mathbf{16.63 {\scriptstyle \pm 0.61}}$    & $-67.00 {\scriptstyle \pm 0.51}$ \\ \midrule
bare      & $35.36 {\scriptstyle \pm 0.05}$             & $8.78 {\scriptstyle \pm 0.10}$              & $-80.63 {\scriptstyle \pm 0.22}$ \\
Joint     & -                                           & $39.47 {\scriptstyle \pm 0.33}$             & -                                \\ \bottomrule
\end{tabular}
\vspace{-1em}
\end{center}
\end{table}

\begin{figure*}[h!]
\centering
\subfloat{\includegraphics[width=0.6\textwidth]{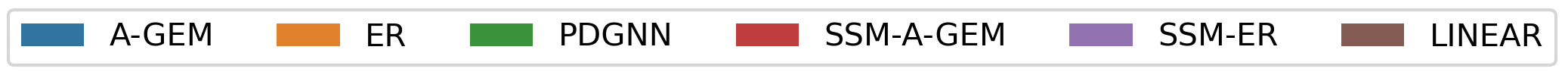}}
\\
\subfloat[CoraFull]{%
  \includegraphics[width=0.245\textwidth]{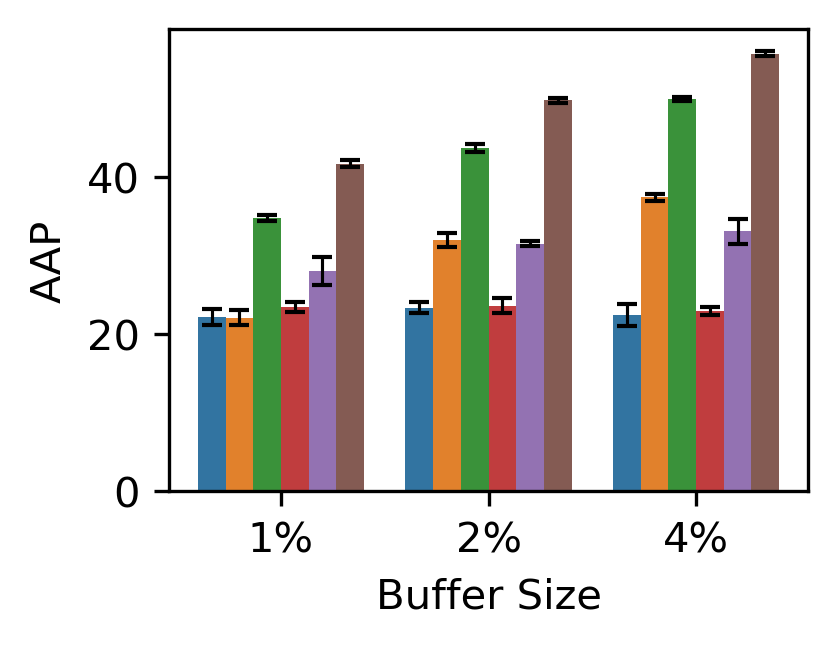}%
  \label{fig:buffer_cora_AAA}}
\hfill
\subfloat[Arxiv]{%
  \includegraphics[width=0.245\textwidth]{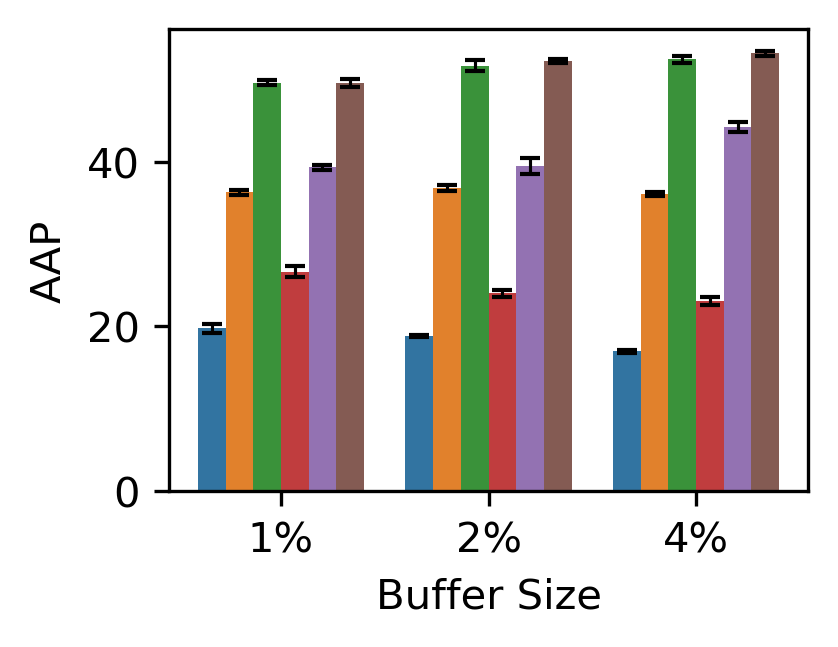}%
  \label{fig:buffer_arxiv_AAA}}
\hfill
\subfloat[Reddit]{%
  \includegraphics[width=0.245\textwidth]{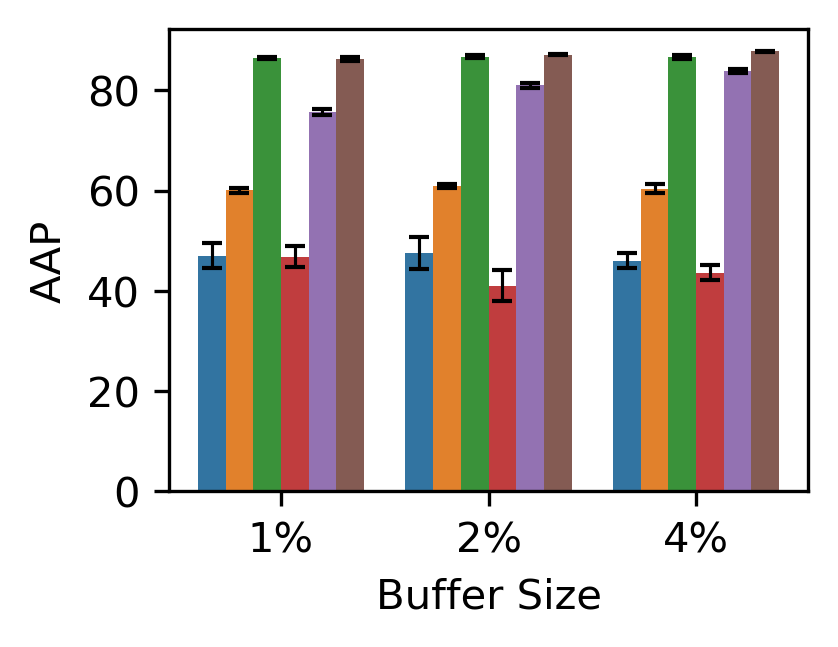}%
  \label{fig:buffer_reddit_AAA}}
\hfill
\subfloat[Amazon Computer]{%
  \includegraphics[width=0.245\textwidth]{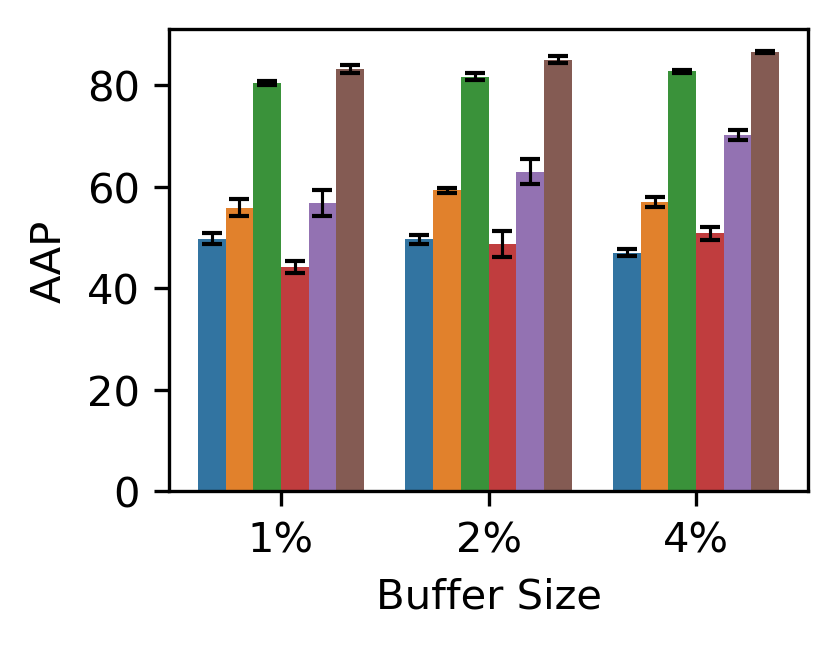}%
  \label{fig:buffer_computer_AAA}}
\\
\subfloat[Roman Empire]{%
  \includegraphics[width=0.245\textwidth]{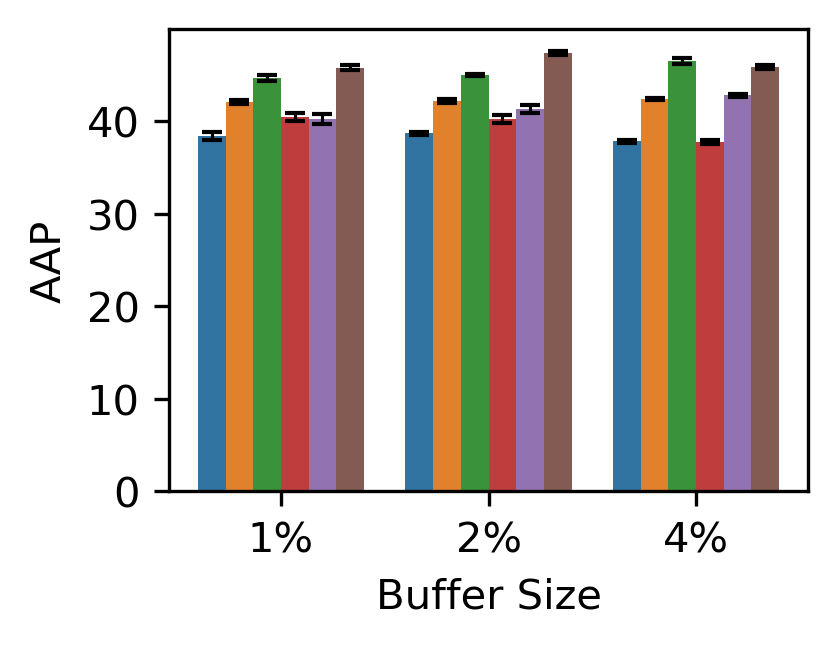}%
  \label{fig:buffer_roman_AAA}}
\hfill
\subfloat[Elliptic]{%
  \includegraphics[width=0.245\textwidth]{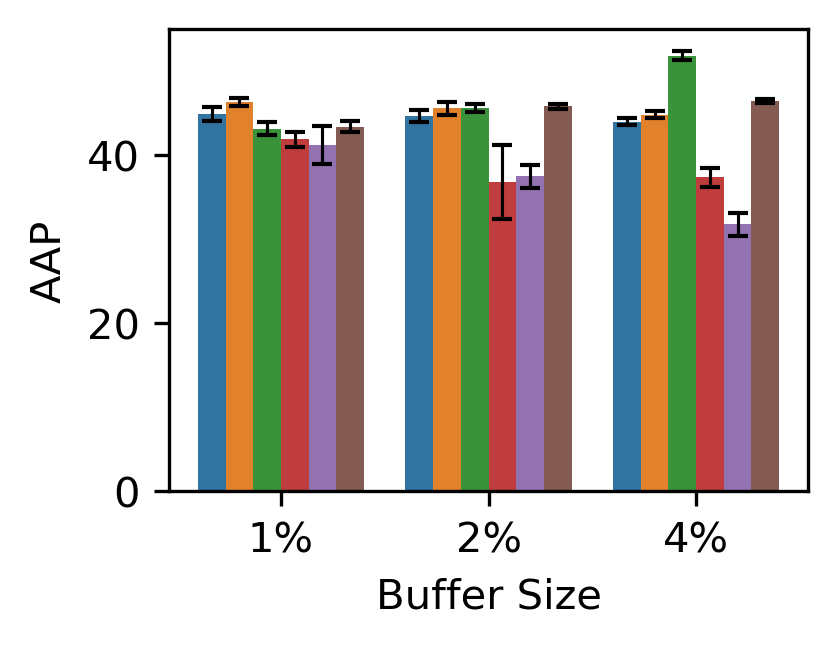}%
  \label{fig:buffer_elliptic_AAA}}
\hfill
\subfloat[Arxiv (time-incr.)]{%
  \includegraphics[width=0.245\textwidth]{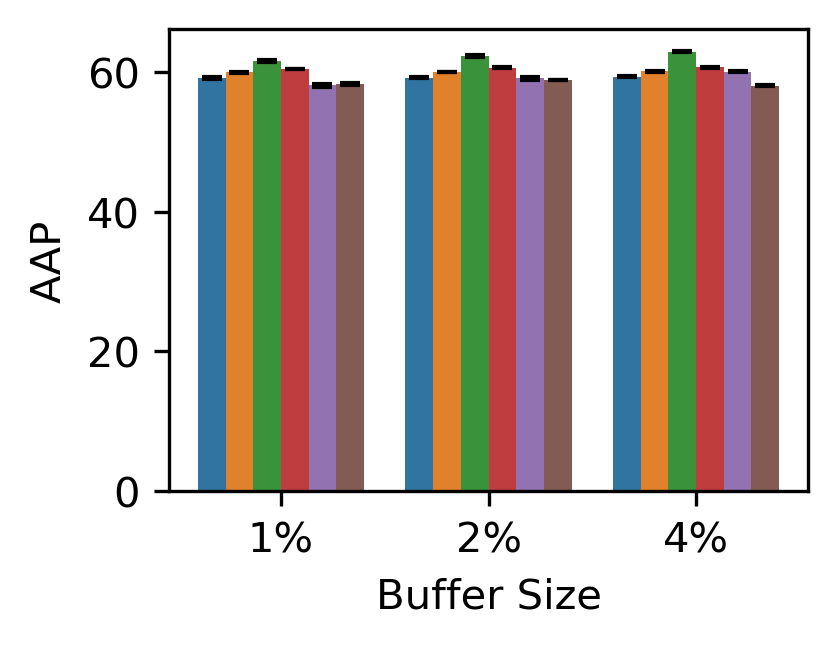}%
  \label{fig:buffer_arxivTI_AAA}}
\hfill
\subfloat[Products]{%
  \includegraphics[width=0.245\textwidth]{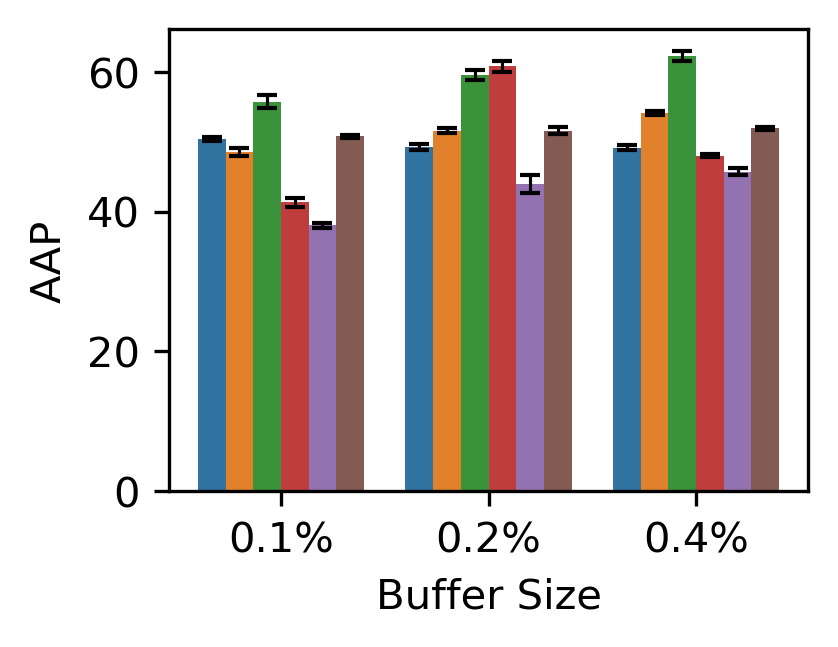}%
  \label{fig:buffer_products_AAA}}
\caption{Results (Average Anytime Performance) with different buffer sizes for replay methods across the datasets.}
\label{fig:buffer_results}
\end{figure*}

\subsection{LINEAR}

Despite its simplicity, LINEAR reliably obtains very good results: it ranks first in four out of eight benchmarks and within the top three in all but one. The only exception is Arxiv with time-incremental stream, where LINEAR underperforms compared to other methods. In this case the distribution drift is limited, as can be inferred by the positive forgetting in Table \ref{tab:arxivTI} and observed in appendix Figures \ref{fig:arxivTI_anytime} and \ref{fig:heatmaps_arxivTI}, so even the bare 2-layer GCN baseline without CL strategy performs well. Instead, LINEAR here suffers from its limited expressive power, since it only aggregates 1-hop information, and with a linear classifier. 
Except for this limit case, we believe that such simplicity is precisely what allows LINEAR to overperform elsewhere: it makes it more robust to forgetting, and faster to adapt. Additionally, its simplicity also results in much lower computational requirements, making it a suitable choice when compute is limited or a high update frequency is required.

\subsection{Design choices and ablation results}

In the previous subsections, we have discussed empirical results obtained on mini-batches with neighborhood sampling and with some fixed design choices. Ablation results with respect to some of these choices are provided in Appendix \ref{app:extended_results}, on the datasets CoraFull, Amazon Computer and Arxiv. We briefly discuss here the impact of these configuration decisions.

\textbf{Memory buffer size.}
In Figure \ref{fig:buffer_results}, we observe the results obtained by the replay methods with different buffer sizes. While overall a larger memory buffer is associated with better performance, this is significantly visible only on CoraFull, as it is a smaller dataset with a large number of classes, and therefore a larger number of samples is required to be representative enough. The need for an increased buffer size is apparent for SSM-based buffers, especially for SSM-ER: as the number of examples is lower than the size of the buffer due to the storage of some neighboring nodes in the form of a sparsified subgraph, we can understand how with lower memory capacity this strategy may not be viable, potentially overfitting to a very small set of examples. PDGNN and LINEAR on the other hand show a consistently high performance even with smaller buffer size, indicating a more efficient use of memory, as it is topology aware.

\textbf{Impact of batch size.} 
Regarding the dimensions of the node batches in the stream, we can compare the results of Tables \ref{tab:cora}, \ref{tab:computer} and \ref{tab:arxiv} with those in Table \ref{tab:larger_batch} of Appendix \ref{app:extended_results}, which are obtained with larger batch sizes (from 10 to 50 and from 50 to 250).
Overall, the two considered sizes have limited differences, as only on CoraFull we observe generally better performances with the larger batch size. On this dataset we also see a significant increase in the performance of regularization methods, with MAS becoming more competitive with the larger batch size. Nonetheless, the fact that very small batch sizes in an online setting can lead to relatively good performance is also encouraging for the future development of OCGL techniques.

\textbf{Impact of sampling.} 
By performing neighborhood sampling to process node mini-batches, we lose some information that might be helpful for the classification task. We expect therefore to trade some performance to remain within the efficiency constraint. Looking at appendix Table \ref{tab:full_neighborhood} (Appendix \ref{app:extended_results}) with results obtained with full neighborhood, we see that this is indeed generally true, although to a very limited extent. There are actually cases in which neighborhood sampling does not significantly affect performance, or even proves beneficial, possibly acting as a regularizer.

\textbf{Sensitivity to hyperparameters.} In our experiments the backbone model architecture, including number of layers and hidden units, is kept fixed. We conducted an ablation study to assess the impact of these choices, evaluating also a backbone GCN with 1 or 3 layers, or with 128 and 512 hidden units, with full results reported in Appendix \ref{app:extended_results}. While changing number of hidden units does not lead to particularly relevant changes, the use of 3 layers of GCN instead of 2 generally reduces performance, in agreement with literature suggesting that deeper networks are more prone to forgetting \cite{Mirzadeh_2022_WideNeuralNetworks}. Results with 1 GCN layer are mixed, but still lower than the ones for LINEAR.

\section{Conclusions}

In this paper, we introduced the Online Continual Graph Learning (OCGL) setting, which bridges the gap between Continual Graph Learning and Online Continual Learning. 
The OCGL formulation establishes a foundation for studying graph-based learning in streaming environments where data arrives sequentially, providing information at the level of nodes. 
A key problem that emerges in this setting is that
of neighborhood expansion, which we addressed with neighborhood sampling as a straightforward solution to bound the computational and memory cost of training on each mini-batch, even as the graph grows through time. 
To support research in this area, we developed a benchmarking environment that adapted seven node-classification datasets to align with the proposed OCGL setting, constructing node streams under both class-incremental and time-incremental learning scenarios.
Our evaluation compares nine suitably adapted methods from the CL literature along with LINEAR, a newly proposed simple and efficient baseline. The results indicate that replay-based methods perform best overall, especially when using tailored strategies to capture topological information in the memory buffer,
while LINEAR achieves highly competitive results despite its simplicity, establishing it as a strong reference point for forthcoming research.
In future works, we plan to further study the neighborhood expansion problem, developing tailored strategies that can ensure computational efficiency while better addressing catastrophic forgetting. 
We also intend to consider more diverse node stream construction and additional tasks such as link prediction.

\bibliographystyle{IEEEtran}

\begin{thebibliography}{10}
\providecommand{\url}[1]{#1}
\csname url@samestyle\endcsname
\providecommand{\newblock}{\relax}
\providecommand{\bibinfo}[2]{#2}
\providecommand{\BIBentrySTDinterwordspacing}{\spaceskip=0pt\relax}
\providecommand{\BIBentryALTinterwordstretchfactor}{4}
\providecommand{\BIBentryALTinterwordspacing}{\spaceskip=\fontdimen2\font plus
\BIBentryALTinterwordstretchfactor\fontdimen3\font minus \fontdimen4\font\relax}
\providecommand{\BIBforeignlanguage}[2]{{%
\expandafter\ifx\csname l@#1\endcsname\relax
\typeout{** WARNING: IEEEtran.bst: No hyphenation pattern has been}%
\typeout{** loaded for the language `#1'. Using the pattern for}%
\typeout{** the default language instead.}%
\else
\language=\csname l@#1\endcsname
\fi
#2}}
\providecommand{\BIBdecl}{\relax}
\BIBdecl

\bibitem{parisi_continual_2019}
G.~I. Parisi, R.~Kemker, J.~L. Part, C.~Kanan, and S.~Wermter, ``Continual lifelong learning with neural networks: {A} review,'' \emph{Neural Networks}, vol. 113, pp. 54--71, May 2019.

\bibitem{de_lange_continual_2022}
M.~De~Lange, R.~Aljundi, M.~Masana, S.~Parisot, X.~Jia, A.~Leonardis, G.~Slabaugh, and T.~Tuytelaars, ``A {Continual} {Learning} {Survey}: {Defying} {Forgetting} in {Classification} {Tasks},'' \emph{IEEE Transactions on Pattern Analysis and Machine Intelligence}, vol.~44, no.~7, pp. 3366--3385, Jul. 2022.

\bibitem{chaudhry_efficient_2018}
A.~Chaudhry, M.~Ranzato, M.~Rohrbach, and M.~Elhoseiny, ``\BIBforeignlanguage{en}{Efficient {Lifelong} {Learning} with {A}-{GEM}},'' in \emph{\BIBforeignlanguage{en}{ICLR}}, Sep. 2018.

\bibitem{mai_online_2022}
Z.~Mai, R.~Li, J.~Jeong, D.~Quispe, H.~Kim, and S.~Sanner, ``\BIBforeignlanguage{en}{Online continual learning in image classification: {An} empirical survey},'' \emph{\BIBforeignlanguage{en}{Neurocomputing}}, vol. 469, pp. 28--51, Jan. 2022.

\bibitem{zliobaite_overview_2016}
I.~Zliobaite, M.~Pechenizkiy, and J.~Gama, ``An {Overview} of {Concept} {Drift} {Applications},'' in \emph{Big {Data} {Analysis}}, ser. Studies in {Big} {Data}, N.~Japkowicz and J.~Stefanowski, Eds.\hskip 1em plus 0.5em minus 0.4em\relax Cham: Springer International Publishing AG, 2016, pp. 91--114.

\bibitem{gunasekara_survey_2023}
N.~Gunasekara, B.~Pfahringer, H.~M. Gomes, and A.~Bifet, ``\BIBforeignlanguage{en}{Survey on {Online} {Streaming} {Continual} {Learning}},'' in \emph{\BIBforeignlanguage{en}{Proceedings of the {Thirty}-{Second} {International} {Joint} {Conference} on {Artificial} {Intelligence}}}.\hskip 1em plus 0.5em minus 0.4em\relax Macau, SAR China: International Joint Conferences on Artificial Intelligence Organization, Aug. 2023, pp. 6628--6637.

\bibitem{yuan_continual_2023}
Q.~Yuan, S.-U. Guan, P.~Ni, T.~Luo, K.~L. Man, P.~Wong, and V.~Chang, ``Continual {Graph} {Learning}: {A} {Survey},'' Jan. 2023, arXiv:2301.12230.

\bibitem{liu_overcoming_2021}
H.~Liu, Y.~Yang, and X.~Wang, ``Overcoming {Catastrophic} {Forgetting} in {Graph} {Neural} {Networks},'' \emph{Proceedings of the AAAI Conference on Artificial Intelligence}, vol.~35, no.~10, pp. 8653--8661, May 2021.

\bibitem{zhou_overcoming_2021}
F.~Zhou and C.~Cao, ``\BIBforeignlanguage{en}{Overcoming {Catastrophic} {Forgetting} in {Graph} {Neural} {Networks} with {Experience} {Replay}},'' \emph{\BIBforeignlanguage{en}{Proceedings of the AAAI Conference on Artificial Intelligence}}, vol.~35, no.~5, pp. 4714--4722, May 2021.

\bibitem{van_de_ven_three_2022}
G.~M. Van De~Ven, T.~Tuytelaars, and A.~S. Tolias, ``\BIBforeignlanguage{en}{Three types of incremental learning},'' \emph{\BIBforeignlanguage{en}{Nature Machine Intelligence}}, vol.~4, no.~12, pp. 1185--1197, Dec. 2022.

\bibitem{kirkpatrick_overcoming_2017}
J.~Kirkpatrick, R.~Pascanu, N.~Rabinowitz, J.~Veness, G.~Desjardins, A.~A. Rusu, K.~Milan, J.~Quan, T.~Ramalho, A.~Grabska-Barwinska, D.~Hassabis, C.~Clopath, D.~Kumaran, and R.~Hadsell, ``\BIBforeignlanguage{en}{Overcoming catastrophic forgetting in neural networks},'' \emph{\BIBforeignlanguage{en}{Proceedings of the National Academy of Sciences}}, vol. 114, no.~13, pp. 3521--3526, Mar. 2017.

\bibitem{rolnick_experience_2019}
D.~Rolnick, A.~Ahuja, J.~Schwarz, T.~Lillicrap, and G.~Wayne, ``Experience {Replay} for {Continual} {Learning},'' in \emph{Advances in {Neural} {Information} {Processing} {Systems}}, vol.~32.\hskip 1em plus 0.5em minus 0.4em\relax Curran Associates, Inc., 2019.

\bibitem{rebuffi_icarl_2017}
S.-A. Rebuffi, A.~Kolesnikov, G.~Sperl, and C.~H. Lampert, ``{iCaRL}: {Incremental} {Classifier} and {Representation} {Learning},'' in \emph{2017 {IEEE} {Conference} on {Computer} {Vision} and {Pattern} {Recognition} ({CVPR})}.\hskip 1em plus 0.5em minus 0.4em\relax Honolulu, HI: IEEE, Jul. 2017, pp. 5533--5542.

\bibitem{lopez-paz_gradient_2017}
D.~Lopez-Paz and M.~A. Ranzato, ``Gradient {Episodic} {Memory} for {Continual} {Learning},'' in \emph{Advances in {Neural} {Information} {Processing} {Systems}}, vol.~30.\hskip 1em plus 0.5em minus 0.4em\relax Curran Associates, Inc., 2017.

\bibitem{aljundi_memory_2018}
R.~Aljundi, F.~Babiloni, M.~Elhoseiny, M.~Rohrbach, and T.~Tuytelaars, ``Memory {Aware} {Synapses}: {Learning} what (not) to forget,'' in \emph{Proceedings of the {European} {Conference} on {Computer} {Vision} ({ECCV})}, 2018, pp. 139--154.

\bibitem{li_learning_2018}
Z.~Li and D.~Hoiem, ``Learning without {Forgetting},'' \emph{IEEE Transactions on Pattern Analysis and Machine Intelligence}, vol.~40, no.~12, pp. 2935--2947, Dec. 2018.

\bibitem{masana_class-incremental_2022}
M.~Masana, X.~Liu, B.~Twardowski, M.~Menta, A.~D. Bagdanov, and J.~Van De~Weijer, ``Class-{Incremental} {Learning}: {Survey} and {Performance} {Evaluation} on {Image} {Classification},'' \emph{IEEE Transactions on Pattern Analysis and Machine Intelligence}, pp. 1--20, 2022.

\bibitem{zenke_continual_2017}
F.~Zenke, B.~Poole, and S.~Ganguli, ``\BIBforeignlanguage{en}{Continual {Learning} {Through} {Synaptic} {Intelligence}},'' in \emph{\BIBforeignlanguage{en}{Proceedings of the 34th {International} {Conference} on {Machine} {Learning}}}.\hskip 1em plus 0.5em minus 0.4em\relax PMLR, Jul. 2017, pp. 3987--3995.

\bibitem{chaudhry_riemannian_2018}
A.~Chaudhry, P.~K. Dokania, T.~Ajanthan, and P.~H.~S. Torr, ``Riemannian {Walk} for {Incremental} {Learning}: {Understanding} {Forgetting} and {Intransigence},'' in \emph{Proceedings of the European Conference on Computer Vision (ECCV)}, 2018, pp. 532--547.

\bibitem{chaudhry_tiny_2019}
A.~Chaudhry, M.~Rohrbach, M.~Elhoseiny, T.~Ajanthan, P.~K. Dokania, P.~H.~S. Torr, and M.~Ranzato, ``On {Tiny} {Episodic} {Memories} in {Continual} {Learning},'' Jun. 2019, arXiv:1902.10486.

\bibitem{fernando_pathnet_2017}
C.~Fernando, D.~Banarse, C.~Blundell, Y.~Zwols, D.~Ha, A.~A. Rusu, A.~Pritzel, and D.~Wierstra, ``{PathNet}: {Evolution} {Channels} {Gradient} {Descent} in {Super} {Neural} {Networks},'' Jan. 2017, arXiv:1701.08734.

\bibitem{masse_alleviating_2018}
N.~Y. Masse, G.~D. Grant, and D.~J. Freedman, ``\BIBforeignlanguage{en}{Alleviating catastrophic forgetting using context-dependent gating and synaptic stabilization},'' \emph{\BIBforeignlanguage{en}{Proceedings of the National Academy of Sciences}}, vol. 115, no.~44, Oct. 2018.

\bibitem{rusu_progressive_2022}
A.~A. Rusu, N.~C. Rabinowitz, G.~Desjardins, H.~Soyer, J.~Kirkpatrick, K.~Kavukcuoglu, R.~Pascanu, and R.~Hadsell, ``Progressive {Neural} {Networks},'' Oct. 2022, arXiv:1606.04671.

\bibitem{soutifcormerais_comprehensive_2023}
A.~Soutif–Cormerais, A.~Carta, A.~Cossu, J.~Hurtado, V.~Lomonaco, J.~Van De~Weijer, and H.~Hemati, ``A {Comprehensive} {Empirical} {Evaluation} on {Online} {Continual} {Learning},'' in \emph{2023 {IEEE}/{CVF} {International} {Conference} on {Computer} {Vision} {Workshops} ({ICCVW})}, Oct. 2023, pp. 3510--3520.

\bibitem{koh_online_2021}
H.~Koh, D.~Kim, J.-W. Ha, and J.~Choi, ``\BIBforeignlanguage{en}{Online {Continual} {Learning} on {Class} {Incremental} {Blurry} {Task} {Configuration} with {Anytime} {Inference}},'' in \emph{\BIBforeignlanguage{en}{International Conference on Learning Representations}}, Oct. 2021.

\bibitem{sperduti_supervised_1997}
A.~Sperduti and A.~Starita, ``Supervised neural networks for the classification of structures,'' \emph{IEEE Transactions on Neural Networks}, vol.~8, no.~3, pp. 714--735, May 1997.

\bibitem{scarselli_graph_2009}
F.~Scarselli, M.~Gori, A.~C. Tsoi, M.~Hagenbuchner, and G.~Monfardini, ``The {Graph} {Neural} {Network} {Model},'' \emph{IEEE Transactions on Neural Networks}, vol.~20, no.~1, pp. 61--80, Jan. 2009.

\bibitem{micheli_neural_2009}
A.~Micheli, ``Neural {Network} for {Graphs}: {A} {Contextual} {Constructive} {Approach},'' \emph{IEEE Transactions on Neural Networks}, vol.~20, no.~3, pp. 498--511, Mar. 2009.

\bibitem{kipf2017semisupervised}
T.~N. Kipf and M.~Welling, ``Semi-supervised classification with graph convolutional networks,'' in \emph{International Conference on Learning Representations}, 2017.

\bibitem{gilmer_neural_2017}
J.~Gilmer, S.~S. Schoenholz, P.~F. Riley, O.~Vinyals, and G.~E. Dahl, ``\BIBforeignlanguage{en}{Neural {Message} {Passing} for {Quantum} {Chemistry}},'' in \emph{\BIBforeignlanguage{en}{Proceedings of the 34th {International} {Conference} on {Machine} {Learning}}}.\hskip 1em plus 0.5em minus 0.4em\relax PMLR, Jul. 2017, pp. 1263--1272, iSSN: 2640-3498.

\bibitem{kazemi2020representation}
S.~M. Kazemi, R.~Goel, K.~Jain, I.~Kobyzev, A.~Sethi, P.~Forsyth, and P.~Poupart, ``Representation learning for dynamic graphs: A survey,'' \emph{Journal of Machine Learning Research}, vol.~21, no.~70, pp. 1--73, 2020.

\bibitem{gravina2024deep}
A.~Gravina and D.~Bacciu, ``Deep learning for dynamic graphs: Models and benchmarks,'' \emph{IEEE Transactions on Neural Networks and Learning Systems}, vol.~35, no.~9, pp. 11\,788--11\,801, 2024.

\bibitem{Liu_2025_Multiview}
M.~Liu, K.~Liang, H.~Yu, L.~Meng, S.~Wang, S.~Zhou, and X.~Liu, ``Multiview temporal graph clustering,'' \emph{IEEE Transactions on Neural Networks and Learning Systems}, vol.~36, no.~10, pp. 18\,383--18\,396, 2025.

\bibitem{Liu_2025_Rethinking}
M.~Liu, Y.~Liu, Q.~Ren, and M.~Han, ``Rethinking multi-level information fusion in temporal graphs: Pre-training then distilling for better embedding,'' \emph{Information Fusion}, vol. 121, p. 103127, 2025.

\bibitem{cini2023graph}
A.~Cini, I.~Marisca, D.~Zambon, and C.~Alippi, ``Graph {{Deep Learning}} for {{Time Series Forecasting}},'' Oct. 2023, arXiv:2310.15978.

\bibitem{jin2024survey}
M.~Jin, H.~Y. Koh, Q.~Wen, D.~Zambon, C.~Alippi, G.~I. Webb, I.~King, and S.~Pan, ``A survey on graph neural networks for time series: Forecasting, classification, imputation, and anomaly detection,'' \emph{IEEE Transactions on Pattern Analysis and Machine Intelligence}, 2024.

\bibitem{cini2023taming}
A.~Cini, I.~Marisca, D.~Zambon, and C.~Alippi, ``Taming {{Local Effects}} in {{Graph-based Spatiotemporal Forecasting}},'' in \emph{Advances in {{Neural Information Processing Systems}}}, vol.~36.\hskip 1em plus 0.5em minus 0.4em\relax Curran Associates, Inc., 2023, pp. 55\,375--55\,393.

\bibitem{huang2023temporal}
S.~Huang, F.~Poursafaei, J.~Danovitch, M.~Fey, W.~Hu, E.~Rossi, J.~Leskovec, M.~Bronstein, G.~Rabusseau, and R.~Rabbany, ``Temporal graph benchmark for machine learning on temporal graphs,'' in \emph{Advances in Neural Information Processing Systems}, A.~Oh, T.~Naumann, A.~Globerson, K.~Saenko, M.~Hardt, and S.~Levine, Eds., vol.~36.\hskip 1em plus 0.5em minus 0.4em\relax Curran Associates, Inc., 2023, pp. 2056--2073.

\bibitem{wang_streaming_2020}
J.~Wang, G.~Song, Y.~Wu, and L.~Wang, ``\BIBforeignlanguage{en}{Streaming {Graph} {Neural} {Networks} via {Continual} {Learning}},'' in \emph{\BIBforeignlanguage{en}{Proceedings of the 29th {ACM} {International} {Conference} on {Information} \& {Knowledge} {Management}}}.\hskip 1em plus 0.5em minus 0.4em\relax Virtual Event Ireland: ACM, Oct. 2020, pp. 1515--1524.

\bibitem{xu_graphsail_2020}
Y.~Xu, Y.~Zhang, W.~Guo, H.~Guo, R.~Tang, and M.~Coates, ``{GraphSAIL}: {Graph} {Structure} {Aware} {Incremental} {Learning} for {Recommender} {Systems},'' in \emph{Proceedings of the 29th {ACM} {International} {Conference} on {Information} \& {Knowledge} {Management}}, ser. {CIKM} '20.\hskip 1em plus 0.5em minus 0.4em\relax New York, NY, USA: Association for Computing Machinery, 2020, pp. 2861--2868.

\bibitem{ijcai2021p498}
X.~Chen, J.~Wang, and K.~Xie, ``Trafficstream: A streaming traffic flow forecasting framework based on graph neural networks and continual learning,'' in \emph{Proceedings of the Thirtieth International Joint Conference on Artificial Intelligence, {IJCAI-21}}.\hskip 1em plus 0.5em minus 0.4em\relax International Joint Conferences on Artificial Intelligence Organization, 8 2021, pp. 3620--3626.

\bibitem{febrinanto_graph_2023}
F.~G. Febrinanto, F.~Xia, K.~Moore, C.~Thapa, and C.~Aggarwal, ``Graph {Lifelong} {Learning}: {A} {Survey},'' \emph{IEEE Computational Intelligence Magazine}, vol.~18, no.~1, pp. 32--51, Feb. 2023.

\bibitem{zhang_continual_2024}
X.~Zhang, D.~Song, and D.~Tao, ``Continual {Learning} on {Graphs}: {Challenges}, {Solutions}, and {Opportunities},'' Feb. 2024, arXiv:2402.11565.

\bibitem{tian_continual_2024}
Z.~Tian, D.~Zhang, and H.-N. Dai, ``Continual {Learning} on {Graphs}: {A} {Survey},'' Feb. 2024, arXiv:2402.06330.

\bibitem{zhang_cglb_2022}
X.~Zhang, D.~Song, and D.~Tao, ``\BIBforeignlanguage{en}{{CGLB}: {Benchmark} {Tasks} for {Continual} {Graph} {Learning}},'' \emph{\BIBforeignlanguage{en}{Advances in Neural Information Processing Systems}}, vol.~35, pp. 13\,006--13\,021, Dec. 2022.

\bibitem{carta_catastrophic_2022}
A.~Carta, A.~Cossu, F.~Errica, and D.~Bacciu, ``Catastrophic {Forgetting} in {Deep} {Graph} {Networks}: {A} {Graph} {Classification} {Benchmark},'' \emph{Frontiers in Artificial Intelligence}, vol.~5, p. 824655, Feb. 2022.

\bibitem{Su_2023_RobustGraphIncremental}
J.~Su, D.~Zou, Z.~Zhang, and C.~Wu, ``Towards robust graph incremental learning on evolving graphs,'' in \emph{Proceedings of the 40th {International} {Conference} on {Machine} {Learning}}, vol. 202, 2023, pp. 32\,728--32\,748.

\bibitem{parisi_online_2020}
G.~I. Parisi and V.~Lomonaco, ``Online {Continual} {Learning} on {Sequences},'' in \emph{Recent trends in learning from data: tutorials from the {INNS} {Big} {Data} and {Deep} {Learning} {Conference} ({INNSBDDL2019})}, ser. Studies in computational intelligence.\hskip 1em plus 0.5em minus 0.4em\relax Cham: Springer, 2020, vol. 896, pp. 197--221, arXiv:2003.09114.

\bibitem{perini_learning_2022}
M.~Perini, G.~Ramponi, P.~Carbone, and V.~Kalavri, ``\BIBforeignlanguage{en}{Learning on streaming graphs with experience replay},'' in \emph{\BIBforeignlanguage{en}{Proceedings of the 37th {ACM}/{SIGAPP} {Symposium} on {Applied} {Computing}}}.\hskip 1em plus 0.5em minus 0.4em\relax Virtual Event: ACM, Apr. 2022, pp. 470--478.

\bibitem{leskovec_graphs_2005}
J.~Leskovec, J.~Kleinberg, and C.~Faloutsos, ``Graphs over time: densification laws, shrinking diameters and possible explanations,'' in \emph{Proceedings of the eleventh {ACM} {SIGKDD} international conference on {Knowledge} discovery in data mining}, 2005, pp. 177--187.

\bibitem{shah_growing_2019}
H.~Shah, S.~Kuma, and H.~Sundaram, ``Growing {Attributed} {Networks} through {Local} {Processes},'' in \emph{The {World} {Wide} {Web} {Conference}}, 2019, pp. 3208--3214.

\bibitem{hamilton_inductive_2017}
W.~Hamilton, Z.~Ying, and J.~Leskovec, ``Inductive {Representation} {Learning} on {Large} {Graphs},'' in \emph{Advances in {Neural} {Information} {Processing} {Systems}}, vol.~30.\hskip 1em plus 0.5em minus 0.4em\relax Curran Associates, Inc., 2017.

\bibitem{chen_fastgcn_2018}
J.~Chen, T.~Ma, and C.~Xiao, ``\BIBforeignlanguage{en}{{FastGCN}: {Fast} {Learning} with {Graph} {Convolutional} {Networks} via {Importance} {Sampling}},'' in \emph{\BIBforeignlanguage{en}{International Conference on Learning Representations}}, Feb. 2018.

\bibitem{chiang_cluster-gcn_2019}
W.-L. Chiang, X.~Liu, S.~Si, Y.~Li, S.~Bengio, and C.-J. Hsieh, ``Cluster-{GCN}: {An} {Efficient} {Algorithm} for {Training} {Deep} and {Large} {Graph} {Convolutional} {Networks},'' in \emph{Proceedings of the 25th {ACM} {SIGKDD} {International} {Conference} on {Knowledge} {Discovery} \& {Data} {Mining}}, Jul. 2019, pp. 257--266.

\bibitem{chen_lifelong_2018}
Z.~Chen and B.~Liu, \emph{\BIBforeignlanguage{en}{Lifelong {Machine} {Learning}}}, ser. Synthesis {Lectures} on {Artificial} {Intelligence} and {Machine} {Learning}.\hskip 1em plus 0.5em minus 0.4em\relax Cham: Springer International Publishing, 2018.

\bibitem{Zhang_2024}
X.~Zhang, D.~Song, Y.~Chen, and D.~Tao, ``Topology-aware {Embedding} {Memory} for {Continual} {Learning} on {Expanding} {Networks},'' in \emph{Proceedings of the 30th {ACM} {SIGKDD} {Conference} on {Knowledge} {Discovery} and {Data} {Mining}}, 2024, pp. 4326--4337.

\bibitem{zhang_sparsified_2022}
X.~Zhang, D.~Song, and D.~Tao, ``Sparsified {Subgraph} {Memory} for {Continual} {Graph} {Representation} {Learning},'' in \emph{2022 {IEEE} {International} {Conference} on {Data} {Mining} ({ICDM})}, 2022, pp. 1335--1340.

\bibitem{liu_cat_2023}
Y.~Liu, R.~Qiu, and Z.~Huang, ``{CaT}: {Balanced} {Continual} {Graph} {Learning} with {Graph} {Condensation},'' in \emph{2023 {IEEE} {International} {Conference} on {Data} {Mining} ({ICDM})}, 2023, pp. 1157--1162.

\bibitem{hoang_universal_2023}
T.~D. Hoang, D.~V. Tung, D.-H. Nguyen, B.-S. Nguyen, H.~H. Nguyen, and H.~Le, ``Universal {Graph} {Continual} {Learning},'' \emph{Transactions on Machine Learning Research}, 2023.

\bibitem{Cai_2022_MultimodalContinualGraph}
J.~Cai, X.~Wang, C.~Guan, Y.~Tang, J.~Xu, B.~Zhong, and W.~Zhu, ``Multimodal {Continual} {Graph} {Learning} with {Neural} {Architecture} {Search},'' in \emph{Proceedings of the {ACM} {Web} {Conference} 2022}, Virtual Event, Lyon France, Apr. 2022, pp. 1292--1300.

\bibitem{prabhu_gdumb_2020}
A.~Prabhu, P.~H.~S. Torr, and P.~K. Dokania, ``\BIBforeignlanguage{en}{{GDumb}: {A} {Simple} {Approach} that {Questions} {Our} {Progress} in {Continual} {Learning}},'' in \emph{\BIBforeignlanguage{en}{Computer {Vision} – {ECCV} 2020}}.\hskip 1em plus 0.5em minus 0.4em\relax Cham: Springer International Publishing, 2020, pp. 524--540.

\bibitem{vitter_random_1985}
J.~S. Vitter, ``Random sampling with a reservoir,'' \emph{ACM Trans. Math. Softw.}, vol.~11, no.~1, pp. 37--57, Mar. 1985.

\bibitem{hinton_distilling_2015}
G.~Hinton, O.~Vinyals, and J.~Dean, ``Distilling the {Knowledge} in a {Neural} {Network},'' Mar. 2015, arXiv:1503.02531.

\bibitem{Wu_2019_SimplifyingGraphConvolutional}
F.~Wu, A.~Souza, T.~Zhang, C.~Fifty, T.~Yu, and K.~Weinberger, ``Simplifying {Graph} {Convolutional} {Networks},'' in \emph{Proceedings of the 36th {International} {Conference} on {Machine} {Learning}}, 2019, pp. 6861--6871.

\bibitem{velickovic_graph_2018}
P.~Veličković, G.~Cucurull, A.~Casanova, A.~Romero, P.~Liò, and Y.~Bengio, ``\BIBforeignlanguage{en}{Graph {Attention} {Networks}},'' in \emph{\BIBforeignlanguage{en}{International Conference on Learning Representations}}, Feb. 2018.

\bibitem{bojchevski_deep_2018}
A.~Bojchevski and S.~Günnemann, ``\BIBforeignlanguage{en}{Deep {Gaussian} {Embedding} of {Graphs}: {Unsupervised} {Inductive} {Learning} via {Ranking}},'' in \emph{\BIBforeignlanguage{en}{International Conference on Learning Representations}}, Feb. 2018.

\bibitem{hu_open_2021}
W.~Hu, M.~Fey, M.~Zitnik, Y.~Dong, H.~Ren, B.~Liu, M.~Catasta, and J.~Leskovec, ``Open {Graph} {Benchmark}: {Datasets} for {Machine} {Learning} on {Graphs},'' Feb. 2021, arXiv:2005.00687.

\bibitem{shchur_pitfalls_2019}
O.~Shchur, M.~Mumme, A.~Bojchevski, and S.~Günnemann, ``Pitfalls of {Graph} {Neural} {Network} {Evaluation},'' Jun. 2019, arXiv:1811.05868.

\bibitem{Platonov_2022_CriticalLookEvaluation}
O.~Platonov, D.~Kuznedelev, M.~Diskin, A.~Babenko, and L.~Prokhorenkova, ``A critical look at the evaluation of {GNNs} under heterophily: {Are} we really making progress?'' in \emph{International Conference on Learning Representations}, 2022.

\bibitem{Weber_2019_AntiMoneyLaunderingBitcoin}
M.~Weber, G.~Domeniconi, J.~Chen, D.~K.~I. Weidele, C.~Bellei, T.~Robinson, and C.~E. Leiserson, ``Anti-{Money} {Laundering} in {Bitcoin}: {Experimenting} with {Graph} {Convolutional} {Networks} for {Financial} {Forensics},'' 2019, arXiv:1908.02591.

\bibitem{caccia_new_2021}
L.~Caccia, R.~Aljundi, N.~Asadi, T.~Tuytelaars, J.~Pineau, and E.~Belilovsky, ``\BIBforeignlanguage{en}{New {Insights} on {Reducing} {Abrupt} {Representation} {Change} in {Online} {Continual} {Learning}},'' in \emph{\BIBforeignlanguage{en}{International Conference on Learning Representations}}, Oct. 2021.

\bibitem{kingma_adam_2017}
D.~P. Kingma and J.~Ba, ``Adam: {A} {Method} for {Stochastic} {Optimization},'' Jan. 2017, arXiv:1412.6980.

\bibitem{aljundi_online_2019}
R.~Aljundi, E.~Belilovsky, T.~Tuytelaars, L.~Charlin, M.~Caccia, M.~Lin, and L.~Page-Caccia, ``Online {Continual} {Learning} with {Maximal} {Interfered} {Retrieval},'' in \emph{Advances in {Neural} {Information} {Processing} {Systems}}, vol.~32.\hskip 1em plus 0.5em minus 0.4em\relax Curran Associates, Inc., 2019.

\bibitem{Zhao_2024_AGALEGraphAwareContinual}
T.~Zhao, A.~Hanjalic, and M.~Khosla, ``{AGALE}: {A} {Graph}-{Aware} {Continual} {Learning} {Evaluation} {Framework},'' \emph{Transactions on Machine Learning Research}, 2024.

\bibitem{Mirzadeh_2022_WideNeuralNetworks}
S.~I. Mirzadeh, A.~Chaudhry, D.~Yin, H.~Hu, R.~Pascanu, D.~Gorur, and M.~Farajtabar, ``Wide {Neural} {Networks} {Forget} {Less} {Catastrophically},'' in \emph{Proceedings of the 39th {International} {Conference} on {Machine} {Learning}}, Jun. 2022, pp. 15\,699--15\,717.

\end{thebibliography}

\onecolumn
\appendices

\section{Datasets}\label{app:dataset}

In the experiments for this paper, we used seven node-level classification graph datasets. The CoraFull dataset \cite{bojchevski_deep_2018} is a citation network where nodes represent research papers and edges indicate citations between them, with labels based on paper topics. Arxiv \cite{hu_open_2021} is a larger citation network derived from arXiv papers in the Computer Science category. The Reddit dataset \cite{hamilton_inductive_2017} consists of posts from different communities of the Reddit platform, where nodes represent posts, and edges connect posts commented on by the same user, forming a large interaction graph. Amazon Computer \cite{shchur_pitfalls_2019} and Products \cite{hu_open_2021} are co-purchase networks, where nodes are products and edges indicate frequently co-purchased items on Amazon. In contrast to previous dataset which are homophilous, Roman Empire \cite{Platonov_2022_CriticalLookEvaluation} is an heterophilous dataset obtained from the homonymous Wikipedia page, where nodes represent words connected by syntactic relationships or adjacency in the sentence. Finally, Elliptic \cite{Weber_2019_AntiMoneyLaunderingBitcoin} is a Bitcoin transaction dataset, consisting of transactions connected by flows of Bitcoins. Only some of the nodes are labeled, either as \textit{licit} (42,019 nodes) or \textit{illicit} transaction (4,545 nodes).
Summary statistics for the four graphs are reported in Table \ref{tab:datasets}.

\begin{table*}[htbp]
\caption{Dataset statistics.}
\label{tab:datasets}
\begin{center}
\setlength{\tabcolsep}{8pt}
\begin{tabular}{@{}l|ccccccc@{}}
\toprule
Dataset    & CoraFull & Arxiv     & Reddit      & Amazon Computer & Products   & Roman Empire & Elliptic \\ \midrule
\# nodes   & 19,793   & 169,343   & 227,853     & 13,752          & 2,449,028  & 22,662       & 203,769  \\
\# edges   & 130,622  & 1,166,243 & 114,615,892 & 491,722         & 61,859,036 & 32,927       & 234,355  \\
\# classes & 70       & 40        & 40          & 10              & 46         & 18           & 2        \\ \bottomrule
\end{tabular}
\end{center}
\end{table*}

\section{Metrics} 
\label{app:Metrics}
Thanks to the construction of the node stream starting from the class-incremental setting (or the artificial definition of tasks for Elliptic), we can exploit two widely used metrics in CL: \textit{Average Performance (AP)} and \textit{Average Forgetting (AF)} \cite{lopez-paz_gradient_2017}. The most comprehensive metric for CL, from which AP and AF are derived, is the performance matrix $\mM \in \R^{T \times T}$, where $T$ is the number of tasks and $M_{i,j}$ is the test classification performance on task $j$ after the model has observed task $i$. AP is then defined as $\text{AP} = \frac{1}{T} \sum_{i=1}^T M_{T,i}$, and average forgetting as $\text{AF} = \frac{1}{T-1} \sum_{i=1}^{T-1} M_{T,i} - M_{i,i}$. AP serves as a single value to quantify the performance of the model after having observed the entire sequence of tasks, or stream in our case. AF measures the performance degradation (forgetting), that occurs from when a task was just observed to the end of training. As a performance metric, we use accuracy for all datasets except Elliptic. Since the latter is highly unbalanced, on it we use the F1 score of the \textit{illicit} class.

More importantly, to assess the performance of the model throughout the node stream, we also perform anytime evaluation, meaning that we evaluate the model on validation nodes after training on each mini-batch \cite{koh_online_2021}. This allows us to capture the performance at any point in time, and observe also graphically how the model reacts to changes in data distribution. We measure this with \textit{Average Anytime Performance (AAP)} \cite{caccia_new_2021}, which is a generalization of average incremental accuracy for the online setting. Indicating with AP$_t$ the average performance after training on batch $t$, and having $n$ batches in total, $\text{AAP} = \frac{1}{n}\sum_{t=1}^n \text{AP}_t$. This can be interpreted as an Area Under the Curve accuracy score \cite{koh_online_2021}.

\section{Hyperparameters}\label{app:hyperparams}

A standard grid search was performed to select training hyperparameters for the models used in all experiments. We detail here the specific search space for each of the methods used in our comparisons. Two hyperparameters are common for all techniques: the learning rate, selected in the set $\{0.01, 0.001, 0.0001, 0.00001\}$, and the number of passes on each batch before passing to the next one, chosen between 1 and 5. No weight decay or dropout were used. Method specific hyperparameters are reported in Table \ref{tab:hyperparameters}, and specific details can be found in the original papers. In particular, the hyperparameters of regularization methods regulate the strength of the regularization. For LwF a new hyperparameter has been introduced to adapt to the online setting: the number of batches after which to update the teacher model. For replay based methods we consider the proportion of memories to use with respect to each training batch, as using the entire buffer like in CGL is unfeasible in an online setting. The neighbors budget for SSM (both SSM-ER and SSM-A-GEM) refers to the number of the first- and second-hop neighbors to preserve in the memory budget, which influences the effective number of example in the buffer.

\begin{table*}[]
\vspace{-1em}
\caption{Method specific hyperparameters.}
\label{tab:hyperparameters}
\begin{center}
\resizebox{0.8\textwidth}{!}{
\begin{tabular}{@{}lc@{}}
\toprule
Method & Hyperparameter candidates                                                                        \\ \midrule
A-GEM, ER, PDGNN, LINEAR     &  memory\_proportion: \{1,2,3\}                                             \\
SSM     &  neighbhors\_budget: \{(5,5), (10,10), (25,25)\}                                             \\
EWC, MAS    & lambda: \{$10^0, 10^2, 10^4, 10^6, 10^8, 10^{10}$\}                                                \\
LwF    & lambda\_dist: \{0.1,1,10\}; T: \{0.2,2,20\}, update\_every: \{1, 10, 100\}                       \\
TWP    & lambda\_l: \{$10^2, 10^4, 10^6$\}; lambda\_t: \{$10^2, 10^4, 10^6$\}; beta: \{0.001, 0.01, 0.1\} \\ \bottomrule
\end{tabular}
}
\end{center}
\end{table*}

\section{Ablation Study}\label{app:extended_results}

In the main experiments of the papers, some hyperparameter and design choices, such as batch sizes and the structure of the GCN model, are kept fixed. In this section we provide results to assess the impact of these choices, for three of the considered benchmarks: CoraFull, Amazon Computer and Arxiv (in class-incremental stream setting). For ease of reference, we report here their aggregated results from the tables in the main paper into Table \ref{tab:cora_amazon_arxiv}.

\begin{table*}[h]
\caption{Results on CoraFull, Amazon Computer and Arxiv in the main setting.}
\label{tab:cora_amazon_arxiv}
\begin{center}
\setlength{\tabcolsep}{3pt}
\begin{tabular}{@{}l|ccc|ccc|ccc@{}}
\toprule
          & \multicolumn{3}{c|}{CoraFull}                                                                                                & \multicolumn{3}{c|}{Amazon Computer}                                                                                         & \multicolumn{3}{c}{Arxiv}                                                                                                    \\
Method    & AAP\% $\uparrow$                            & AP\% $\uparrow$                             & AF\% $\uparrow$                  & AAP\% $\uparrow$                            & AP\% $\uparrow$                             & AF\% $\uparrow$                  & AAP\% $\uparrow$                            & AP\% $\uparrow$                             & AF\% $\uparrow$                  \\ \midrule
A-GEM     & $22.43 {\scriptstyle \pm 1.44}$             & $10.65 {\scriptstyle \pm 1.55}$             & $-31.77 {\scriptstyle \pm 4.19}$ & $47.02 {\scriptstyle \pm 0.74}$             & $20.05 {\scriptstyle \pm 0.45}$             & $-77.84 {\scriptstyle \pm 0.68}$ & $16.97 {\scriptstyle \pm 0.21}$             & $9.24 {\scriptstyle \pm 0.67}$              & $-80.89 {\scriptstyle \pm 0.57}$ \\
ER        & $37.42 {\scriptstyle \pm 0.43}$             & $\underline{29.73 {\scriptstyle \pm 1.08}}$ & $-63.66 {\scriptstyle \pm 1.62}$ & $56.94 {\scriptstyle \pm 1.00}$             & $45.45 {\scriptstyle \pm 6.07}$             & $-50.68 {\scriptstyle \pm 4.99}$ & $36.09 {\scriptstyle \pm 0.19}$             & $20.77 {\scriptstyle \pm 1.38}$             & $-72.98 {\scriptstyle \pm 1.22}$ \\
EWC       & $30.14 {\scriptstyle \pm 2.57}$             & $8.50 {\scriptstyle \pm 1.45}$              & $-19.64 {\scriptstyle \pm 2.98}$ & $41.41 {\scriptstyle \pm 0.24}$             & $19.00 {\scriptstyle \pm 0.73}$             & $-77.68 {\scriptstyle \pm 0.35}$ & $12.98 {\scriptstyle \pm 0.33}$             & $4.79 {\scriptstyle \pm 0.55}$              & $-56.96 {\scriptstyle \pm 7.95}$ \\
LwF       & $33.24 {\scriptstyle \pm 0.57}$             & $12.19 {\scriptstyle \pm 1.24}$             & $-39.41 {\scriptstyle \pm 1.24}$ & $44.49 {\scriptstyle \pm 0.28}$             & $24.63 {\scriptstyle \pm 2.10}$             & $-63.29 {\scriptstyle \pm 3.54}$ & $12.96 {\scriptstyle \pm 0.02}$             & $4.61 {\scriptstyle \pm 0.47}$              & $-70.73 {\scriptstyle \pm 1.37}$ \\
MAS       & $\smash{\dashuline{39.56 {\scriptstyle \pm 2.60}}}$             & $18.04 {\scriptstyle \pm 3.11}$             & $-19.21 {\scriptstyle \pm 2.74}$ & $45.32 {\scriptstyle \pm 2.23}$             & $21.86 {\scriptstyle \pm 4.19}$             & $-63.52 {\scriptstyle \pm 8.57}$ & $13.50 {\scriptstyle \pm 0.43}$             & $6.66 {\scriptstyle \pm 0.68}$              & $-72.52 {\scriptstyle \pm 2.23}$ \\
PDGNN     & $\underline{49.95 {\scriptstyle \pm 0.29}}$ & $\mathbf{30.78 {\scriptstyle \pm 1.99}}$    & $-61.17 {\scriptstyle \pm 2.15}$ & $\underline{82.78 {\scriptstyle \pm 0.31}}$ & $\underline{75.17 {\scriptstyle \pm 2.08}}$ & $-19.47 {\scriptstyle \pm 3.42}$ & $\underline{52.45 {\scriptstyle \pm 0.42}}$ & $\underline{37.83 {\scriptstyle \pm 1.47}}$ & $-50.99 {\scriptstyle \pm 1.69}$ \\
SSM-A-GEM & $22.92 {\scriptstyle \pm 0.49}$             & $11.33 {\scriptstyle \pm 1.84}$             & $-32.29 {\scriptstyle \pm 2.60}$ & $50.87 {\scriptstyle \pm 1.28}$             & $32.82 {\scriptstyle \pm 8.61}$             & $-65.03 {\scriptstyle \pm 8.85}$ & $23.07 {\scriptstyle \pm 0.47}$             & $15.01 {\scriptstyle \pm 2.04}$             & $-77.04 {\scriptstyle \pm 2.24}$ \\
SSM-ER    & $33.07 {\scriptstyle \pm 1.61}$             & $19.43 {\scriptstyle \pm 1.83}$             & $-22.74 {\scriptstyle \pm 2.80}$ & $\smash{\dashuline{70.16 {\scriptstyle \pm 0.99}}}$             & $\smash{\dashuline{56.20 {\scriptstyle \pm 8.32}}}$             & $-40.88 {\scriptstyle \pm 8.44}$ & $\smash{\dashuline{44.19 {\scriptstyle \pm 0.58}}}$             & $\smash{\dashuline{24.76 {\scriptstyle \pm 1.14}}}$             & $-66.99 {\scriptstyle \pm 1.16}$ \\
TWP       & $23.28 {\scriptstyle \pm 0.95}$             & $9.97 {\scriptstyle \pm 0.81}$              & $-33.55 {\scriptstyle \pm 2.09}$ & $42.51 {\scriptstyle \pm 0.49}$             & $19.34 {\scriptstyle \pm 1.51}$             & $-68.10 {\scriptstyle \pm 6.64}$ & $13.92 {\scriptstyle \pm 0.22}$             & $5.20 {\scriptstyle \pm 0.57}$              & $-77.50 {\scriptstyle \pm 1.48}$ \\ \midrule
LINEAR      & $\mathbf{55.70 {\scriptstyle \pm 0.33}}$    & $\smash{\dashuline{27.13 {\scriptstyle \pm 1.29}}}$             & $-67.02 {\scriptstyle \pm 1.30}$ & $\mathbf{86.53 {\scriptstyle \pm 0.20}}$    & $\mathbf{81.55 {\scriptstyle \pm 0.64}}$    & $-15.18 {\scriptstyle \pm 0.61}$ & $\mathbf{53.15 {\scriptstyle \pm 0.28}}$    & $\mathbf{41.16 {\scriptstyle \pm 1.46}}$    & $-49.78 {\scriptstyle \pm 1.74}$ \\ \midrule
bare      & $23.66 {\scriptstyle \pm 0.20}$             & $15.19 {\scriptstyle \pm 2.69}$             & $-67.19 {\scriptstyle \pm 3.85}$ & $42.37 {\scriptstyle \pm 0.47}$             & $18.99 {\scriptstyle \pm 0.69}$             & $-77.65 {\scriptstyle \pm 1.30}$ & $12.33 {\scriptstyle \pm 0.02}$             & $4.82 {\scriptstyle \pm 0.13}$              & $-90.07 {\scriptstyle \pm 0.49}$ \\ \bottomrule
\end{tabular}
\end{center}
\end{table*}

\subsection{Full results with larger batch size}

In Table \ref{tab:larger_batch}, we provide extended results with the same setting configurations as the one explored in Section \ref{sec:neighborhood} of the main paper, yet with larger mini-batch size. Compared to size 10 for CoraFull and Amazon Computer, and 50 for Arxiv, here we use mini-batches of 50 and 250 nodes respectively. While on CoraFull and Arxiv the differences compared to the main setting are limited, on CoraFull we observe generally better performances (except for ER), with significantly higher results for regularization methods EWC and MAS.

\begin{table*}[h]
\caption{Results with larger batch size.}
\label{tab:larger_batch}
\begin{center}
\setlength{\tabcolsep}{3pt}
\begin{tabular}{@{}l|ccc|ccc|ccc@{}}
\toprule
          & \multicolumn{3}{c|}{CoraFull}                                                                                                & \multicolumn{3}{c|}{Amazon Computer}                                                                                         & \multicolumn{3}{c}{Arxiv}                                                                                                    \\
Method    & AAP\% $\uparrow$                            & AP\% $\uparrow$                             & AF\% $\uparrow$                  & AAP\% $\uparrow$                            & AP\% $\uparrow$                             & AF\% $\uparrow$                  & AAP\% $\uparrow$                            & AP\% $\uparrow$                             & AF\% $\uparrow$                  \\ \midrule
A-GEM     & $33.95 {\scriptstyle \pm 0.46}$             & $30.40 {\scriptstyle \pm 4.91}$             & $-57.50 {\scriptstyle \pm 5.50}$ & $48.57 {\scriptstyle \pm 0.95}$             & $19.48 {\scriptstyle \pm 0.30}$             & $-78.37 {\scriptstyle \pm 0.49}$ & $19.63 {\scriptstyle \pm 0.41}$             & $10.62 {\scriptstyle \pm 0.87}$             & $-79.40 {\scriptstyle \pm 1.32}$ \\
ER        & $35.70 {\scriptstyle \pm 0.47}$             & $17.37 {\scriptstyle \pm 3.22}$             & $-74.63 {\scriptstyle \pm 3.33}$ & $54.13 {\scriptstyle \pm 0.58}$             & $33.57 {\scriptstyle \pm 2.55}$             & $-62.78 {\scriptstyle \pm 1.84}$ & $29.94 {\scriptstyle \pm 0.43}$             & $14.23 {\scriptstyle \pm 1.15}$             & $-78.78 {\scriptstyle \pm 1.41}$ \\
EWC       & $45.54 {\scriptstyle \pm 2.72}$             & $29.26 {\scriptstyle \pm 2.91}$             & $-18.69 {\scriptstyle \pm 1.50}$ & $43.25 {\scriptstyle \pm 0.69}$             & $20.58 {\scriptstyle \pm 1.83}$             & $-77.77 {\scriptstyle \pm 2.11}$ & $15.39 {\scriptstyle \pm 0.73}$             & $6.55 {\scriptstyle \pm 1.56}$              & $-59.72 {\scriptstyle \pm 4.36}$ \\
LwF       & $28.51 {\scriptstyle \pm 0.75}$             & $12.25 {\scriptstyle \pm 1.23}$             & $-24.17 {\scriptstyle \pm 2.75}$ & $37.66 {\scriptstyle \pm 0.13}$             & $17.36 {\scriptstyle \pm 0.14}$             & $-59.07 {\scriptstyle \pm 0.93}$ & $12.31 {\scriptstyle \pm 0.04}$             & $5.87 {\scriptstyle \pm 1.16}$              & $-64.97 {\scriptstyle \pm 2.43}$ \\
MAS       & $\smash{\dashuline{52.13 {\scriptstyle \pm 1.18}}}$             & $\smash{\dashuline{32.42 {\scriptstyle \pm 1.82}}}$             & $-20.99 {\scriptstyle \pm 1.70}$ & $45.04 {\scriptstyle \pm 2.60}$             & $22.21 {\scriptstyle \pm 5.15}$             & $-43.78 {\scriptstyle \pm 2.37}$ & $12.50 {\scriptstyle \pm 0.21}$             & $4.95 {\scriptstyle \pm 0.15}$              & $-70.67 {\scriptstyle \pm 2.19}$ \\
PDGNN     & $\underline{52.35 {\scriptstyle \pm 0.31}}$ & $\mathbf{33.80 {\scriptstyle \pm 0.94}}$    & $-57.90 {\scriptstyle \pm 1.24}$ & $\underline{82.96 {\scriptstyle \pm 0.34}}$ & $\underline{76.08 {\scriptstyle \pm 1.23}}$ & $-21.00 {\scriptstyle \pm 1.21}$ & $\mathbf{53.70 {\scriptstyle \pm 0.27}}$    & $\mathbf{38.37 {\scriptstyle \pm 2.05}}$    & $-52.71 {\scriptstyle \pm 2.21}$ \\
SSM-A-GEM & $26.76 {\scriptstyle \pm 1.08}$             & $15.86 {\scriptstyle \pm 1.24}$             & $-19.24 {\scriptstyle \pm 1.93}$ & $44.97 {\scriptstyle \pm 1.71}$             & $22.79 {\scriptstyle \pm 2.58}$             & $-74.83 {\scriptstyle \pm 2.59}$ & $21.05 {\scriptstyle \pm 0.65}$             & $14.38 {\scriptstyle \pm 3.07}$             & $-62.42 {\scriptstyle \pm 3.54}$ \\
SSM-ER    & $36.35 {\scriptstyle \pm 1.00}$             & $24.55 {\scriptstyle \pm 1.55}$             & $-8.81 {\scriptstyle \pm 1.29}$  & $\smash{\dashuline{70.88 {\scriptstyle \pm 1.04}}}$             & $\smash{\dashuline{53.25 {\scriptstyle \pm 5.02}}}$             & $-43.91 {\scriptstyle \pm 5.03}$ & $\smash{\dashuline{41.97 {\scriptstyle \pm 0.39}}}$             & $\smash{\dashuline{21.41 {\scriptstyle \pm 1.97}}}$             & $-67.12 {\scriptstyle \pm 1.91}$ \\
TWP       & $25.39 {\scriptstyle \pm 0.32}$             & $14.10 {\scriptstyle \pm 1.67}$             & $-19.13 {\scriptstyle \pm 2.05}$ & $40.01 {\scriptstyle \pm 2.27}$             & $18.08 {\scriptstyle \pm 0.50}$             & $-63.25 {\scriptstyle \pm 7.84}$ & $16.89 {\scriptstyle \pm 1.29}$             & $5.92 {\scriptstyle \pm 2.31}$              & $-52.42 {\scriptstyle \pm 5.79}$ \\ \midrule
LINEAR      & $\mathbf{57.35 {\scriptstyle \pm 0.33}}$    & $\underline{33.28 {\scriptstyle \pm 1.30}}$ & $-61.05 {\scriptstyle \pm 1.26}$ & $\mathbf{86.59 {\scriptstyle \pm 0.34}}$    & $\mathbf{78.94 {\scriptstyle \pm 0.99}}$    & $-16.97 {\scriptstyle \pm 1.27}$ & $\underline{48.46 {\scriptstyle \pm 0.21}}$ & $\underline{38.25 {\scriptstyle \pm 0.73}}$ & $-53.01 {\scriptstyle \pm 0.76}$ \\ \midrule
bare      & $24.76 {\scriptstyle \pm 0.61}$             & $13.05 {\scriptstyle \pm 1.77}$             & $-18.58 {\scriptstyle \pm 3.05}$ & $42.60 {\scriptstyle \pm 0.28}$             & $19.52 {\scriptstyle \pm 0.21}$             & $-79.09 {\scriptstyle \pm 0.12}$ & $12.02 {\scriptstyle \pm 0.15}$             & $4.41 {\scriptstyle \pm 1.04}$              & $-78.64 {\scriptstyle \pm 4.12}$ \\ \bottomrule
\end{tabular}
\end{center}
\end{table*}

\subsection{Results with full neighborhood}

As one of the main requirements for the OCGL setting is to keep a constant computational footprint, in the experiments for the paper we adopt neighborhood sampling to tame the neighborhood expansion issue. Yet, we expect that this may come as a tradeoff with performance, since some information is thus discarded. To assess this, we report in Table \ref{tab:full_neighborhood} results obtained with mini-batches containing the full neighborhoods of newly presented nodes.
We observe nonetheless how these results are in some cases lower than those with neighborhood sampling, suggesting that it could potentially act as a regularizer, and confirming the goodness of this efficient choice for the OCGL setting.

\begin{table*}[h]
\caption{Results with full neighborhood.}
\label{tab:full_neighborhood}
\begin{center}
\setlength{\tabcolsep}{3pt}
\vspace{-1em}
\begin{tabular}{@{}l|ccc|ccc|ccc@{}}
\toprule
          & \multicolumn{3}{c|}{CoraFull}                                                                                                & \multicolumn{3}{c|}{Amazon Computer}                                                                                          & \multicolumn{3}{c}{Arxiv}                                                                                                    \\
Method    & AAP\% $\uparrow$                            & AP\% $\uparrow$                             & AF\% $\uparrow$                  & AAP\% $\uparrow$                            & AP\% $\uparrow$                             & AF\% $\uparrow$                   & AAP\% $\uparrow$                            & AP\% $\uparrow$                             & AF\% $\uparrow$                  \\ \midrule
A-GEM     & $26.36 {\scriptstyle \pm 0.24}$             & $13.40 {\scriptstyle \pm 1.86}$             & $-79.25 {\scriptstyle \pm 1.94}$ & $46.35 {\scriptstyle \pm 1.18}$             & $21.19 {\scriptstyle \pm 1.34}$             & $-74.06 {\scriptstyle \pm 1.42}$  & $27.11 {\scriptstyle \pm 0.19}$             & $11.27 {\scriptstyle \pm 0.78}$             & $-73.55 {\scriptstyle \pm 1.68}$ \\
ER        & $35.03 {\scriptstyle \pm 0.24}$             & $26.28 {\scriptstyle \pm 2.02}$             & $-65.67 {\scriptstyle \pm 1.81}$ & $46.65 {\scriptstyle \pm 0.54}$             & $26.26 {\scriptstyle \pm 5.16}$             & $-70.58 {\scriptstyle \pm 5.82}$  & $26.75 {\scriptstyle \pm 0.55}$             & $10.90 {\scriptstyle \pm 1.76}$             & $-80.39 {\scriptstyle \pm 1.89}$ \\
EWC       & $\smash{\dashuline{45.87 {\scriptstyle \pm 0.77}}}$             & $\underline{28.43 {\scriptstyle \pm 4.52}}$ & $-23.58 {\scriptstyle \pm 5.01}$ & $38.23 {\scriptstyle \pm 3.32}$             & $11.53 {\scriptstyle \pm 10.16}$            & $-60.26 {\scriptstyle \pm 13.98}$ & $13.10 {\scriptstyle \pm 0.36}$             & $1.67 {\scriptstyle \pm 1.65}$              & $-43.20 {\scriptstyle \pm 5.65}$ \\
LwF       & $36.88 {\scriptstyle \pm 0.16}$             & $16.10 {\scriptstyle \pm 0.61}$             & $-49.12 {\scriptstyle \pm 1.10}$ & $50.18 {\scriptstyle \pm 1.46}$             & $22.74 {\scriptstyle \pm 5.74}$             & $-60.41 {\scriptstyle \pm 7.67}$  & $12.97 {\scriptstyle \pm 0.29}$             & $4.71 {\scriptstyle \pm 0.33}$              & $-53.32 {\scriptstyle \pm 2.56}$ \\
MAS       & $37.82 {\scriptstyle \pm 0.28}$             & $19.51 {\scriptstyle \pm 0.79}$             & $-14.84 {\scriptstyle \pm 1.28}$ & $44.76 {\scriptstyle \pm 2.30}$             & $19.77 {\scriptstyle \pm 2.40}$             & $-17.73 {\scriptstyle \pm 13.98}$ & $12.78 {\scriptstyle \pm 1.77}$             & $4.54 {\scriptstyle \pm 0.97}$              & $-11.86 {\scriptstyle \pm 2.55}$ \\
PDGNN     & $\underline{49.02 {\scriptstyle \pm 0.24}}$ & $\mathbf{29.52 {\scriptstyle \pm 3.55}}$    & $-62.26 {\scriptstyle \pm 3.31}$ & $\underline{79.21 {\scriptstyle \pm 0.80}}$ & $\underline{73.31 {\scriptstyle \pm 2.33}}$ & $-22.03 {\scriptstyle \pm 2.53}$  & $\underline{52.00 {\scriptstyle \pm 0.24}}$ & $\underline{36.36 {\scriptstyle \pm 1.48}}$ & $-52.91 {\scriptstyle \pm 1.57}$ \\
SSM-A-GEM & $26.24 {\scriptstyle \pm 0.58}$             & $16.94 {\scriptstyle \pm 2.02}$             & $-40.75 {\scriptstyle \pm 3.02}$ & $41.42 {\scriptstyle \pm 1.37}$             & $20.94 {\scriptstyle \pm 0.97}$             & $-72.79 {\scriptstyle \pm 1.11}$  & $27.68 {\scriptstyle \pm 0.63}$             & $12.80 {\scriptstyle \pm 1.89}$             & $-63.39 {\scriptstyle \pm 1.92}$ \\
SSM-ER    & $39.45 {\scriptstyle \pm 0.84}$             & $23.80 {\scriptstyle \pm 4.24}$             & $-60.89 {\scriptstyle \pm 3.92}$ & $\smash{\dashuline{56.55 {\scriptstyle \pm 0.68}}}$             & $\smash{\dashuline{38.26 {\scriptstyle \pm 5.86}}}$             & $-53.88 {\scriptstyle \pm 5.18}$  & $\smash{\dashuline{39.65 {\scriptstyle \pm 0.63}}}$             & $\smash{\dashuline{19.25 {\scriptstyle \pm 1.75}}}$             & $-68.69 {\scriptstyle \pm 1.93}$ \\
TWP       & $25.88 {\scriptstyle \pm 0.79}$             & $15.99 {\scriptstyle \pm 1.64}$             & $-41.83 {\scriptstyle \pm 1.02}$ & $41.92 {\scriptstyle \pm 1.31}$             & $19.67 {\scriptstyle \pm 3.28}$             & $-71.26 {\scriptstyle \pm 4.44}$  & $14.42 {\scriptstyle \pm 1.54}$             & $2.27 {\scriptstyle \pm 1.50}$              & $-43.55 {\scriptstyle \pm 6.06}$ \\ \midrule
LINEAR      & $\mathbf{56.03 {\scriptstyle \pm 0.45}}$    & $\smash{\dashuline{27.87 {\scriptstyle \pm 0.43}}}$             & $-66.15 {\scriptstyle \pm 0.32}$ & $\mathbf{86.13 {\scriptstyle \pm 0.28}}$    & $\mathbf{79.12 {\scriptstyle \pm 0.73}}$    & $-18.13 {\scriptstyle \pm 0.81}$  & $\mathbf{53.09 {\scriptstyle \pm 0.27}}$    & $\mathbf{41.13 {\scriptstyle \pm 1.37}}$    & $-49.72 {\scriptstyle \pm 1.58}$ \\ \midrule
bare      & $25.58 {\scriptstyle \pm 0.23}$             & $17.87 {\scriptstyle \pm 2.51}$             & $-68.51 {\scriptstyle \pm 3.25}$ & $39.88 {\scriptstyle \pm 0.24}$             & $18.40 {\scriptstyle \pm 0.84}$             & $-77.21 {\scriptstyle \pm 1.21}$  & $12.46 {\scriptstyle \pm 0.19}$             & $4.81 {\scriptstyle \pm 0.10}$              & $-85.16 {\scriptstyle \pm 0.69}$ \\ \bottomrule
\end{tabular}
\end{center}
\end{table*}

\subsection{Number of hidden units}

Still maintaining a 2-layer GCN as in the main experiments, we changed here the number of hidden units, from the original 256 to alternatively 512 and 128. We report the results in Tables \ref{tab:512units} and \ref{tab:128units}. LINEAR is not reported as it is a linear model.

\begin{table*}[h]
\caption{Results with 512 units.}
\label{tab:512units}
\begin{center}
\setlength{\tabcolsep}{3pt}
\vspace{-1em}
\begin{tabular}{@{}l|ccc|ccc|ccc@{}}
\toprule
          & \multicolumn{3}{c|}{CoraFull}                                                                                                & \multicolumn{3}{c|}{Amazon Computer}                                                                                         & \multicolumn{3}{c}{Arxiv}                                                                                                    \\
Method    & AAP\% $\uparrow$                            & AP\% $\uparrow$                             & AF\% $\uparrow$                  & AAP\% $\uparrow$                            & AP\% $\uparrow$                             & AF\% $\uparrow$                  & AAP\% $\uparrow$                            & AP\% $\uparrow$                             & AF\% $\uparrow$                  \\ \midrule
A-GEM     & $24.72 {\scriptstyle \pm 0.78}$             & $15.00 {\scriptstyle \pm 1.18}$             & $-36.05 {\scriptstyle \pm 0.84}$ & $48.30 {\scriptstyle \pm 1.87}$             & $24.21 {\scriptstyle \pm 3.09}$             & $-72.58 {\scriptstyle \pm 3.90}$ & $16.85 {\scriptstyle \pm 0.40}$             & $8.79 {\scriptstyle \pm 0.38}$              & $-82.65 {\scriptstyle \pm 1.17}$ \\
ER        & $\underline{38.59 {\scriptstyle \pm 0.30}}$ & $\underline{24.99 {\scriptstyle \pm 2.28}}$ & $-67.83 {\scriptstyle \pm 2.48}$ & $\smash{\dashuline{60.50 {\scriptstyle \pm 0.24}}}$             & $\smash{\dashuline{44.83 {\scriptstyle \pm 3.81}}}$             & $-52.53 {\scriptstyle \pm 4.25}$ & $\smash{\dashuline{38.25 {\scriptstyle \pm 0.28}}}$             & $\smash{\dashuline{21.38 {\scriptstyle \pm 3.06}}}$             & $-71.99 {\scriptstyle \pm 2.89}$ \\
EWC       & $34.88 {\scriptstyle \pm 2.57}$             & $14.94 {\scriptstyle \pm 1.35}$             & $-5.25 {\scriptstyle \pm 0.77}$  & $38.20 {\scriptstyle \pm 0.91}$             & $2.65 {\scriptstyle \pm 0.63}$              & $-58.26 {\scriptstyle \pm 4.62}$ & $11.49 {\scriptstyle \pm 0.19}$             & $4.63 {\scriptstyle \pm 0.56}$              & $-70.49 {\scriptstyle \pm 2.24}$ \\
LwF       & $31.58 {\scriptstyle \pm 0.25}$             & $9.57 {\scriptstyle \pm 0.33}$              & $-58.31 {\scriptstyle \pm 0.47}$ & $47.17 {\scriptstyle \pm 2.19}$             & $29.69 {\scriptstyle \pm 0.98}$             & $-67.52 {\scriptstyle \pm 1.33}$ & $12.37 {\scriptstyle \pm 0.09}$             & $4.95 {\scriptstyle \pm 0.09}$              & $-73.40 {\scriptstyle \pm 0.34}$ \\
MAS       & $33.95 {\scriptstyle \pm 0.97}$             & $12.90 {\scriptstyle \pm 1.78}$             & $-20.01 {\scriptstyle \pm 1.90}$ & $44.30 {\scriptstyle \pm 0.45}$             & $21.90 {\scriptstyle \pm 2.55}$             & $-68.13 {\scriptstyle \pm 4.20}$ & $13.82 {\scriptstyle \pm 0.21}$             & $5.86 {\scriptstyle \pm 0.91}$              & $-75.34 {\scriptstyle \pm 2.56}$ \\
PDGNN     & $\mathbf{50.46 {\scriptstyle \pm 0.47}}$    & $\mathbf{29.89 {\scriptstyle \pm 3.19}}$    & $-62.09 {\scriptstyle \pm 2.90}$ & $\mathbf{79.64 {\scriptstyle \pm 0.47}}$    & $\mathbf{73.08 {\scriptstyle \pm 4.29}}$    & $-22.16 {\scriptstyle \pm 5.28}$ & $\mathbf{53.16 {\scriptstyle \pm 0.40}}$    & $\mathbf{38.49 {\scriptstyle \pm 1.64}}$    & $-51.35 {\scriptstyle \pm 1.85}$ \\
SSM-A-GEM & $24.95 {\scriptstyle \pm 0.48}$             & $15.81 {\scriptstyle \pm 1.38}$             & $-34.92 {\scriptstyle \pm 2.11}$ & $49.54 {\scriptstyle \pm 1.10}$             & $35.00 {\scriptstyle \pm 1.98}$             & $-63.44 {\scriptstyle \pm 2.05}$ & $23.38 {\scriptstyle \pm 0.36}$             & $13.05 {\scriptstyle \pm 0.67}$             & $-80.68 {\scriptstyle \pm 0.94}$ \\
SSM-ER    & $\smash{\dashuline{34.93 {\scriptstyle \pm 1.58}}}$             & $\smash{\dashuline{21.51 {\scriptstyle \pm 0.99}}}$             & $-29.06 {\scriptstyle \pm 1.89}$ & $\underline{67.42 {\scriptstyle \pm 0.44}}$ & $\underline{45.86 {\scriptstyle \pm 5.50}}$ & $-50.39 {\scriptstyle \pm 5.86}$ & $\underline{42.74 {\scriptstyle \pm 0.50}}$ & $\underline{22.26 {\scriptstyle \pm 1.54}}$ & $-69.66 {\scriptstyle \pm 1.79}$ \\
TWP       & $24.15 {\scriptstyle \pm 0.64}$             & $13.64 {\scriptstyle \pm 0.43}$             & $-36.86 {\scriptstyle \pm 0.84}$ & $39.46 {\scriptstyle \pm 2.31}$             & $8.71 {\scriptstyle \pm 8.14}$              & $-68.13 {\scriptstyle \pm 9.75}$ & $13.43 {\scriptstyle \pm 0.18}$             & $4.98 {\scriptstyle \pm 0.06}$              & $-85.66 {\scriptstyle \pm 1.72}$ \\ \midrule
bare      & $24.20 {\scriptstyle \pm 0.65}$             & $14.62 {\scriptstyle \pm 0.79}$             & $-35.69 {\scriptstyle \pm 1.57}$ & $42.47 {\scriptstyle \pm 0.30}$             & $19.36 {\scriptstyle \pm 0.22}$             & $-78.96 {\scriptstyle \pm 0.30}$ & $12.32 {\scriptstyle \pm 0.02}$             & $4.91 {\scriptstyle \pm 0.04}$              & $-90.78 {\scriptstyle \pm 0.13}$ \\ \bottomrule
\end{tabular}
\end{center}
\end{table*}

\begin{table*}[h]
\caption{Results with 128 units.}
\label{tab:128units}
\begin{center}
\setlength{\tabcolsep}{3pt}
\vspace{-1em}
\begin{tabular}{@{}l|ccc|ccc|ccc@{}}
\toprule
          & \multicolumn{3}{c|}{CoraFull}                                                                                                & \multicolumn{3}{c|}{Amazon Computer}                                                                                          & \multicolumn{3}{c}{Arxiv}                                                                                                    \\
Method    & AAP\% $\uparrow$                            & AP\% $\uparrow$                             & AF\% $\uparrow$                  & AAP\% $\uparrow$                            & AP\% $\uparrow$                             & AF\% $\uparrow$                   & AAP\% $\uparrow$                            & AP\% $\uparrow$                             & AF\% $\uparrow$                  \\ \midrule
A-GEM     & $19.85 {\scriptstyle \pm 1.17}$             & $8.98 {\scriptstyle \pm 2.48}$              & $-28.35 {\scriptstyle \pm 2.22}$ & $47.31 {\scriptstyle \pm 0.29}$             & $20.38 {\scriptstyle \pm 1.03}$             & $-77.88 {\scriptstyle \pm 0.91}$  & $22.82 {\scriptstyle \pm 0.14}$             & $10.80 {\scriptstyle \pm 1.05}$             & $-82.72 {\scriptstyle \pm 1.30}$ \\
ER        & $36.35 {\scriptstyle \pm 0.55}$             & $\mathbf{31.77 {\scriptstyle \pm 1.25}}$    & $-61.08 {\scriptstyle \pm 1.81}$ & $\smash{\dashuline{57.79 {\scriptstyle \pm 1.11}}}$             & $\smash{\dashuline{36.46 {\scriptstyle \pm 9.77}}}$             & $-60.30 {\scriptstyle \pm 9.58}$  & $\smash{\dashuline{35.01 {\scriptstyle \pm 0.41}}}$             & $\smash{\dashuline{20.21 {\scriptstyle \pm 2.09}}}$             & $-74.30 {\scriptstyle \pm 2.13}$ \\
EWC       & $26.58 {\scriptstyle \pm 3.39}$             & $8.14 {\scriptstyle \pm 1.47}$              & $-13.74 {\scriptstyle \pm 1.78}$ & $42.39 {\scriptstyle \pm 0.63}$             & $18.72 {\scriptstyle \pm 0.89}$             & $-78.76 {\scriptstyle \pm 0.11}$  & $12.96 {\scriptstyle \pm 0.48}$             & $3.48 {\scriptstyle \pm 1.64}$              & $-49.56 {\scriptstyle \pm 4.69}$ \\
LwF       & $33.29 {\scriptstyle \pm 0.23}$             & $15.22 {\scriptstyle \pm 0.33}$             & $-45.98 {\scriptstyle \pm 0.30}$ & $42.12 {\scriptstyle \pm 2.23}$             & $20.34 {\scriptstyle \pm 5.81}$             & $-65.18 {\scriptstyle \pm 10.27}$ & $11.51 {\scriptstyle \pm 0.00}$             & $5.03 {\scriptstyle \pm 0.05}$              & $-79.30 {\scriptstyle \pm 0.10}$ \\
MAS       & $\underline{40.46 {\scriptstyle \pm 2.74}}$ & $19.77 {\scriptstyle \pm 2.76}$             & $-18.00 {\scriptstyle \pm 2.51}$ & $42.91 {\scriptstyle \pm 0.88}$             & $21.90 {\scriptstyle \pm 4.02}$             & $-54.26 {\scriptstyle \pm 16.25}$ & $13.02 {\scriptstyle \pm 0.54}$             & $3.79 {\scriptstyle \pm 0.82}$              & $-53.02 {\scriptstyle \pm 3.20}$ \\
PDGNN     & $\mathbf{49.10 {\scriptstyle \pm 0.30}}$    & $\underline{24.56 {\scriptstyle \pm 1.67}}$ & $-66.53 {\scriptstyle \pm 1.57}$ & $\mathbf{83.03 {\scriptstyle \pm 0.18}}$    & $\mathbf{75.57 {\scriptstyle \pm 1.47}}$    & $-20.89 {\scriptstyle \pm 1.57}$  & $\mathbf{53.39 {\scriptstyle \pm 0.49}}$    & $\mathbf{37.10 {\scriptstyle \pm 1.69}}$    & $-52.93 {\scriptstyle \pm 1.83}$ \\
SSM-A-GEM & $20.53 {\scriptstyle \pm 0.76}$             & $9.91 {\scriptstyle \pm 0.81}$              & $-28.65 {\scriptstyle \pm 3.44}$ & $48.18 {\scriptstyle \pm 1.54}$             & $29.20 {\scriptstyle \pm 5.51}$             & $-68.86 {\scriptstyle \pm 5.53}$  & $22.50 {\scriptstyle \pm 0.44}$             & $14.50 {\scriptstyle \pm 1.97}$             & $-74.79 {\scriptstyle \pm 1.61}$ \\
SSM-ER    & $\smash{\dashuline{37.11 {\scriptstyle \pm 0.61}}}$             & $\smash{\dashuline{21.51 {\scriptstyle \pm 3.18}}}$             & $-53.99 {\scriptstyle \pm 4.23}$ & $\underline{68.81 {\scriptstyle \pm 1.27}}$ & $\underline{53.13 {\scriptstyle \pm 6.06}}$ & $-43.63 {\scriptstyle \pm 6.56}$  & $\underline{44.20 {\scriptstyle \pm 0.35}}$ & $\underline{25.53 {\scriptstyle \pm 1.27}}$ & $-66.01 {\scriptstyle \pm 1.66}$ \\
TWP       & $21.47 {\scriptstyle \pm 1.06}$             & $10.16 {\scriptstyle \pm 3.02}$             & $-28.70 {\scriptstyle \pm 4.96}$ & $42.29 {\scriptstyle \pm 1.95}$             & $18.27 {\scriptstyle \pm 0.38}$             & $-66.04 {\scriptstyle \pm 12.34}$ & $14.01 {\scriptstyle \pm 0.63}$             & $5.27 {\scriptstyle \pm 2.92}$              & $-54.97 {\scriptstyle \pm 4.33}$ \\ \midrule
bare      & $21.72 {\scriptstyle \pm 1.14}$             & $11.02 {\scriptstyle \pm 2.16}$             & $-27.49 {\scriptstyle \pm 1.92}$ & $41.23 {\scriptstyle \pm 1.24}$             & $17.78 {\scriptstyle \pm 0.24}$             & $-76.93 {\scriptstyle \pm 2.94}$  & $12.31 {\scriptstyle \pm 0.03}$             & $4.86 {\scriptstyle \pm 0.07}$              & $-90.20 {\scriptstyle \pm 0.27}$ \\ \bottomrule
\end{tabular}
\end{center}
\end{table*}

\subsection{Number of GCN layers}

We considered here a different number of GCN layers compared to the 2 our main results, 3 and 1 specifically, keeping the number of hidden units fixed to 256. Results are reported in Tables \ref{tab:3layers} and \ref{tab:1layer}. LINEAR is not reported as it is a linear model. We observe overall lower results with 3 layers, in accordance with literature suggesting that deeper networks are more prone to forgetting \cite{Mirzadeh_2022_WideNeuralNetworks}. With only 1 layers results are mixed, with also some cases of improved performance over the 2-layer network.

\begin{table*}[h]
\caption{Results with 3 layers.}
\label{tab:3layers}
\begin{center}
\setlength{\tabcolsep}{3pt}
\begin{tabular}{@{}l|ccc|ccc|ccc@{}}
\toprule
          & \multicolumn{3}{c|}{CoraFull}                                                                                                & \multicolumn{3}{c|}{Amazon Computer}                                                                                          & \multicolumn{3}{c}{Arxiv}                                                                                                     \\
Method    & AAP\% $\uparrow$                            & AP\% $\uparrow$                             & AF\% $\uparrow$                  & AAP\% $\uparrow$                            & AP\% $\uparrow$                             & AF\% $\uparrow$                   & AAP\% $\uparrow$                            & AP\% $\uparrow$                             & AF\% $\uparrow$                   \\ \midrule
A-GEM     & $16.98 {\scriptstyle \pm 0.52}$             & $4.47 {\scriptstyle \pm 0.73}$              & $-87.44 {\scriptstyle \pm 0.86}$ & $\smash{\dashuline{50.62 {\scriptstyle \pm 0.60}}}$             & $\smash{\dashuline{25.76 {\scriptstyle \pm 3.52}}}$             & $-71.18 {\scriptstyle \pm 4.35}$  & $22.57 {\scriptstyle \pm 0.48}$             & $\smash{\dashuline{13.87 {\scriptstyle \pm 0.27}}}$             & $-73.26 {\scriptstyle \pm 1.33}$  \\
ER        & $26.28 {\scriptstyle \pm 0.68}$             & $7.91 {\scriptstyle \pm 1.85}$              & $-82.48 {\scriptstyle \pm 1.72}$ & $49.65 {\scriptstyle \pm 0.83}$             & $21.01 {\scriptstyle \pm 1.97}$             & $-74.91 {\scriptstyle \pm 1.84}$  & $\smash{\dashuline{23.30 {\scriptstyle \pm 0.39}}}$             & $11.71 {\scriptstyle \pm 2.36}$             & $-81.09 {\scriptstyle \pm 2.36}$  \\
EWC       & $\underline{41.71 {\scriptstyle \pm 2.91}}$ & $\underline{21.36 {\scriptstyle \pm 4.97}}$ & $-29.44 {\scriptstyle \pm 3.38}$ & $42.21 {\scriptstyle \pm 0.75}$             & $21.29 {\scriptstyle \pm 3.58}$             & $-69.56 {\scriptstyle \pm 5.28}$  & $13.20 {\scriptstyle \pm 0.37}$             & $4.34 {\scriptstyle \pm 0.53}$              & $-70.72 {\scriptstyle \pm 10.43}$ \\
LwF       & $30.50 {\scriptstyle \pm 0.43}$             & $17.48 {\scriptstyle \pm 0.75}$             & $-59.17 {\scriptstyle \pm 0.71}$ & $44.84 {\scriptstyle \pm 1.03}$             & $22.21 {\scriptstyle \pm 5.48}$             & $-65.94 {\scriptstyle \pm 10.92}$ & $13.37 {\scriptstyle \pm 0.10}$             & $6.38 {\scriptstyle \pm 0.21}$              & $-71.16 {\scriptstyle \pm 0.82}$  \\
MAS       & $\smash{\dashuline{37.28 {\scriptstyle \pm 2.84}}}$             & $16.26 {\scriptstyle \pm 1.83}$             & $-22.20 {\scriptstyle \pm 2.42}$ & $50.36 {\scriptstyle \pm 3.89}$             & $22.28 {\scriptstyle \pm 5.30}$             & $-66.80 {\scriptstyle \pm 6.30}$  & $15.54 {\scriptstyle \pm 0.69}$             & $6.21 {\scriptstyle \pm 1.32}$              & $-64.22 {\scriptstyle \pm 3.62}$  \\
PDGNN     & $\mathbf{46.73 {\scriptstyle \pm 0.50}}$    & $\mathbf{28.26 {\scriptstyle \pm 3.06}}$    & $-62.35 {\scriptstyle \pm 3.29}$ & $\mathbf{78.55 {\scriptstyle \pm 0.68}}$    & $\mathbf{71.15 {\scriptstyle \pm 0.57}}$    & $-24.20 {\scriptstyle \pm 0.57}$  & $\mathbf{51.20 {\scriptstyle \pm 0.54}}$    & $\mathbf{33.78 {\scriptstyle \pm 1.70}}$    & $-50.18 {\scriptstyle \pm 1.50}$  \\
SSM-A-GEM & $19.21 {\scriptstyle \pm 0.72}$             & $6.24 {\scriptstyle \pm 0.57}$              & $-85.67 {\scriptstyle \pm 0.56}$ & $47.80 {\scriptstyle \pm 1.42}$             & $22.07 {\scriptstyle \pm 1.34}$             & $-75.57 {\scriptstyle \pm 1.35}$  & $21.09 {\scriptstyle \pm 0.67}$             & $7.66 {\scriptstyle \pm 1.28}$              & $-86.16 {\scriptstyle \pm 1.20}$  \\
SSM-ER    & $28.73 {\scriptstyle \pm 1.35}$             & $\smash{\dashuline{17.76 {\scriptstyle \pm 2.34}}}$             & $-49.42 {\scriptstyle \pm 2.51}$ & $\underline{59.89 {\scriptstyle \pm 2.43}}$ & $\underline{38.39 {\scriptstyle \pm 8.65}}$ & $-58.01 {\scriptstyle \pm 9.45}$  & $\underline{41.42 {\scriptstyle \pm 0.56}}$ & $\underline{20.97 {\scriptstyle \pm 0.83}}$ & $-70.93 {\scriptstyle \pm 0.83}$  \\
TWP       & $17.72 {\scriptstyle \pm 0.32}$             & $8.14 {\scriptstyle \pm 2.70}$              & $-82.75 {\scriptstyle \pm 2.61}$ & $42.90 {\scriptstyle \pm 0.99}$             & $20.96 {\scriptstyle \pm 2.61}$             & $-74.42 {\scriptstyle \pm 3.16}$  & $13.70 {\scriptstyle \pm 0.75}$             & $4.47 {\scriptstyle \pm 2.45}$              & $-71.14 {\scriptstyle \pm 6.86}$  \\ \midrule
bare      & $15.77 {\scriptstyle \pm 0.23}$             & $3.27 {\scriptstyle \pm 0.77}$              & $-88.15 {\scriptstyle \pm 0.34}$ & $41.98 {\scriptstyle \pm 0.20}$             & $19.74 {\scriptstyle \pm 0.12}$             & $-78.77 {\scriptstyle \pm 0.19}$  & $12.38 {\scriptstyle \pm 0.08}$             & $4.88 {\scriptstyle \pm 0.06}$              & $-90.02 {\scriptstyle \pm 0.55}$  \\ \bottomrule
\end{tabular}
\end{center}
\end{table*}

\begin{table*}[h]
\caption{Results with 1 layer.}
\label{tab:1layer}
\begin{center}
\setlength{\tabcolsep}{3pt}
\begin{tabular}{@{}l|ccc|ccc|ccc@{}}
\toprule
          & \multicolumn{3}{c|}{CoraFull}                                                                                                & \multicolumn{3}{c|}{Amazon Computer}                                                                                         & \multicolumn{3}{c}{Arxiv}                                                                                                    \\
Method    & AAP\% $\uparrow$                            & AP\% $\uparrow$                             & AF\% $\uparrow$                  & AAP\% $\uparrow$                            & AP\% $\uparrow$                             & AF\% $\uparrow$                  & AAP\% $\uparrow$                            & AP\% $\uparrow$                             & AF\% $\uparrow$                  \\ \midrule
A-GEM     & $35.59 {\scriptstyle \pm 0.17}$             & $18.53 {\scriptstyle \pm 0.87}$             & $-76.83 {\scriptstyle \pm 0.85}$ & $46.88 {\scriptstyle \pm 0.51}$             & $21.09 {\scriptstyle \pm 0.86}$             & $-77.63 {\scriptstyle \pm 0.82}$ & $20.20 {\scriptstyle \pm 0.10}$             & $14.98 {\scriptstyle \pm 0.54}$             & $-69.34 {\scriptstyle \pm 0.71}$ \\
ER        & $\mathbf{53.48 {\scriptstyle \pm 0.21}}$    & $\smash{\dashuline{27.03 {\scriptstyle \pm 0.61}}}$             & $-67.09 {\scriptstyle \pm 0.55}$ & $\smash{\dashuline{76.29 {\scriptstyle \pm 0.35}}}$             & $\smash{\dashuline{61.49 {\scriptstyle \pm 9.28}}}$             & $-36.11 {\scriptstyle \pm 9.27}$ & $\smash{\dashuline{44.48 {\scriptstyle \pm 0.17}}}$             & $\smash{\dashuline{31.67 {\scriptstyle \pm 0.58}}}$             & $-60.51 {\scriptstyle \pm 0.49}$ \\
EWC       & $23.67 {\scriptstyle \pm 0.03}$             & $14.96 {\scriptstyle \pm 0.25}$             & $-78.78 {\scriptstyle \pm 0.26}$ & $41.87 {\scriptstyle \pm 0.14}$             & $19.71 {\scriptstyle \pm 0.04}$             & $-77.98 {\scriptstyle \pm 0.77}$ & $12.49 {\scriptstyle \pm 0.01}$             & $4.90 {\scriptstyle \pm 0.00}$              & $-90.70 {\scriptstyle \pm 0.05}$ \\
LwF       & $28.41 {\scriptstyle \pm 0.20}$             & $18.46 {\scriptstyle \pm 1.26}$             & $-62.26 {\scriptstyle \pm 1.25}$ & $49.23 {\scriptstyle \pm 0.66}$             & $27.08 {\scriptstyle \pm 2.44}$             & $-66.36 {\scriptstyle \pm 3.36}$ & $11.86 {\scriptstyle \pm 0.00}$             & $4.79 {\scriptstyle \pm 0.01}$              & $-79.94 {\scriptstyle \pm 0.03}$ \\
MAS       & $29.86 {\scriptstyle \pm 0.08}$             & $\underline{27.71 {\scriptstyle \pm 0.43}}$ & $-67.51 {\scriptstyle \pm 0.48}$ & $41.87 {\scriptstyle \pm 0.14}$             & $19.71 {\scriptstyle \pm 0.04}$             & $-77.98 {\scriptstyle \pm 0.77}$ & $15.22 {\scriptstyle \pm 0.03}$             & $4.97 {\scriptstyle \pm 0.00}$              & $-90.89 {\scriptstyle \pm 0.04}$ \\
PDGNN     & $\underline{51.40 {\scriptstyle \pm 0.20}}$ & $\mathbf{28.29 {\scriptstyle \pm 1.09}}$    & $-63.41 {\scriptstyle \pm 1.22}$ & $\mathbf{85.97 {\scriptstyle \pm 0.17}}$    & $\mathbf{77.88 {\scriptstyle \pm 1.73}}$    & $-18.84 {\scriptstyle \pm 2.29}$ & $\mathbf{51.05 {\scriptstyle \pm 0.18}}$    & $\mathbf{36.63 {\scriptstyle \pm 0.80}}$    & $-51.91 {\scriptstyle \pm 0.58}$ \\
SSM-A-GEM & $34.11 {\scriptstyle \pm 0.30}$             & $19.31 {\scriptstyle \pm 0.67}$             & $-74.63 {\scriptstyle \pm 0.64}$ & $51.45 {\scriptstyle \pm 1.69}$             & $43.10 {\scriptstyle \pm 8.17}$             & $-55.54 {\scriptstyle \pm 8.23}$ & $25.34 {\scriptstyle \pm 0.21}$             & $20.77 {\scriptstyle \pm 0.72}$             & $-68.33 {\scriptstyle \pm 0.48}$ \\
SSM-ER    & $\smash{\dashuline{44.05 {\scriptstyle \pm 0.24}}}$             & $21.93 {\scriptstyle \pm 0.62}$             & $-71.92 {\scriptstyle \pm 0.50}$ & $\underline{77.84 {\scriptstyle \pm 0.63}}$ & $\underline{61.69 {\scriptstyle \pm 7.57}}$ & $-36.28 {\scriptstyle \pm 7.72}$ & $\underline{48.41 {\scriptstyle \pm 0.23}}$ & $\underline{34.51 {\scriptstyle \pm 1.06}}$ & $-54.53 {\scriptstyle \pm 1.05}$ \\
TWP       & $23.67 {\scriptstyle \pm 0.02}$             & $14.89 {\scriptstyle \pm 0.19}$             & $-78.72 {\scriptstyle \pm 0.21}$ & $42.92 {\scriptstyle \pm 0.89}$             & $19.63 {\scriptstyle \pm 0.14}$             & $-77.19 {\scriptstyle \pm 1.26}$ & $12.82 {\scriptstyle \pm 0.03}$             & $4.95 {\scriptstyle \pm 0.01}$              & $-90.85 {\scriptstyle \pm 0.04}$ \\ \midrule
bare      & $23.64 {\scriptstyle \pm 0.03}$             & $15.01 {\scriptstyle \pm 0.14}$             & $-78.80 {\scriptstyle \pm 0.11}$ & $42.28 {\scriptstyle \pm 0.12}$             & $19.69 {\scriptstyle \pm 0.06}$             & $-79.05 {\scriptstyle \pm 0.13}$ & $12.47 {\scriptstyle \pm 0.02}$             & $4.91 {\scriptstyle \pm 0.01}$              & $-90.72 {\scriptstyle \pm 0.07}$ \\ \bottomrule
\end{tabular}
\end{center}
\end{table*}

\clearpage
\section{Performance visualization}\label{app:plots}

\subsection{Impact of buffer size}

\begin{figure*}[h!]
\centering
\subfloat{\includegraphics[width=0.6\textwidth]{figures/buffer/legend.png}}
\vspace{-0.25em}
\\
\subfloat[CoraFull]{%
  \includegraphics[width=0.24\textwidth]{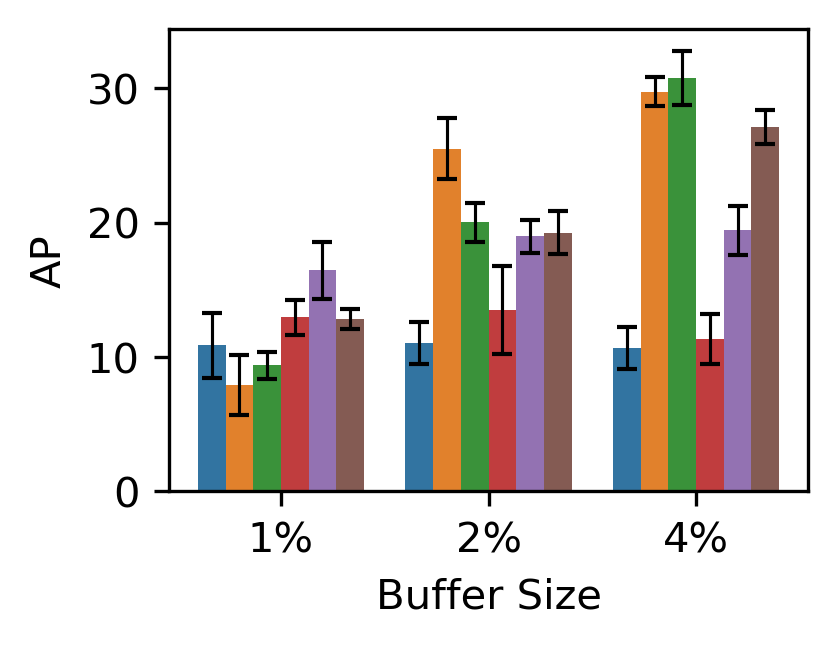}%
  \label{fig:buffer_cora_AA}}
\hfill
\subfloat[Arxiv]{%
  \includegraphics[width=0.24\textwidth]{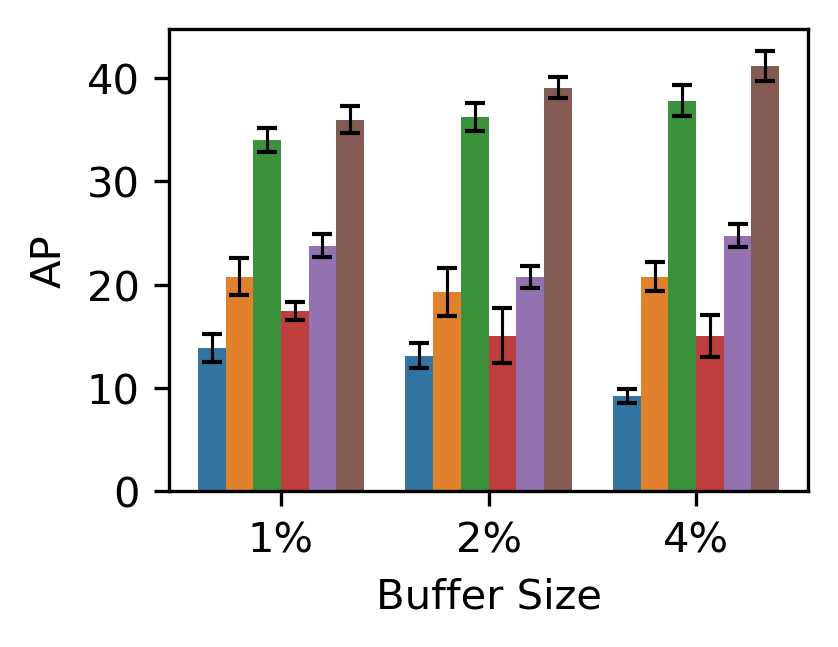}%
  \label{fig:buffer_arxiv_AA}}
\hfill
\subfloat[Reddit]{%
  \includegraphics[width=0.24\textwidth]{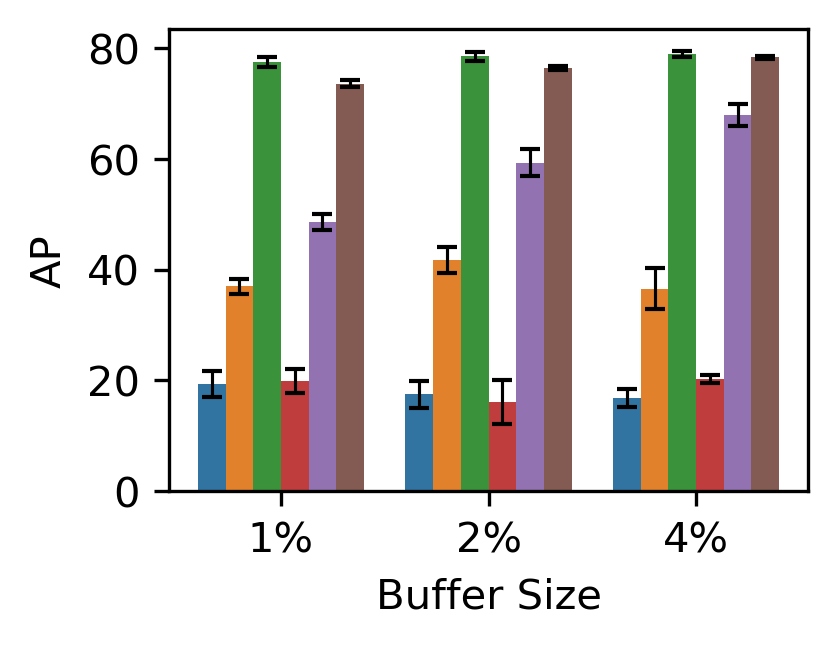}%
  \label{fig:buffer_reddit_AA}}
\hfill
\subfloat[Amazon Computer]{%
  \includegraphics[width=0.24\textwidth]{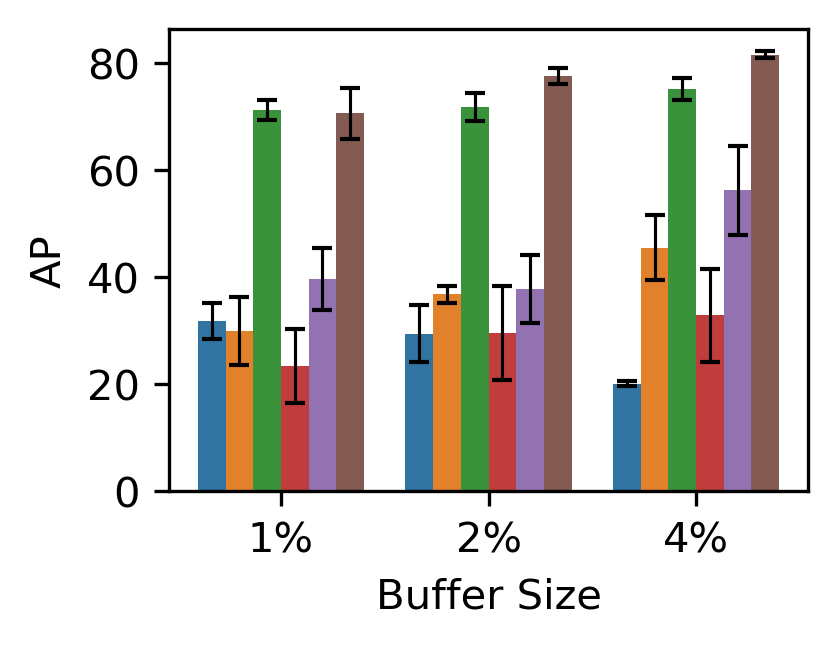}%
  \label{fig:buffer_computer_AA}}
\\
\subfloat[Roman Empire]{%
  \includegraphics[width=0.24\textwidth]{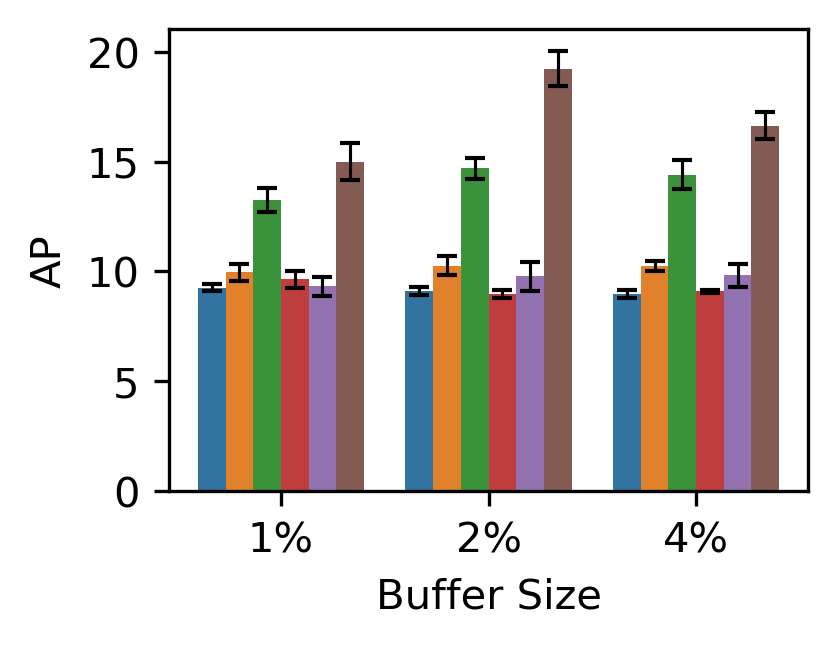}%
  \label{fig:buffer_roman_AA}}
\hfill
\subfloat[Elliptic]{%
  \includegraphics[width=0.24\textwidth]{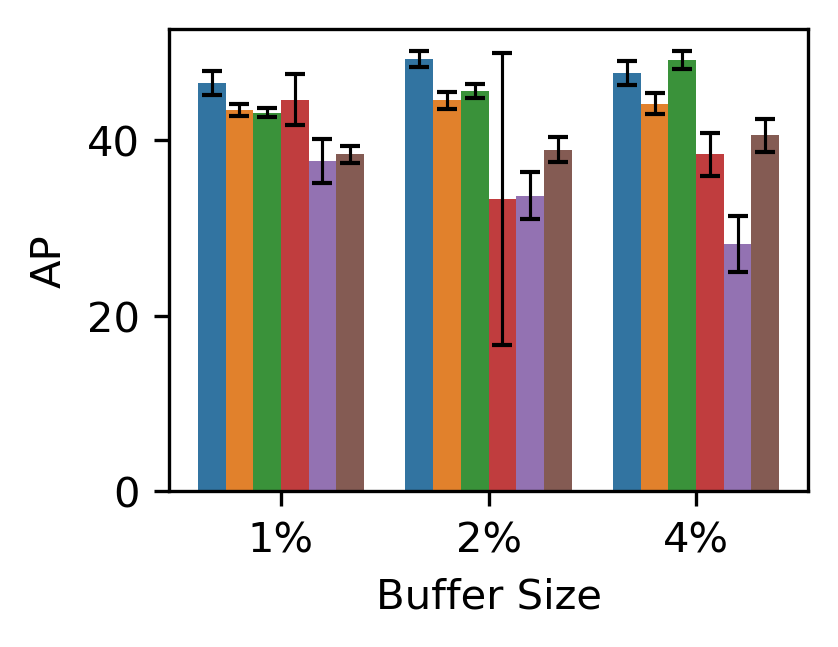}%
  \label{fig:buffer_elliptic_AA}}
\hfill
\subfloat[Arxiv (time-incr.)]{%
  \includegraphics[width=0.24\textwidth]{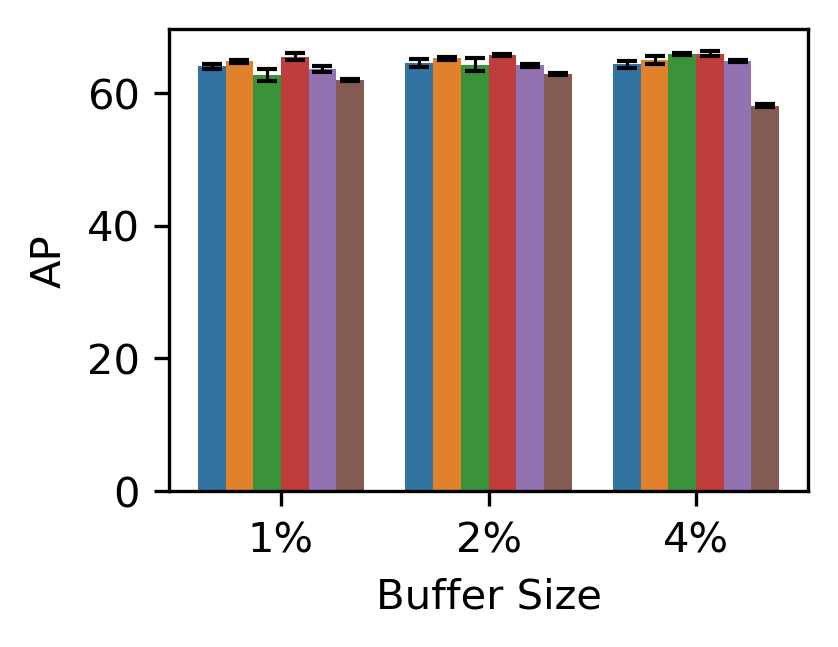}%
  \label{fig:buffer_arxivTI_AA}}
\hfill
\subfloat[Products]{%
  \includegraphics[width=0.24\textwidth]{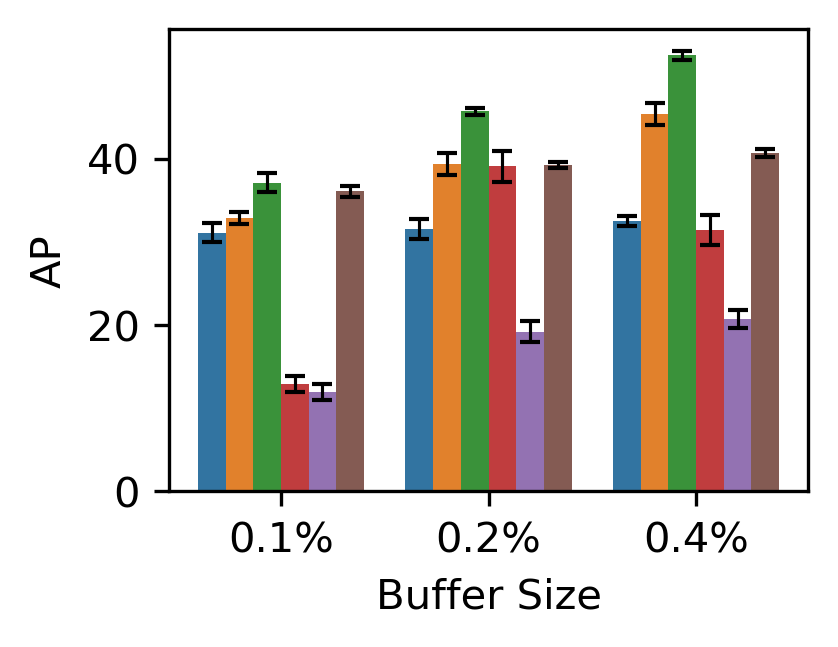}%
  \label{fig:buffer_products_AA}}
\caption{Results (Average Performance) with different buffer sizes for replay methods across the datasets.}
\label{fig:buffer_results_appendix}
\end{figure*}

For completeness of our results on the impact of buffer size on model performance for replay methods, for which Figure \ref{fig:buffer_results} of the main body of the paper shows AAP, we show here in Figure \ref{fig:buffer_results_appendix} the impact on AP.

\subsection{Anytime evaluation plots}

\begin{figure}[h!]
\begin{minipage}[c]{0.49\linewidth}
    \centering
    \includegraphics[width=\textwidth]{figures/cora.png}
    \caption{Anytime evaluation on CoraFull dataset.}
    \label{fig:cora_anytime}
\end{minipage}
\hfill
\begin{minipage}[c]{0.49\linewidth}
    \centering
    \includegraphics[width=\textwidth]{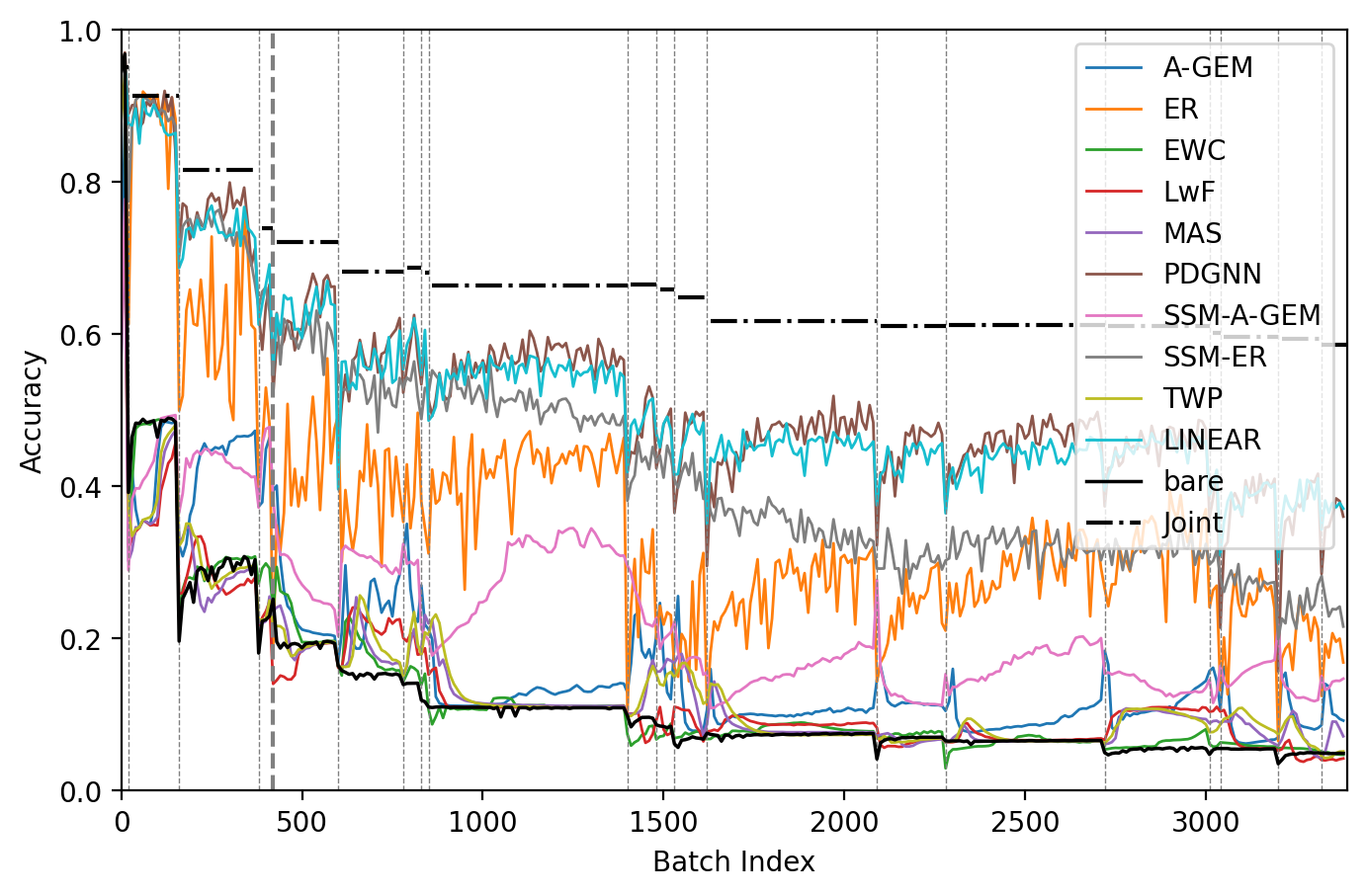}
    \caption{Anytime evaluation on Arxiv dataset.}
    \label{fig:arxiv_anytime}
\end{minipage}
\end{figure}

\begin{figure}[h!]
\begin{minipage}[c]{0.49\linewidth}
    \centering
    \includegraphics[width=\textwidth]{figures/reddit.png}
    \caption{Anytime evaluation on Reddit dataset.}
    \label{fig:reddit_anytime}
\end{minipage}
\hfill
\begin{minipage}[c]{0.49\linewidth}
    \centering
    \includegraphics[width=\textwidth]{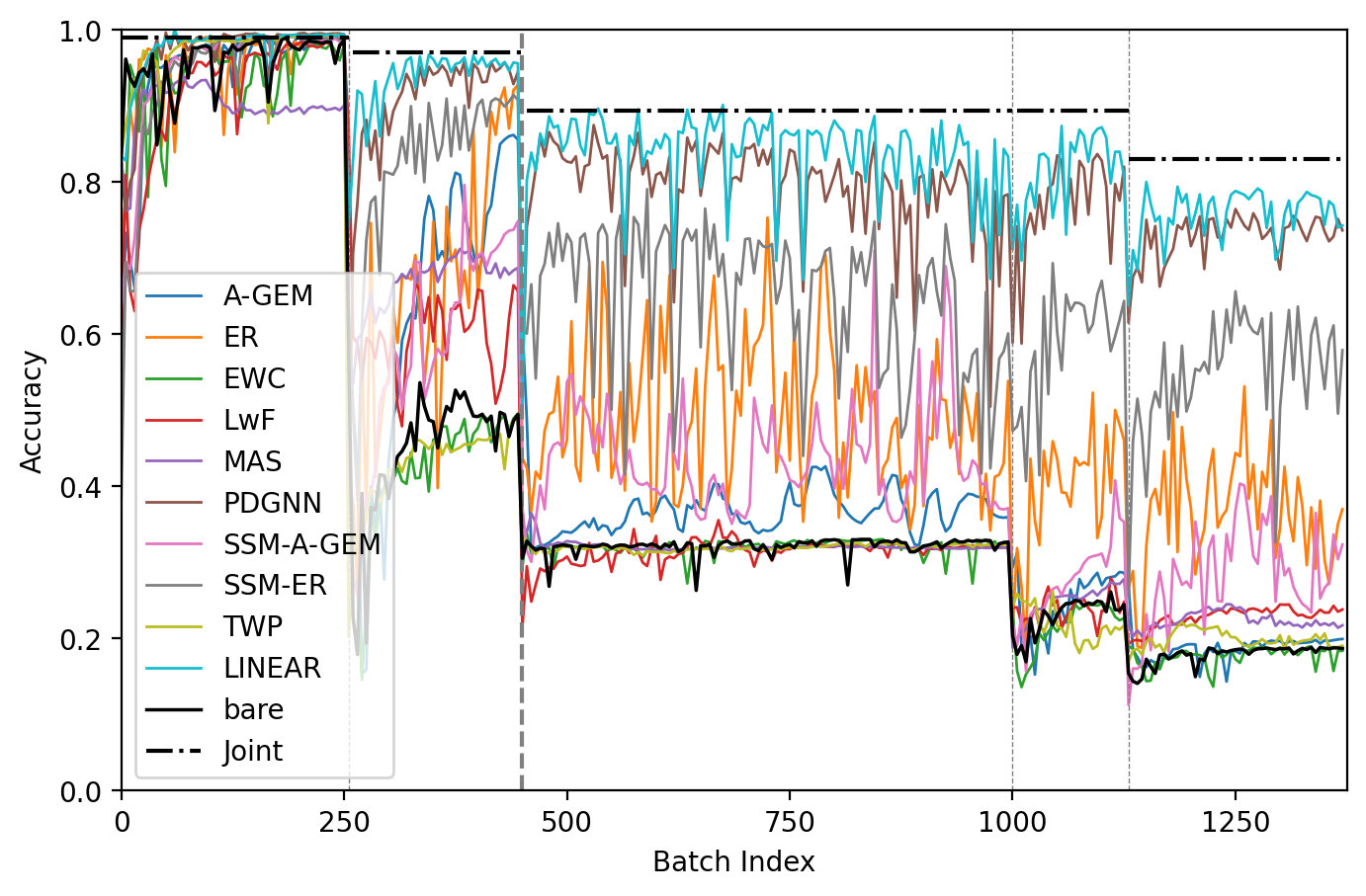}
    \caption{Anytime evaluation on Amazon Computer dataset.}
    \label{fig:computer_anytime}
\end{minipage}
\end{figure}

\begin{figure}[h!]
\begin{minipage}[c]{0.49\linewidth}
    \centering
    \includegraphics[width=\textwidth]{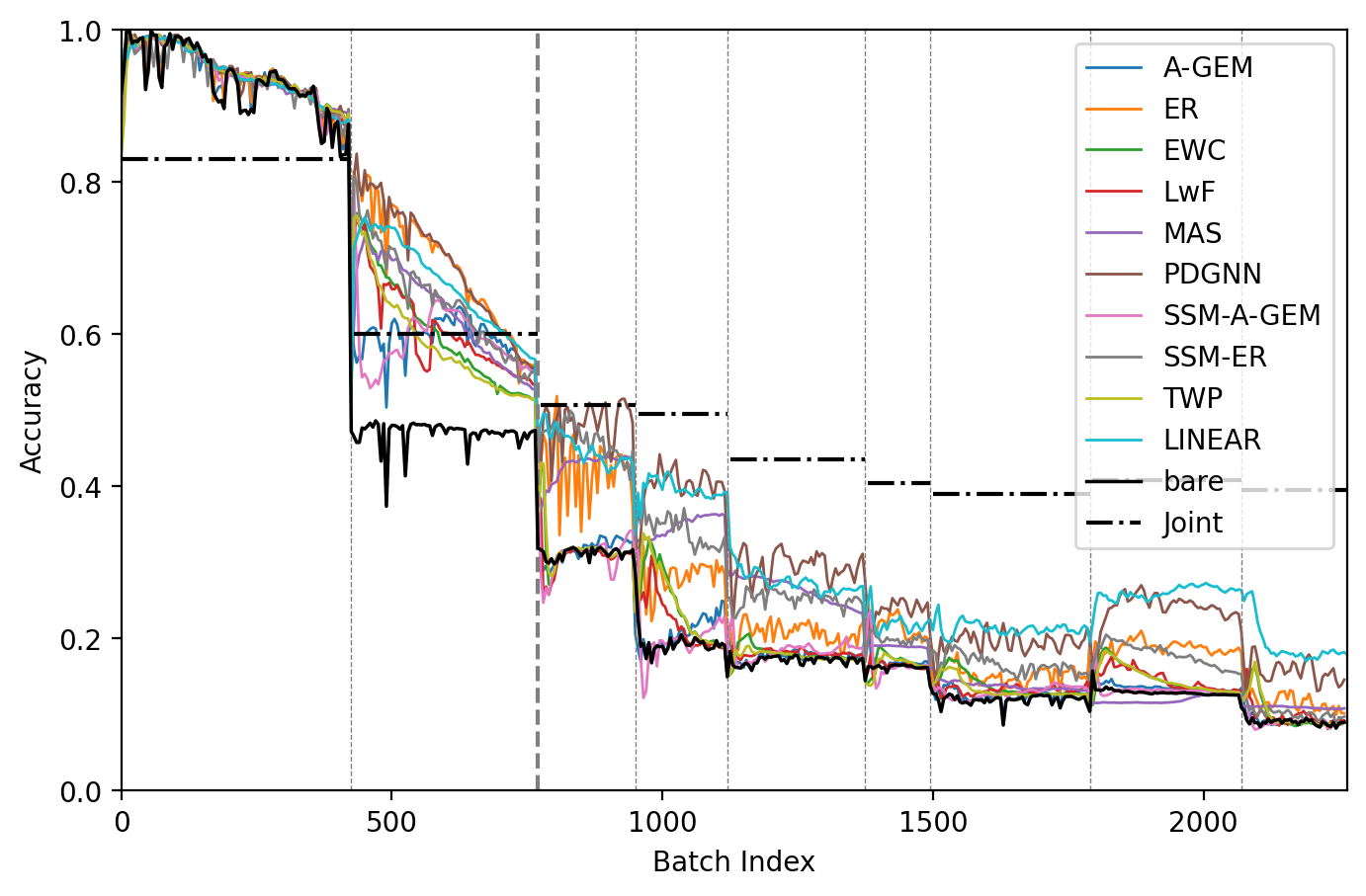}
    \caption{Anytime evaluation on Roman Empire dataset.}
    \label{fig:roman_anytime}
\end{minipage}
\hfill
\begin{minipage}[c]{0.49\linewidth}
    \centering
    \includegraphics[width=\textwidth]{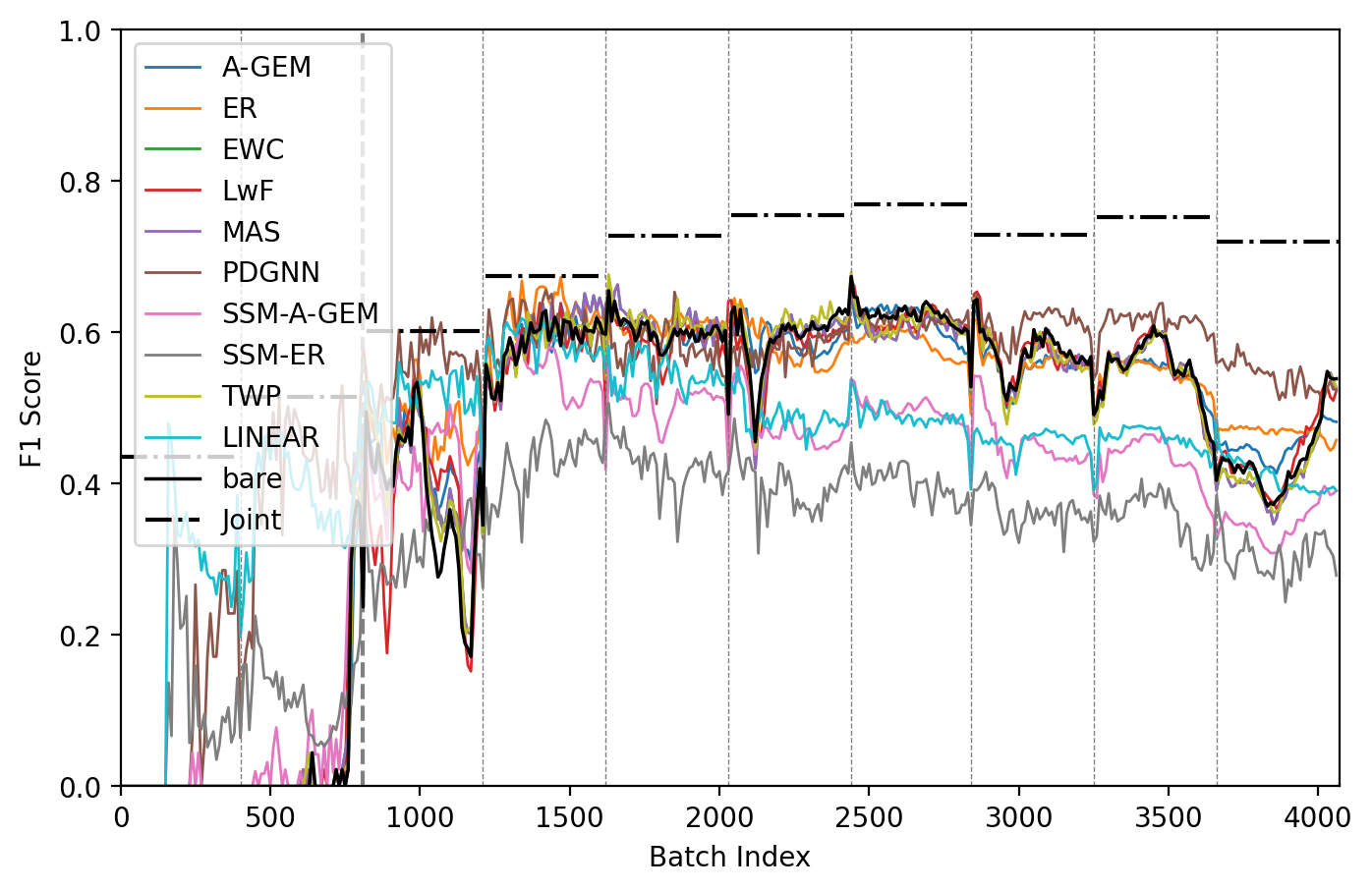}
    \caption{Anytime evaluation on Elliptic dataset.}
    \label{fig:elliptic_anytime}
\end{minipage}
\end{figure}

\begin{figure}[h!]
\begin{minipage}[c]{0.49\linewidth}
    \centering
    \includegraphics[width=\textwidth]{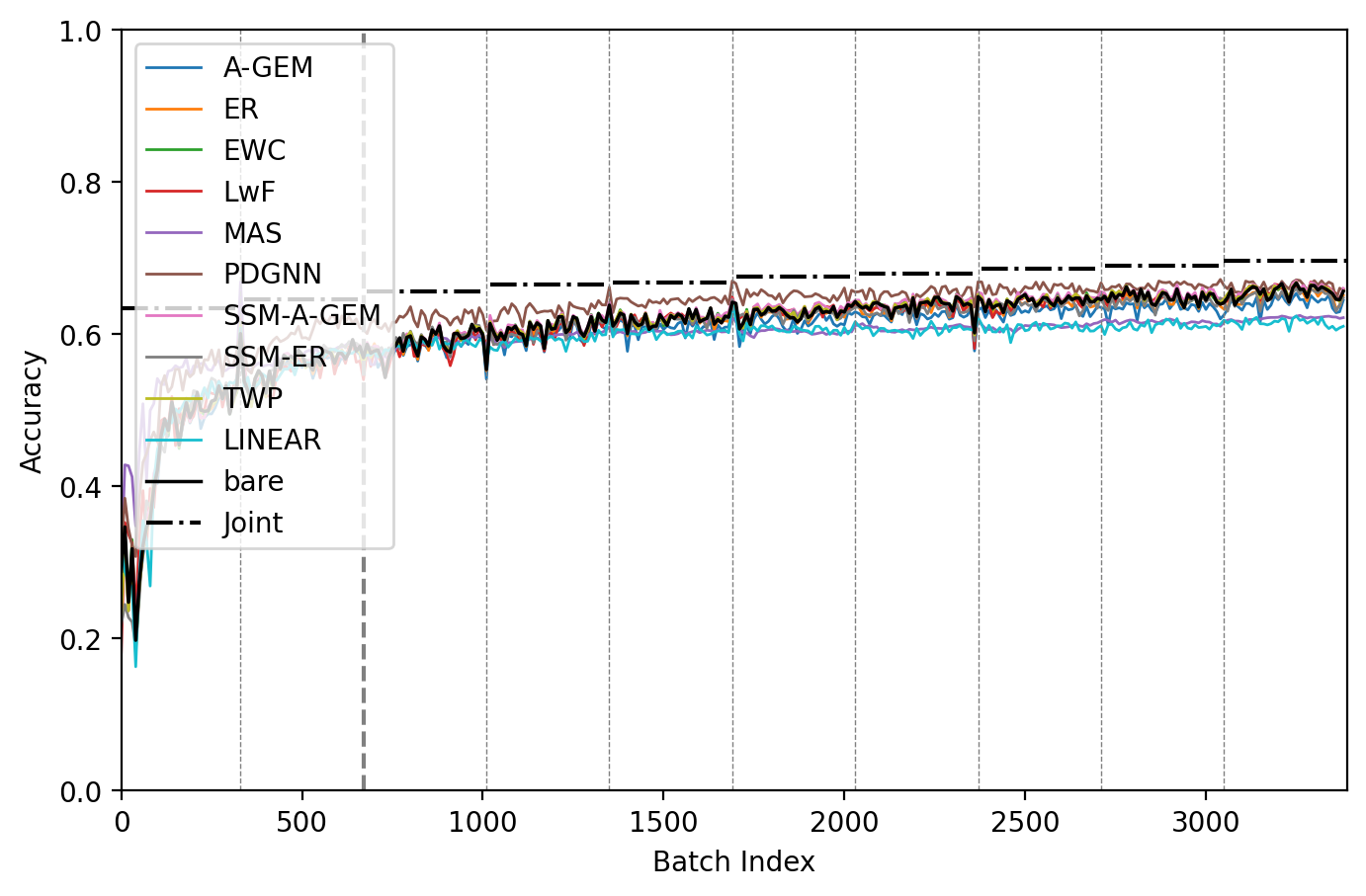}
    \caption{Anytime evaluation on Arxiv dataset with time-incremental stream.}
    \label{fig:arxivTI_anytime}
\end{minipage}
\hfill
\begin{minipage}[c]{0.49\linewidth}
    \centering
    \includegraphics[width=\textwidth]{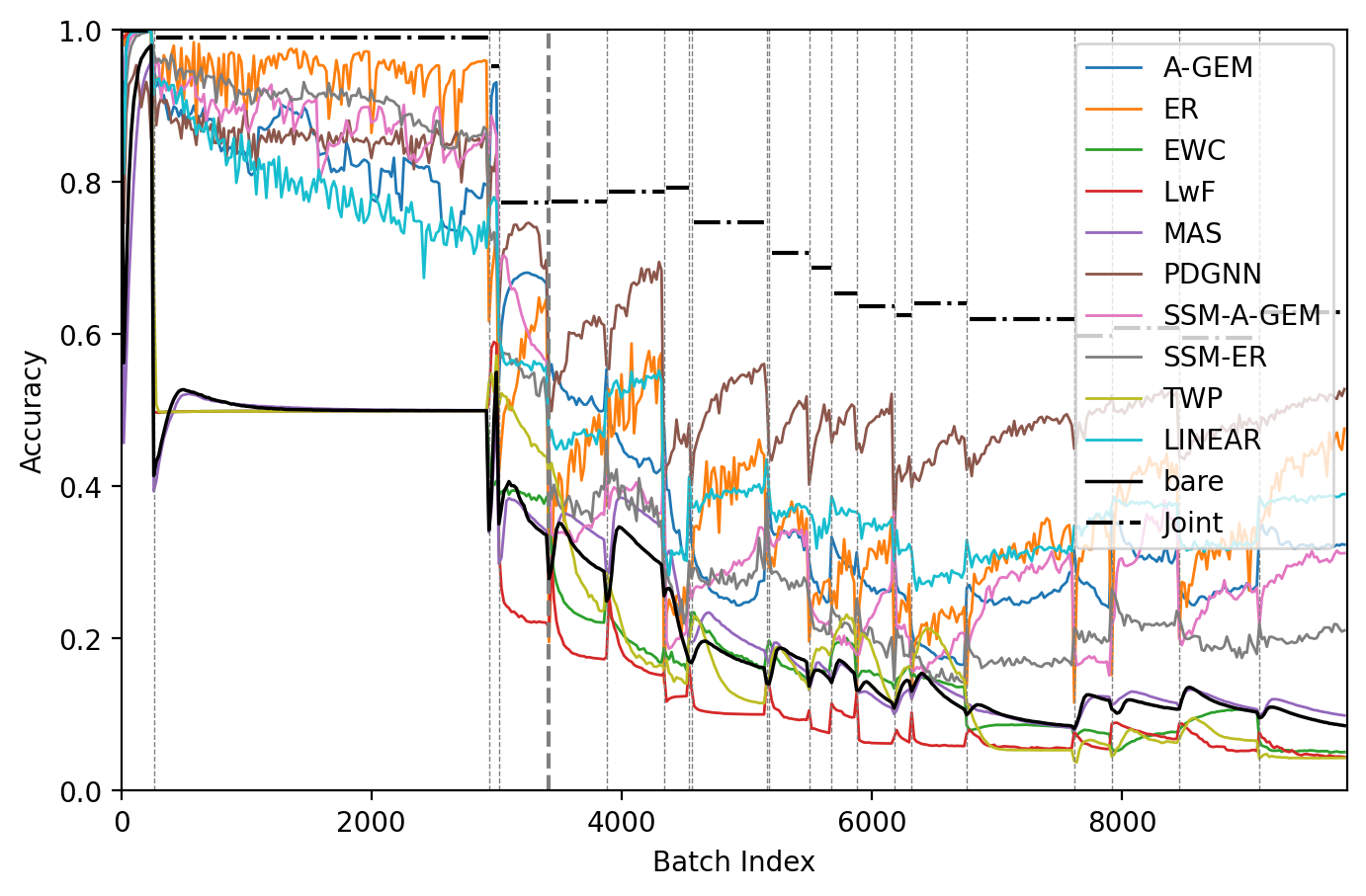}
    \caption{Anytime evaluation on Products dataset.}
    \label{fig:products_anytime}
\end{minipage}
\end{figure}

\begin{figure*}[!t]
\centering
\subfloat[A-GEM]{%
    \includegraphics[width=0.32\textwidth]{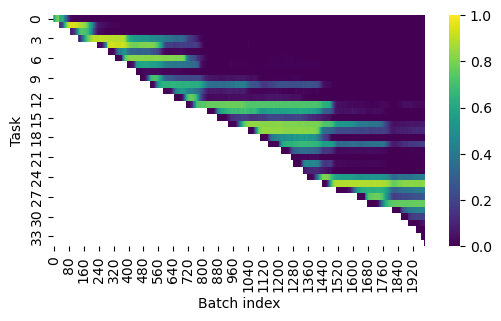}%
    \label{fig:cora_agem}}
\hfill
\subfloat[ER]{%
    \includegraphics[width=0.32\textwidth]{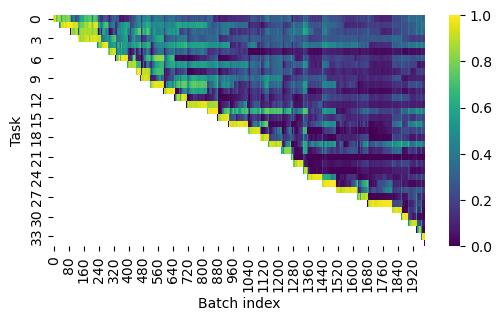}%
    \label{fig:cora_er}}
\hfill
\subfloat[EWC]{%
    \includegraphics[width=0.32\textwidth]{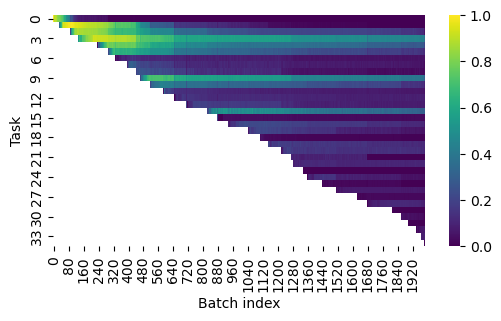}%
    \label{fig:cora_ewc}}
\\
\subfloat[LwF]{%
    \includegraphics[width=0.32\textwidth]{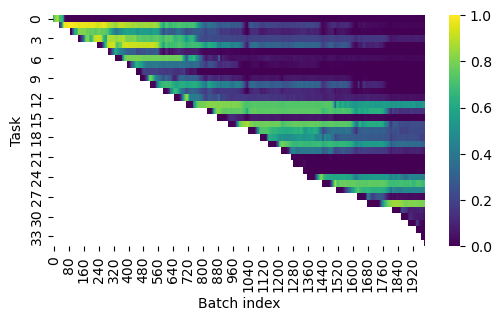}%
    \label{fig:cora_lwf}}
\hfill
\subfloat[MAS]{%
    \includegraphics[width=0.32\textwidth]{figures/anytime/corafull_mas.png}%
    \label{fig:cora_mas}}
\hfill
\subfloat[PDGNN]{%
    \includegraphics[width=0.32\textwidth]{figures/anytime/corafull_pdgnn.png}%
    \label{fig:cora_pdgnn}}
\\
\subfloat[SSM-A-GEM]{%
    \includegraphics[width=0.32\textwidth]{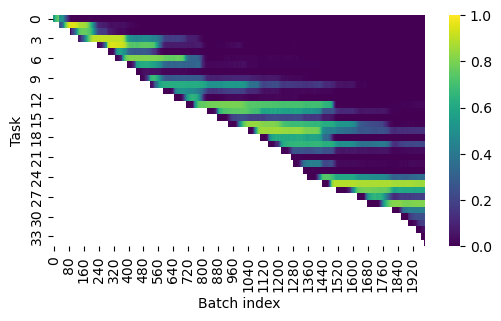}%
    \label{fig:cora_ssmgem}}
\hfill
\subfloat[SSM-ER]{%
    \includegraphics[width=0.32\textwidth]{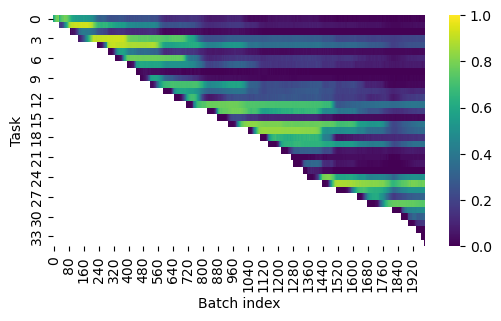}%
    \label{fig:cora_ssm}}
\hfill
\subfloat[TWP]{%
    \includegraphics[width=0.32\textwidth]{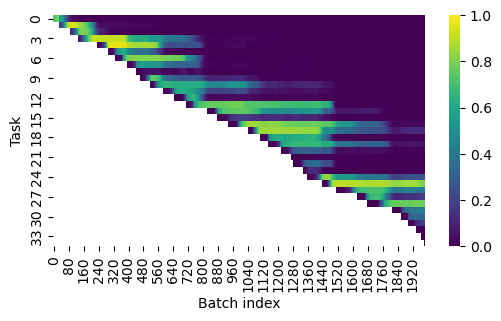}%
    \label{fig:cora_twp}}
\\
\subfloat[LINEAR]{%
    \includegraphics[width=0.32\textwidth]{figures/anytime/corafull_malc.png}%
    \label{fig:cora_malc}}
\hspace{2em}
\subfloat[bare]{%
    \includegraphics[width=0.32\textwidth]{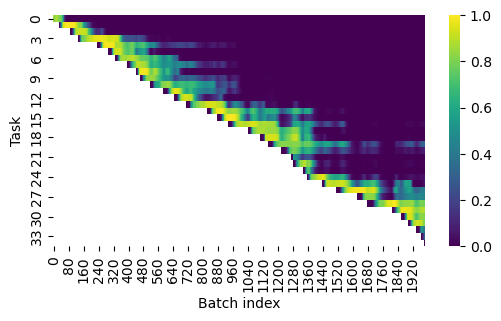}%
    \label{fig:cora_bare}}
\caption{Anytime evaluation by task for the CoraFull dataset.}
\label{fig:heatmaps_cora}
\end{figure*}

\begin{figure*}[!t]
\centering
\subfloat[A-GEM]{%
    \includegraphics[width=0.32\textwidth]{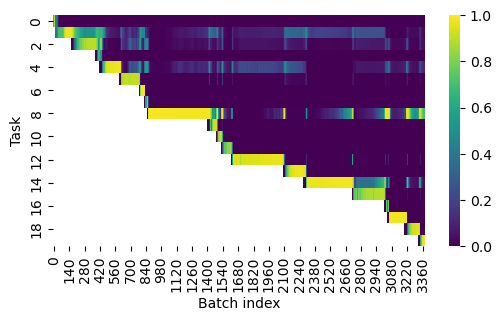}%
    \label{fig:arxiv_agem}}
\hfill
\subfloat[ER]{%
    \includegraphics[width=0.32\textwidth]{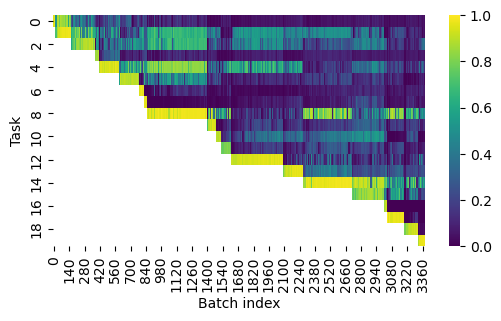}%
    \label{fig:arxiv_er}}
\hfill
\subfloat[EWC]{%
    \includegraphics[width=0.32\textwidth]{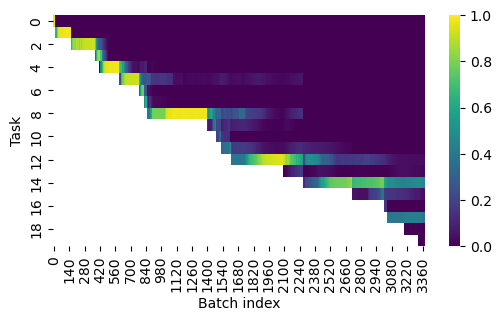}%
    \label{fig:arxiv_ewc}}
\\
\subfloat[LwF]{%
    \includegraphics[width=0.32\textwidth]{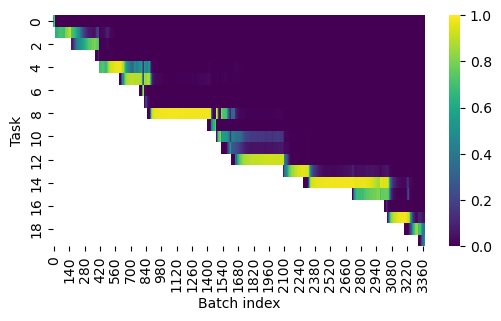}%
    \label{fig:arxiv_lwf}}
\hfill
\subfloat[MAS]{%
    \includegraphics[width=0.32\textwidth]{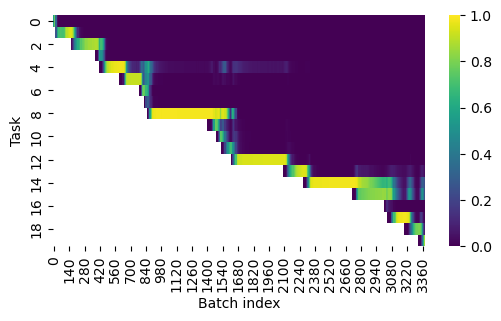}%
    \label{fig:arxiv_mas}}
\hfill
\subfloat[PDGNN]{%
    \includegraphics[width=0.32\textwidth]{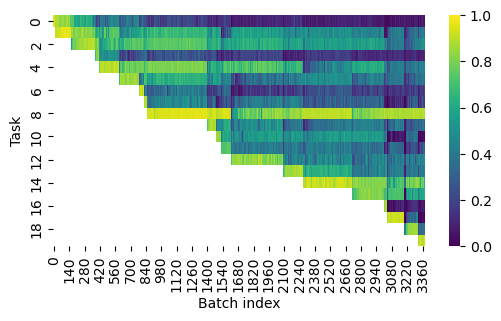}%
    \label{fig:arxiv_pdgnn}}
\\
\subfloat[SSM-A-GEM]{%
    \includegraphics[width=0.32\textwidth]{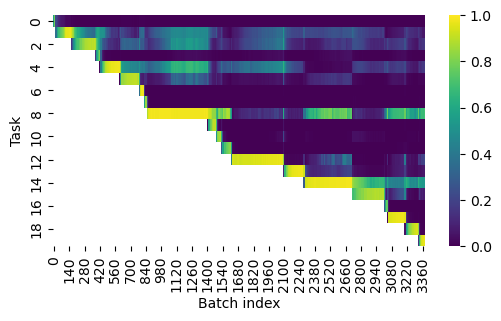}%
    \label{fig:arxiv_ssmgem}}
\hfill
\subfloat[SSM-ER]{%
    \includegraphics[width=0.32\textwidth]{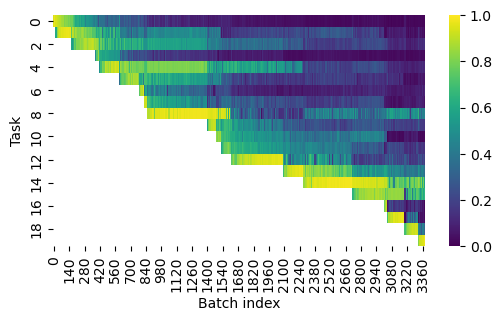}%
    \label{fig:arxiv_ssm}}
\hfill
\subfloat[TWP]{%
    \includegraphics[width=0.32\textwidth]{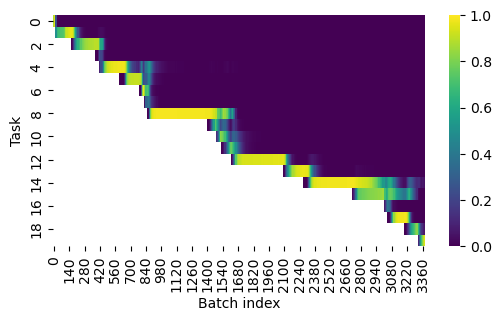}%
    \label{fig:arxiv_twp}}
\\
\subfloat[LINEAR]{%
    \includegraphics[width=0.32\textwidth]{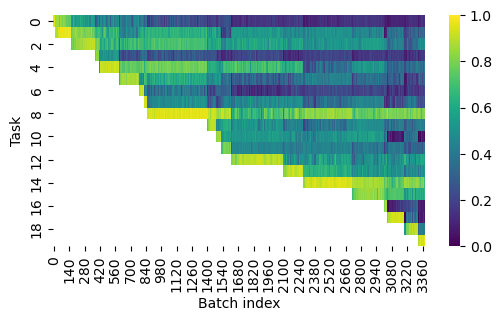}%
    \label{fig:arxiv_malc}}
\hspace{2em}
\subfloat[bare]{%
    \includegraphics[width=0.32\textwidth]{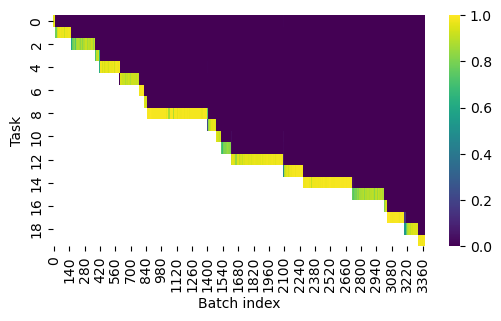}%
    \label{fig:arxiv_bare}}
\caption{Anytime evaluation by task for the Arxiv dataset.}
\label{fig:heatmaps_arxiv}
\end{figure*}

\begin{figure*}[!t]
\centering
\subfloat[A-GEM]{%
    \includegraphics[width=0.32\textwidth]{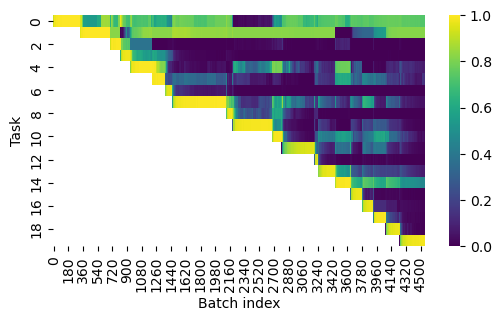}%
    \label{fig:reddit_agem}}
\hfill
\subfloat[ER]{%
    \includegraphics[width=0.32\textwidth]{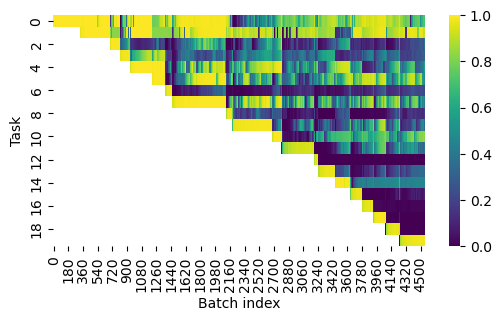}%
    \label{fig:reddit_er}}
\hfill
\subfloat[EWC]{%
    \includegraphics[width=0.32\textwidth]{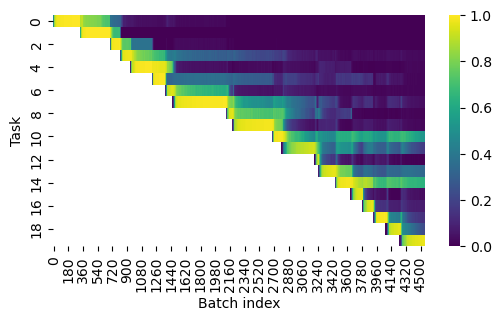}%
    \label{fig:reddit_ewc}}
\\
\subfloat[LwF]{%
    \includegraphics[width=0.32\textwidth]{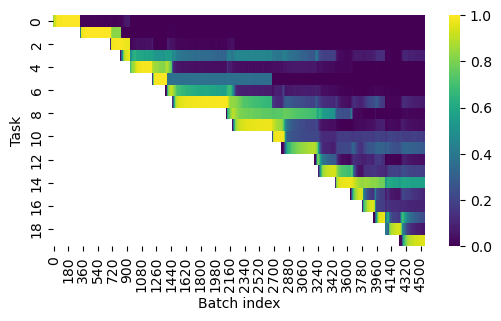}%
    \label{fig:reddit_lwf}}
\hfill
\subfloat[MAS]{%
    \includegraphics[width=0.32\textwidth]{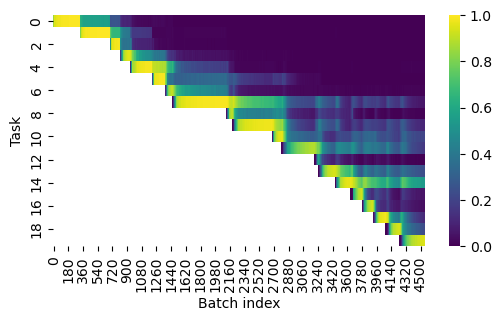}%
    \label{fig:reddit_mas}}
\hfill
\subfloat[PDGNN]{%
    \includegraphics[width=0.32\textwidth]{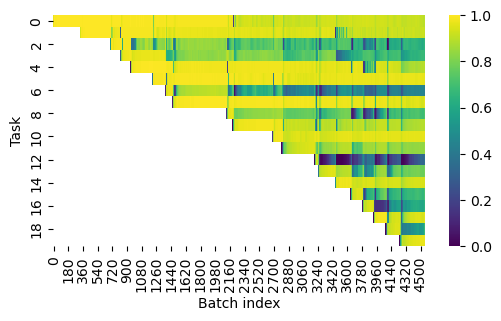}%
    \label{fig:reddit_pdgnn}}
\\
\subfloat[SSM-A-GEM]{%
    \includegraphics[width=0.32\textwidth]{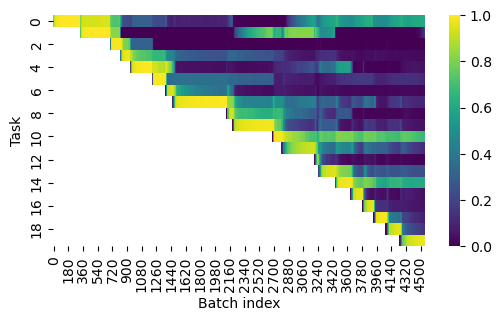}%
    \label{fig:reddit_ssmgem}}
\hfill
\subfloat[SSM-ER]{%
    \includegraphics[width=0.32\textwidth]{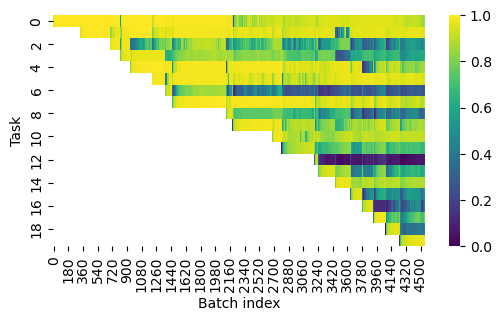}%
    \label{fig:reddit_ssm}}
\hfill
\subfloat[TWP]{%
    \includegraphics[width=0.32\textwidth]{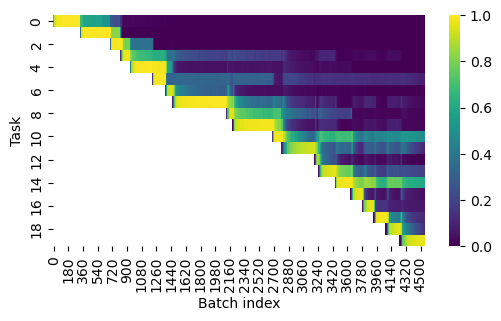}%
    \label{fig:reddit_twp}}
\\
\subfloat[LINEAR]{%
    \includegraphics[width=0.32\textwidth]{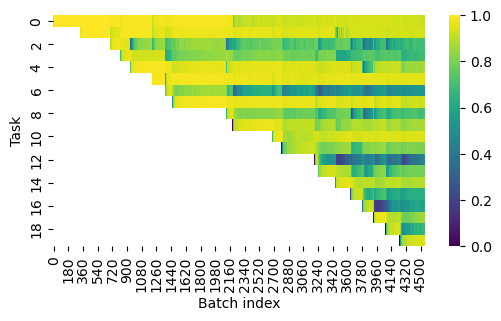}%
    \label{fig:reddit_malc}}
\hspace{2em}
\subfloat[bare]{%
    \includegraphics[width=0.32\textwidth]{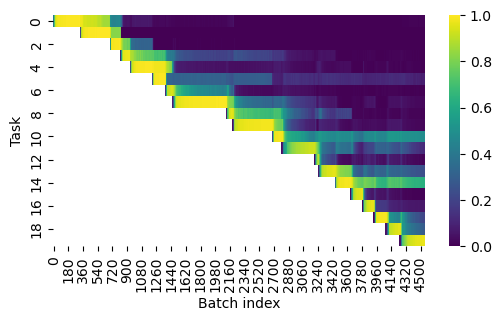}%
    \label{fig:reddit_bare}}
\caption{Anytime evaluation by task for the Reddit dataset.}
\label{fig:heatmaps_reddit}
\end{figure*}

\begin{figure*}[!t]
\centering
\subfloat[A-GEM]{%
    \includegraphics[width=0.32\textwidth]{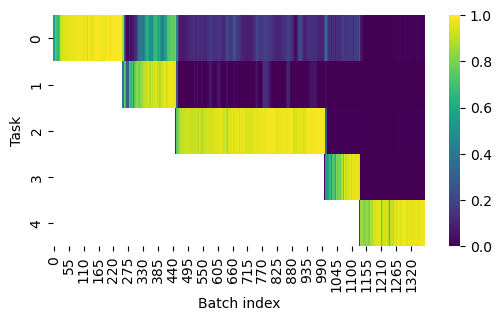}%
    \label{fig:amazoncomputer_agem}}
\hfill
\subfloat[ER]{%
    \includegraphics[width=0.32\textwidth]{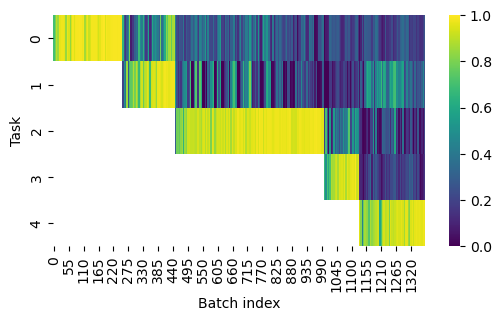}%
    \label{fig:amazoncomputer_er}}
\hfill
\subfloat[EWC]{%
    \includegraphics[width=0.32\textwidth]{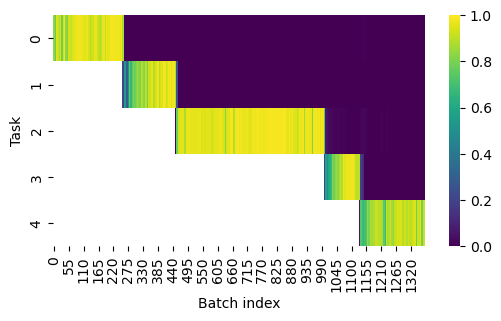}%
    \label{fig:amazoncomputer_ewc}}
\\
\subfloat[LwF]{%
    \includegraphics[width=0.32\textwidth]{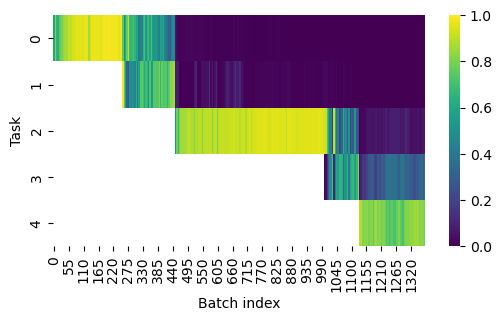}%
    \label{fig:amazoncomputer_lwf}}
\hfill
\subfloat[MAS]{%
    \includegraphics[width=0.32\textwidth]{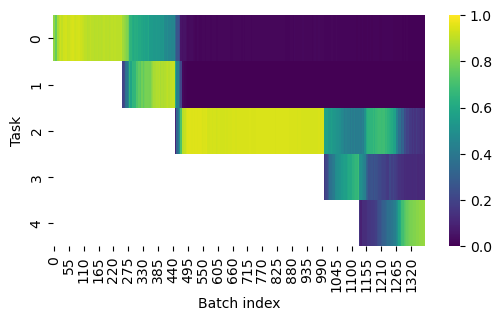}%
    \label{fig:amazoncomputer_mas}}
\hfill
\subfloat[PDGNN]{%
    \includegraphics[width=0.32\textwidth]{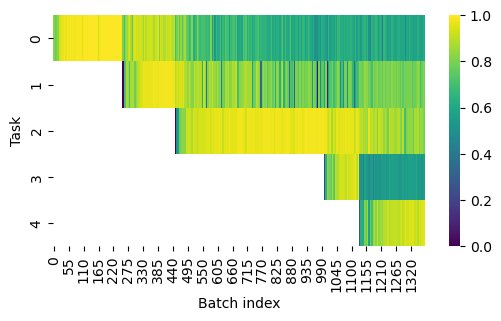}%
    \label{fig:amazoncomputer_pdgnn}}
\\
\subfloat[SSM-A-GEM]{%
    \includegraphics[width=0.32\textwidth]{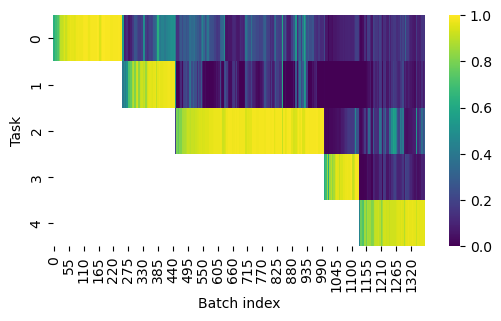}%
    \label{fig:amazoncomputer_ssmgem}}
\hfill
\subfloat[SSM-ER]{%
    \includegraphics[width=0.32\textwidth]{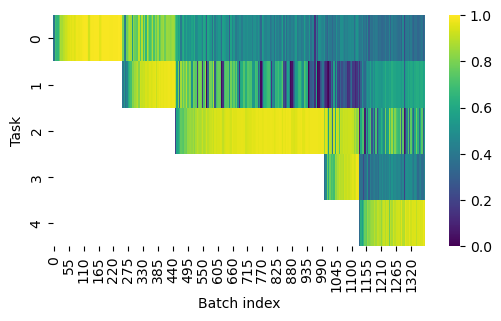}%
    \label{fig:amazoncomputer_ssm}}
\hfill
\subfloat[TWP]{%
    \includegraphics[width=0.32\textwidth]{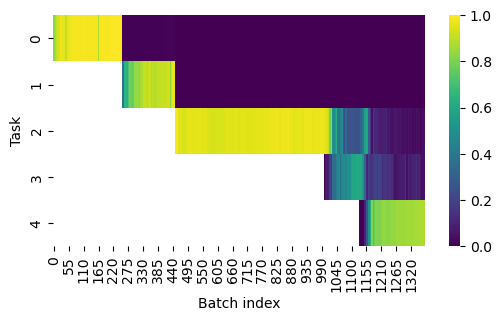}%
    \label{fig:amazoncomputer_twp}}
\\
\subfloat[LINEAR]{%
    \includegraphics[width=0.32\textwidth]{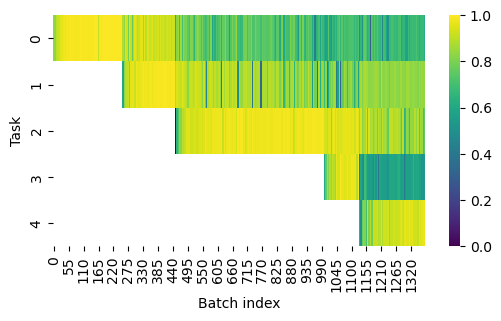}%
    \label{fig:amazoncomputer_malc}}
\hspace{2em}
\subfloat[bare]{%
    \includegraphics[width=0.32\textwidth]{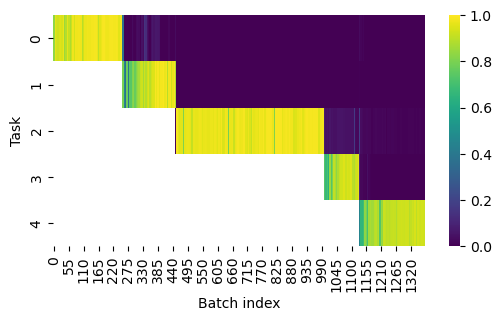}%
    \label{fig:amazoncomputer_bare}}
\caption{Anytime evaluation by task for the Amazon Computer dataset.}
\label{fig:heatmaps_amazoncomputer}
\end{figure*}

\begin{figure*}[!t]
\centering
\subfloat[A-GEM]{%
    \includegraphics[width=0.32\textwidth]{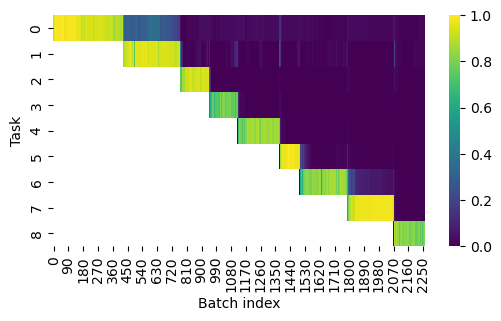}%
    \label{fig:romanempire_agem}}
\hfill
\subfloat[ER]{%
    \includegraphics[width=0.32\textwidth]{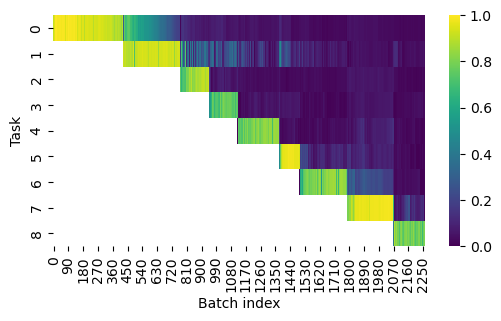}%
    \label{fig:romanempire_er}}
\hfill
\subfloat[EWC]{%
    \includegraphics[width=0.32\textwidth]{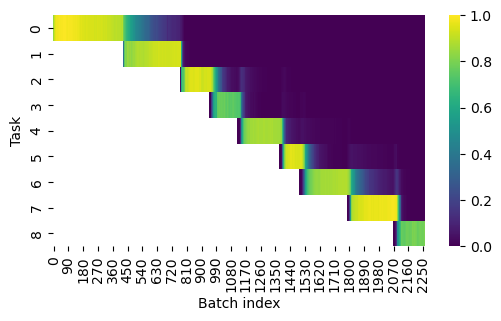}%
    \label{fig:romanempire_ewc}}
\\
\subfloat[LwF]{%
    \includegraphics[width=0.32\textwidth]{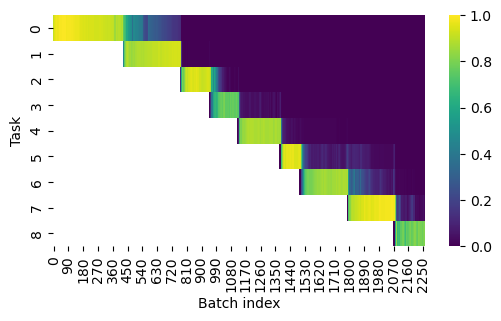}%
    \label{fig:romanempire_lwf}}
\hfill
\subfloat[MAS]{%
    \includegraphics[width=0.32\textwidth]{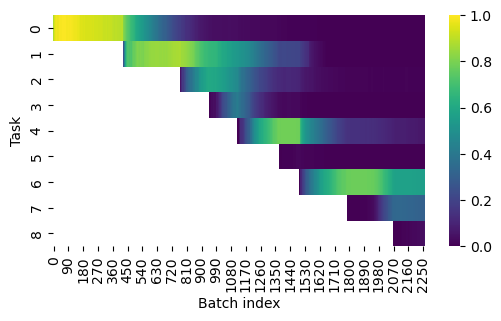}%
    \label{fig:romanempire_mas}}
\hfill
\subfloat[PDGNN]{%
    \includegraphics[width=0.32\textwidth]{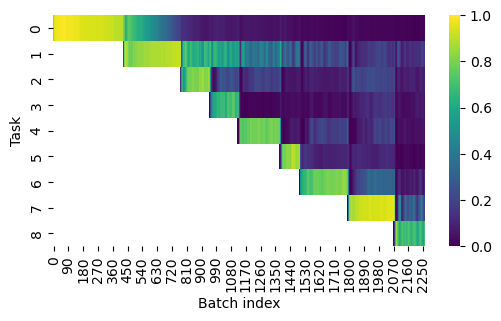}%
    \label{fig:romanempire_pdgnn}}
\\
\subfloat[SSM-A-GEM]{%
    \includegraphics[width=0.32\textwidth]{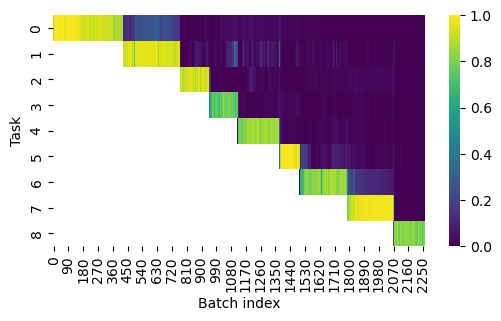}%
    \label{fig:romanempire_ssmgem}}
\hfill
\subfloat[SSM-ER]{%
    \includegraphics[width=0.32\textwidth]{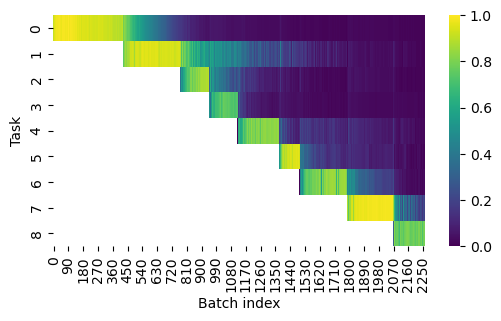}%
    \label{fig:romanempire_ssm}}
\hfill
\subfloat[TWP]{%
    \includegraphics[width=0.32\textwidth]{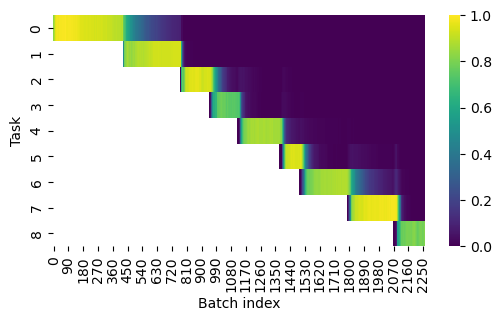}%
    \label{fig:romanempire_twp}}
\\
\subfloat[LINEAR]{%
    \includegraphics[width=0.32\textwidth]{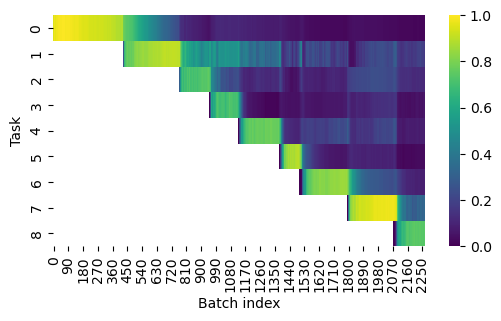}%
    \label{fig:romanempire_malc}}
\hspace{2em}
\subfloat[bare]{%
    \includegraphics[width=0.32\textwidth]{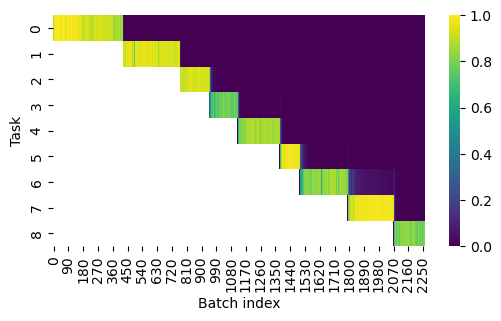}%
    \label{fig:romanempire_bare}}
\caption{Anytime evaluation by task for the Roman Empire dataset.}
\label{fig:heatmaps_romanempire}
\end{figure*}

\begin{figure*}[!t]
\centering
\subfloat[A-GEM]{%
    \includegraphics[width=0.32\textwidth]{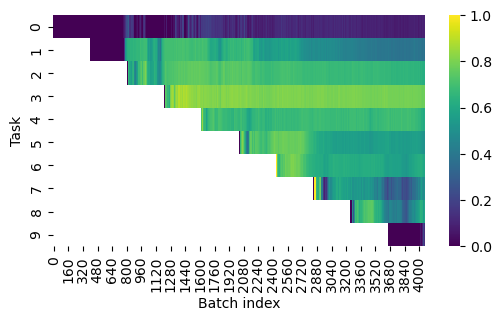}%
    \label{fig:elliptic_agem}}
\hfill
\subfloat[ER]{%
    \includegraphics[width=0.32\textwidth]{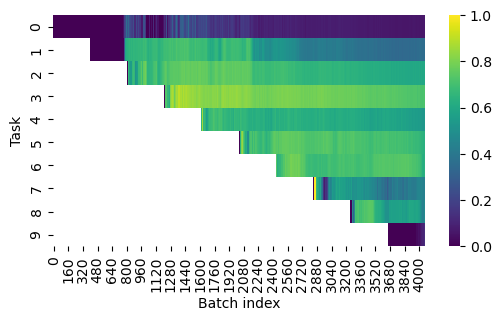}%
    \label{fig:elliptic_er}}
\hfill
\subfloat[EWC]{%
    \includegraphics[width=0.32\textwidth]{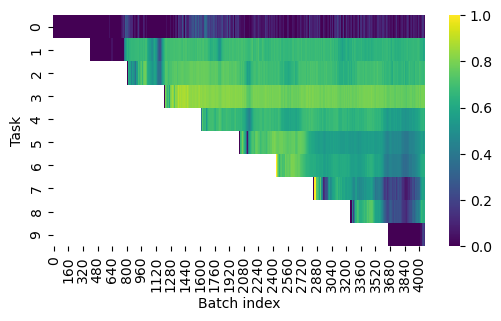}%
    \label{fig:elliptic_ewc}}
\\
\subfloat[LwF]{%
    \includegraphics[width=0.32\textwidth]{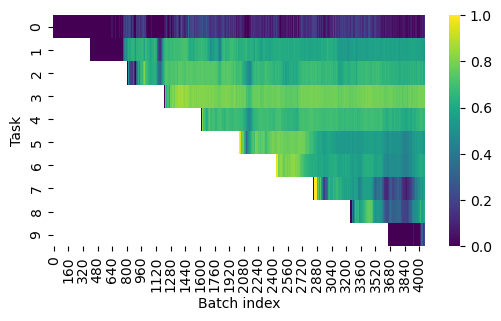}%
    \label{fig:elliptic_lwf}}
\hfill
\subfloat[MAS]{%
    \includegraphics[width=0.32\textwidth]{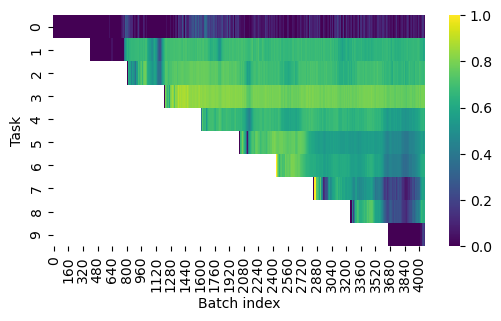}%
    \label{fig:elliptic_mas}}
\hfill
\subfloat[PDGNN]{%
    \includegraphics[width=0.32\textwidth]{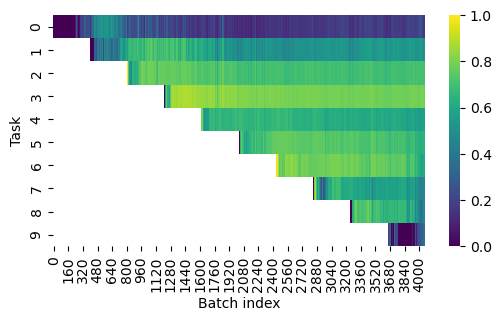}%
    \label{fig:elliptic_pdgnn}}
\\
\subfloat[SSM-A-GEM]{%
    \includegraphics[width=0.32\textwidth]{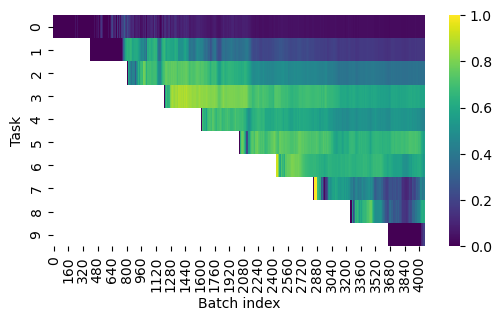}%
    \label{fig:elliptic_ssmgem}}
\hfill
\subfloat[SSM-ER]{%
    \includegraphics[width=0.32\textwidth]{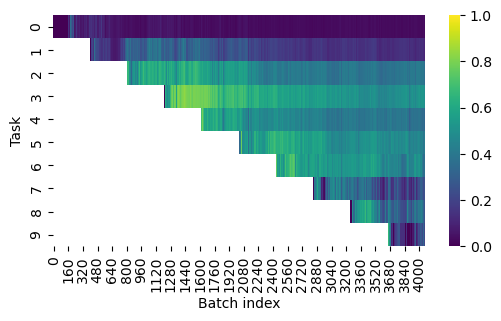}%
    \label{fig:elliptic_ssm}}
\hfill
\subfloat[TWP]{%
    \includegraphics[width=0.32\textwidth]{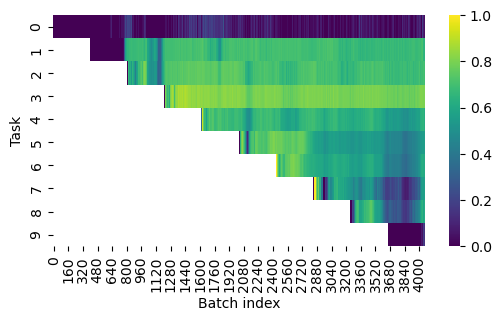}%
    \label{fig:elliptic_twp}}
\\
\subfloat[LINEAR]{%
    \includegraphics[width=0.32\textwidth]{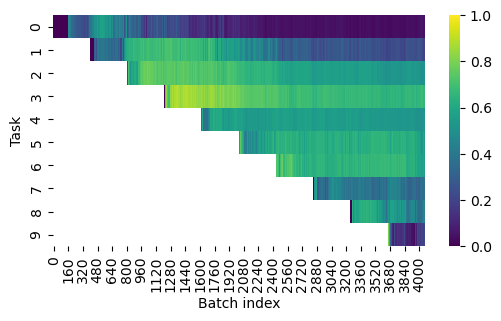}%
    \label{fig:elliptic_malc}}
\hspace{2em}
\subfloat[bare]{%
    \includegraphics[width=0.32\textwidth]{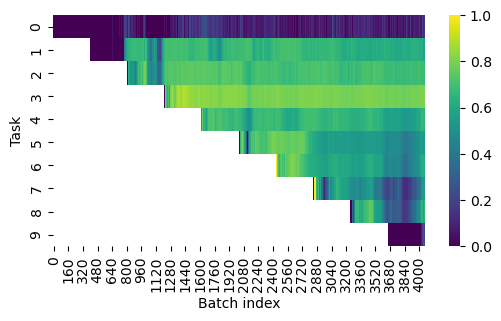}%
    \label{fig:elliptic_bare}}
\caption{Anytime evaluation by task for the Elliptic dataset.}
\label{fig:heatmaps_elliptic}
\end{figure*}

\begin{figure*}[!t]
\centering
\subfloat[A-GEM]{%
    \includegraphics[width=0.32\textwidth]{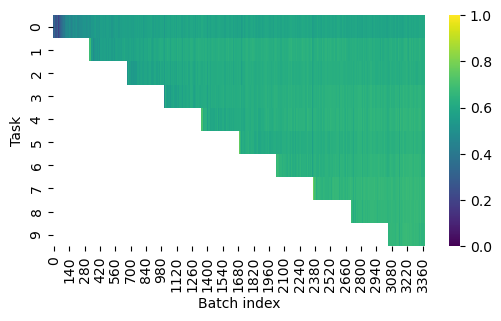}%
    \label{fig:arxivTI_agem}}
\hfill
\subfloat[ER]{%
    \includegraphics[width=0.32\textwidth]{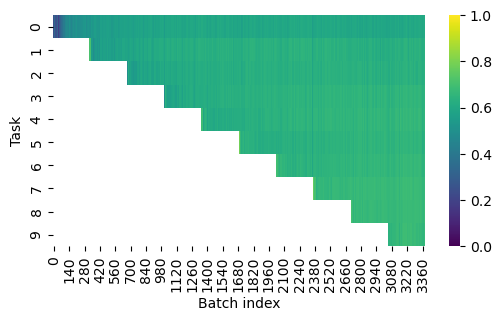}%
    \label{fig:arxivTI_er}}
\hfill
\subfloat[EWC]{%
    \includegraphics[width=0.32\textwidth]{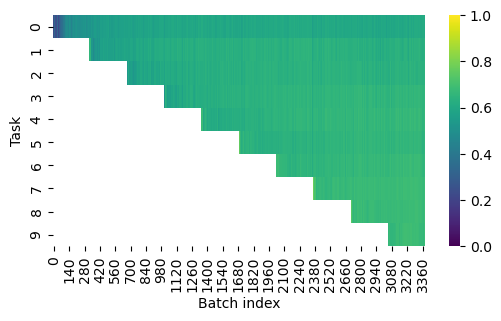}%
    \label{fig:arxivTI_ewc}}
\\
\subfloat[LwF]{%
    \includegraphics[width=0.32\textwidth]{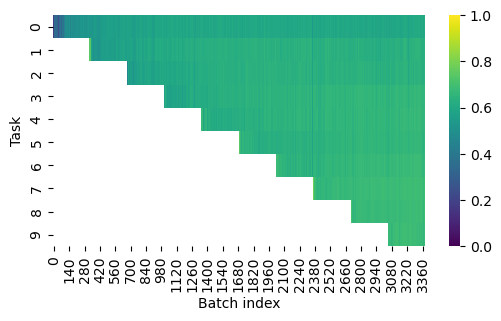}%
    \label{fig:arxivTI_lwf}}
\hfill
\subfloat[MAS]{%
    \includegraphics[width=0.32\textwidth]{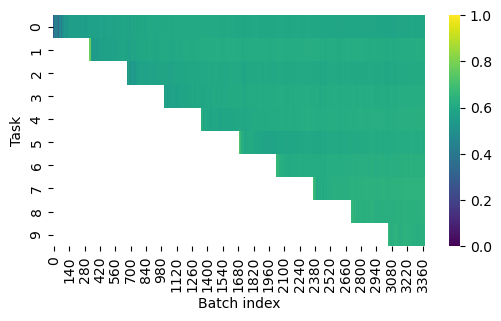}%
    \label{fig:arxivTI_mas}}
\hfill
\subfloat[PDGNN]{%
    \includegraphics[width=0.32\textwidth]{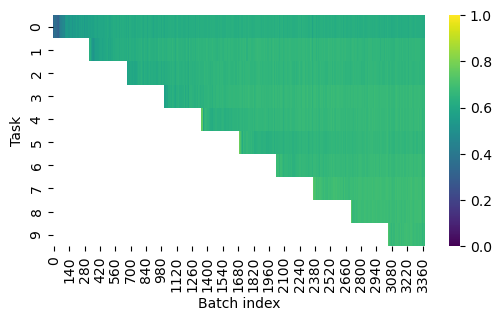}%
    \label{fig:arxivTI_pdgnn}}
\\
\subfloat[SSM-A-GEM]{%
    \includegraphics[width=0.32\textwidth]{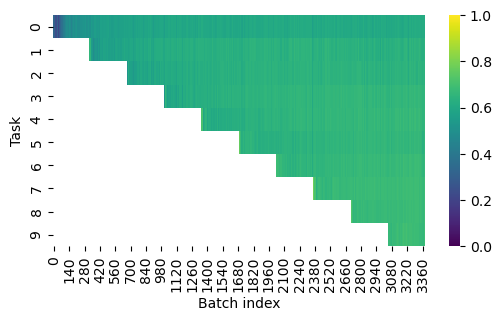}%
    \label{fig:arxivTI_ssmgem}}
\hfill
\subfloat[SSM-ER]{%
    \includegraphics[width=0.32\textwidth]{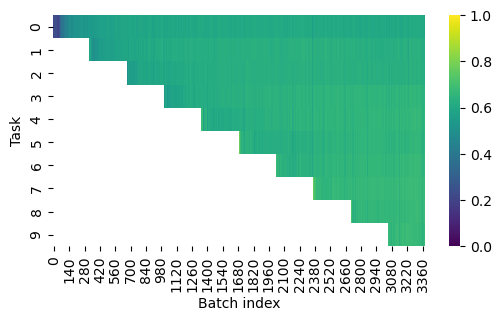}%
    \label{fig:arxivTI_ssm}}
\hfill
\subfloat[TWP]{%
    \includegraphics[width=0.32\textwidth]{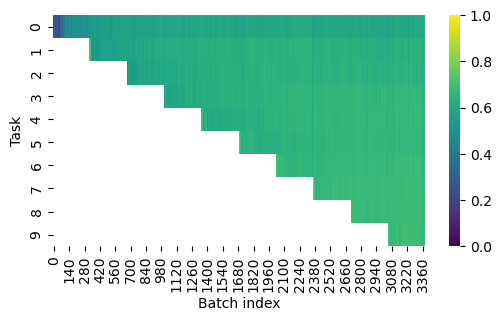}%
    \label{fig:arxivTI_twp}}
\\
\subfloat[LINEAR]{%
    \includegraphics[width=0.32\textwidth]{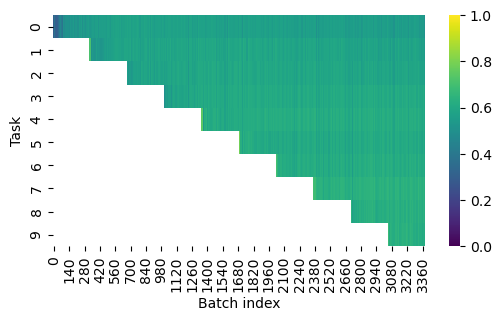}%
    \label{fig:arxivTI_malc}}
\hspace{2em}
\subfloat[bare]{%
    \includegraphics[width=0.32\textwidth]{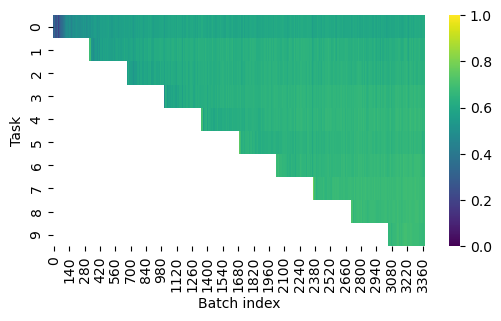}%
    \label{fig:arxivTI_bare}}
\caption{Anytime evaluation by task for the Arxiv dataset with time-incremental stream.}
\label{fig:heatmaps_arxivTI}
\end{figure*}

\begin{figure*}[!t]
\centering
\subfloat[A-GEM]{%
    \includegraphics[width=0.32\textwidth]{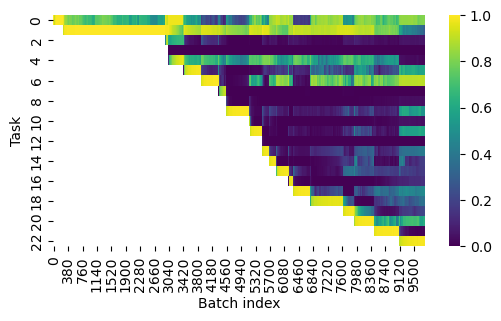}%
    \label{fig:products_agem}}
\hfill
\subfloat[ER]{%
    \includegraphics[width=0.32\textwidth]{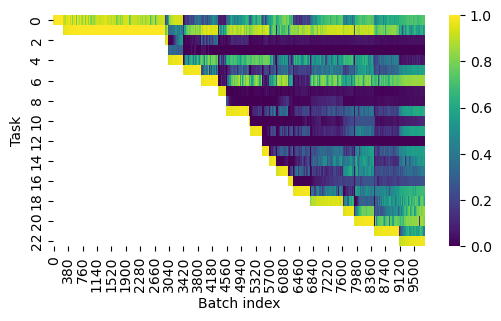}%
    \label{fig:products_er}}
\hfill
\subfloat[EWC]{%
    \includegraphics[width=0.32\textwidth]{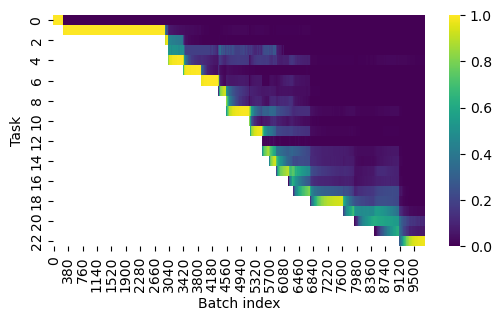}%
    \label{fig:products_ewc}}
\\
\subfloat[LwF]{%
    \includegraphics[width=0.32\textwidth]{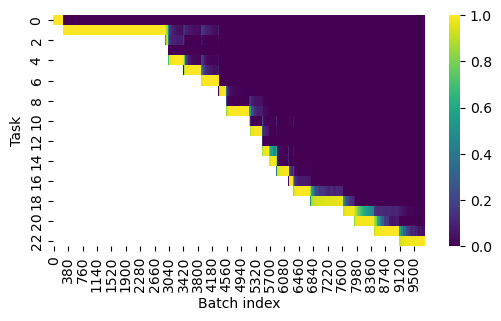}%
    \label{fig:products_lwf}}
\hfill
\subfloat[MAS]{%
    \includegraphics[width=0.32\textwidth]{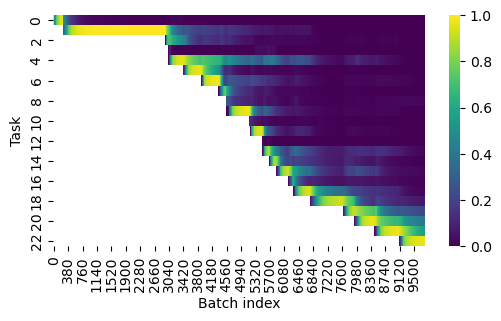}%
    \label{fig:products_mas}}
\hfill
\subfloat[PDGNN]{%
    \includegraphics[width=0.32\textwidth]{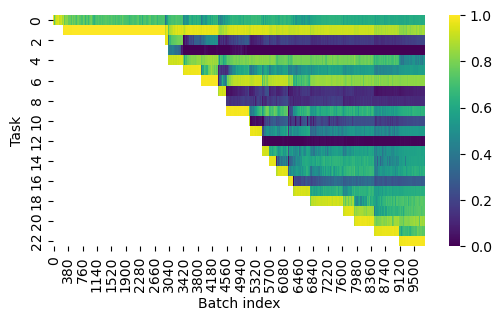}%
    \label{fig:products_pdgnn}}
\\
\subfloat[SSM-A-GEM]{%
    \includegraphics[width=0.32\textwidth]{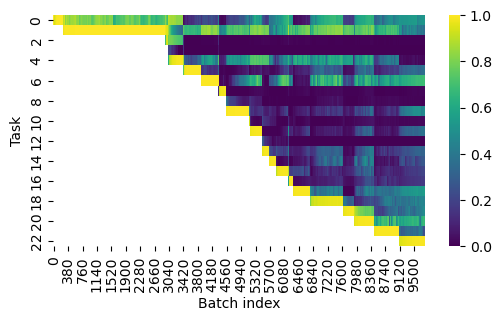}%
    \label{fig:products_ssmgem}}
\hfill
\subfloat[SSM-ER]{%
    \includegraphics[width=0.32\textwidth]{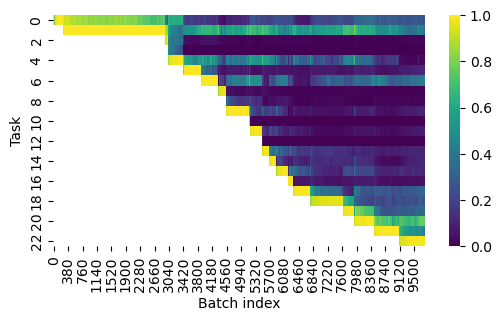}%
    \label{fig:products_ssm}}
\hfill
\subfloat[TWP]{%
    \includegraphics[width=0.32\textwidth]{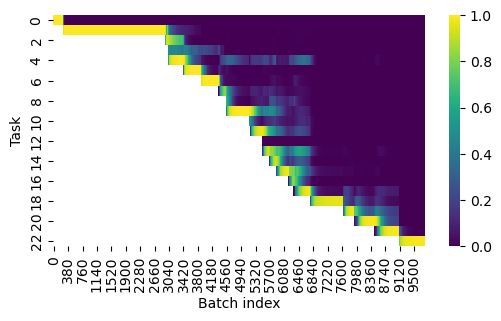}%
    \label{fig:products_twp}}
\\
\subfloat[LINEAR]{%
    \includegraphics[width=0.32\textwidth]{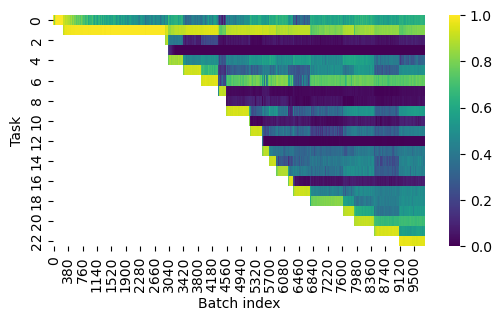}%
    \label{fig:products_malc}}
\hspace{2em}
\subfloat[bare]{%
    \includegraphics[width=0.32\textwidth]{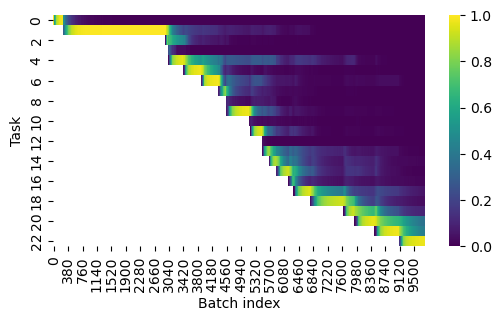}%
    \label{fig:products_bare}}
\caption{Anytime evaluation by task for the Products dataset.}
\label{fig:heatmaps_products}
\end{figure*}

We show here the plots with anytime evaluation, on all datasets and with the settings used in the main experiments for the paper. In Figures \ref{fig:cora_anytime}-\ref{fig:products_anytime} the lines indicate AP on validation nodes after training on each mini-batch, for all considered methods. We highlight the boundaries between tasks, the threshold up to which hyperparameter selection is performed, and the upper bound of joint training up to the current task. We note how it is natural and expected that accuracy tends to decrease with the batch index, as new classes are introduced and the classification task gets increasingly complex.
In general, we observe that while the performance of regularization methods tends to decrease smoothly, the performance of replay-based methods shows much higher variations. 
In Figures \ref{fig:heatmaps_cora}-\ref{fig:heatmaps_products}, instead, we show the performance of each method in more detail, with a breakdown of performance by each task.


\end{document}